\documentclass{article}
\PassOptionsToPackage{numbers,sort&compress}{natbib}

\usepackage[main, final]{neurips_2026}

\usepackage[utf8]{inputenc} 
\usepackage[T1]{fontenc}    
\usepackage{hyperref}       
\usepackage{url}            
\usepackage{booktabs}       
\usepackage{amsfonts}       
\usepackage{nicefrac}       
\usepackage{microtype}      
\usepackage{xcolor}         
\usepackage{microtype}
\usepackage{graphicx}
\usepackage{subcaption}
\usepackage{booktabs} 
\usepackage{amsmath}
\usepackage{amssymb}
\usepackage{mathtools}
\usepackage{amsthm}
\usepackage{hyperref}
\usepackage{url}
\usepackage{enumitem}
\usepackage{booktabs}
\usepackage{tabularx}
\usepackage{array}
\usepackage{float}
\usepackage{multirow}
\usepackage{graphicx}
\usepackage{threeparttable}
\usepackage{algorithm}
\usepackage{algorithmic}

\usepackage{algpseudocode}
\usepackage{subcaption}
\usepackage{caption}
\usepackage[table]{xcolor}
\newcolumntype{Y}{>{\centering\arraybackslash}X}
\usepackage{amsthm}
\newtheorem{proposition}{Proposition}
\theoremstyle{definition}
\newtheorem{definition}{Definition}
\usepackage{xcolor}
\definecolor{myblue}{RGB}{0,120,200}
\usepackage{textcomp}
\usepackage{etoc}
\usepackage{wrapfig}

\theoremstyle{remark}
\newtheorem{remark}{Remark}

\usepackage[capitalize,noabbrev]{cleveref}

\usepackage[textsize=tiny]{todonotes}

\title{
\begin{minipage}[c]{0.11\textwidth}
    \centering
    \includegraphics[width=0.9\linewidth]{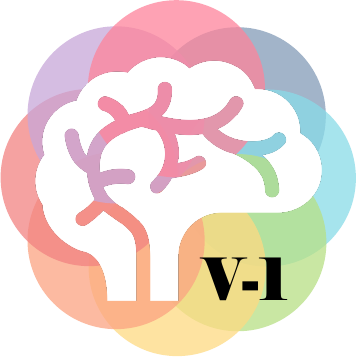}
\end{minipage}
\hspace{0.001\textwidth}
\begin{minipage}[c]{0.85\textwidth}
    \raggedright
    \fontsize{16}{19}\selectfont
    \textbf{BrainPro: Process-conditioned Pretraining with Topographic Retrieval for Large EEG Models}
\end{minipage}
}

\author{%
  Yi Ding\textsuperscript{1,}\thanks{Equal contribution \qquad
  $\dagger$ Corresponding author (ctguan@ntu.edu.sg)} \quad
  Muyun Jiang\textsuperscript{1,}\footnotemark[1] \quad
  Weibang Jiang\textsuperscript{1, 2} \quad
  Shuailei Zhang\textsuperscript{1} \quad
  Xinliang Zhou\textsuperscript{1} \\
  \textbf{Chenyu Liu\textsuperscript{1}} \quad
  \textbf{Shanglin Li\textsuperscript{3}} \quad
  \textbf{Yong Li\textsuperscript{1, 4}} \quad
  \textbf{Cuntai Guan\textsuperscript{1,$\dagger$}} \\
  \textsuperscript{1}Nanyang Technological University,
  \textsuperscript{2}Shanghai Jiao Tong University \\
  \textsuperscript{3}Berlin Institute for the Foundations of Learning and Data,
  \textsuperscript{4}Southeast University
}

\begin{document}
\etocdepthtag.toc{main}

\maketitle

\begin{abstract}
EEG foundation models are pretrained on heterogeneous corpora, but two challenges remain underexplored: (1) experimental protocols provide mental-process cues that are rarely used in masked pretraining, and (2) electrode montages vary across datasets, making spatial representations montage-dependent. Existing methods typically learn a single general-purpose representation and model spatial structure implicitly, limiting their ability to exploit process-related information under montage heterogeneity. We introduce \textbf{BrainPro}, a process-conditioned self-supervised pretraining framework that couples topographic retrieval with shared and process-associated representation learning. BrainPro maps dataset-specific montages to a universal channel-region template and retrieves channel- and region-level spatial filters to form a topographically aligned spatial basis. Over this basis, BrainPro learns a shared encoder for general EEG structure and additional affective, motor-related, and auxiliary residual branches for process-associated variation. Protocol-derived mental-process cues condition branch activation and region-weighted masked reconstruction, incorporating topographic spatial priors and process information. Across nine public BCI benchmarks, BrainPro achieves strong performance among evaluated baselines. Ablations, channel-drop analysis, encoder-configuration studies, and spatial-filter visualizations suggest that topographic retrieval and process-conditioned representation learning jointly improve EEG decoding. Our code is available at \url{https://github.com/yi-ding-cs/BrainPro}.

\end{abstract}

\section{Introduction}
\label{sec:intro}
Electroencephalography (EEG) measures scalp electrical activity and provides a non-invasive, indirect view of neural dynamics associated with cognition, affect, and behavior~\citep{10646518}. 
Different mental processes can involve both shared and process-associated neural substrates. 
For example, affective processing can engage motor-related cortical regions that overlap with motor processes~\citep{SCHNITZLER1997201}, while also showing associations with frontal and temporal regions~\citep{10.1093/cercor/bhaa373,Greene2023}. 
Motor-related processes, in contrast, are strongly associated with sensorimotor rhythms and central scalp regions~\citep{SCHNITZLER1997201,Greene2023}. 
Although EEG reflects these neural patterns indirectly and at coarse spatial resolution, it can capture coarse process-associated differences in spectral activity and scalp distributions~\citep{Greene2023}. 
These observations motivate EEG representations that preserve both shared dynamics and spatial-spectral variation associated with different mental processes.

Recent EEG foundation models (EFMs) improve transferability mainly through large-scale masked pretraining~\citep{zhou2025brainfoundationmodelssurvey,jiang2024large,NEURIPS2024_4540d267,wang2025cbramod,NEURIPS2023_f6b30f3e}. 
However, two challenges remain underexplored. 
First, existing EFMs typically learn a single general-purpose representation, although multi-dataset EEG corpora contain protocol-derived information about the recording context, such as affective stimulation~\citep{zheng2015investigating} and motor imagery~\citep{10.1093/gigascience/giz002}. 
We define such information a \emph{mental-process cue}: it indicates the dominant mental process types associated with an EEG segment. Ignoring these cues may mix general and process-associated variation in one representation space, limiting process-relevant features for downstream decoding.

Second, EEG datasets differ substantially in electrode montage, channel count, spacing, and regional coverage~\citep{jiang2024large}. 
The same scalp region may correspond to different channel sets across datasets, and some regions may be missing in low-density montages. 
Thus, implicit spatial mechanisms such as channel embeddings or self-attention may learn montage-dependent spatial patterns~\citep{jiang2024large,wang2025cbramod}. 
This is especially problematic for process-conditioned pretraining: without an aligned spatial basis, process-associated branches may capture dataset-specific channel patterns rather than reusable scalp-level structure. These challenges are coupled. Without process conditioning, large EEG models may underuse protocol-derived mental-process information that helps to learn representation useful for downstream decoding; without topographic alignment, process conditioning may amplify montage-specific spatial biases. 
This motivates our central question:
\textbf{How can large EEG models learn process-conditioned representations over a topographically aligned spatial basis?}

We propose \textbf{BrainPro}, a process-conditioned self-supervised pretraining framework with topographic retrieval. 
BrainPro maps dataset-specific montages to a universal channel-region template and retrieves channel- and region-level spatial filters according to the available electrodes in the EEG sample. This allows EEG signals recorded from the same or nearby scalp regions to share the same spatial filter weights, forming a montage-consistent spatial basis. 
Using this basis, BrainPro learns a shared encoder for the general EEG structure and additional affective, motor-related, and auxiliary residual branches for process-associated variation. 
During masked pretraining, the mental-process cue determines the active process-associated branch and the regional reconstruction prior. 
The auxiliary residual branch handles heterogeneous non-affect/non-motor data.
We evaluate whether this design improves downstream decoding on unseen datasets with related or overlapping EEG structure.

The main contributions of this work are:
\begin{itemize}
    \item \textbf{Topographically aligned spatial retrieval.} 
    We design a channel- and region-level retrieval mechanism that maps heterogeneous EEG montages to a universal channel-region template, enabling spatial filters to be shared according to scalp-topographic correspondence across datasets.

    \item \textbf{Process-conditioned self-supervised pretraining.} 
    We propose a multi-branch pretraining architecture in which protocol-derived mental-process cues determine the active process-associated branch during masked pretraining, allowing BrainPro to learn shared, affective, motor-related, and auxiliary residual representations.

    \item \textbf{Process-guided region-weighted reconstruction.} 
    We incorporate coarse neurophysiological priors into masked EEG reconstruction by emphasizing scalp regions associated with affective and motor-related processes, while using uniform weights for residual data.

    \item \textbf{Broad empirical evaluation.} 
    We evaluate BrainPro on nine public BCI benchmarks spanning six task families. 
    Ablations, channel-drop analysis, encoder-configuration studies, and spatial-filter visualizations suggest that topographic retrieval and process-conditioned self-supervised pretraining improve EEG decoding.
\end{itemize}

\begin{figure*}[t!]
    \centering
    \includegraphics[width=\linewidth]{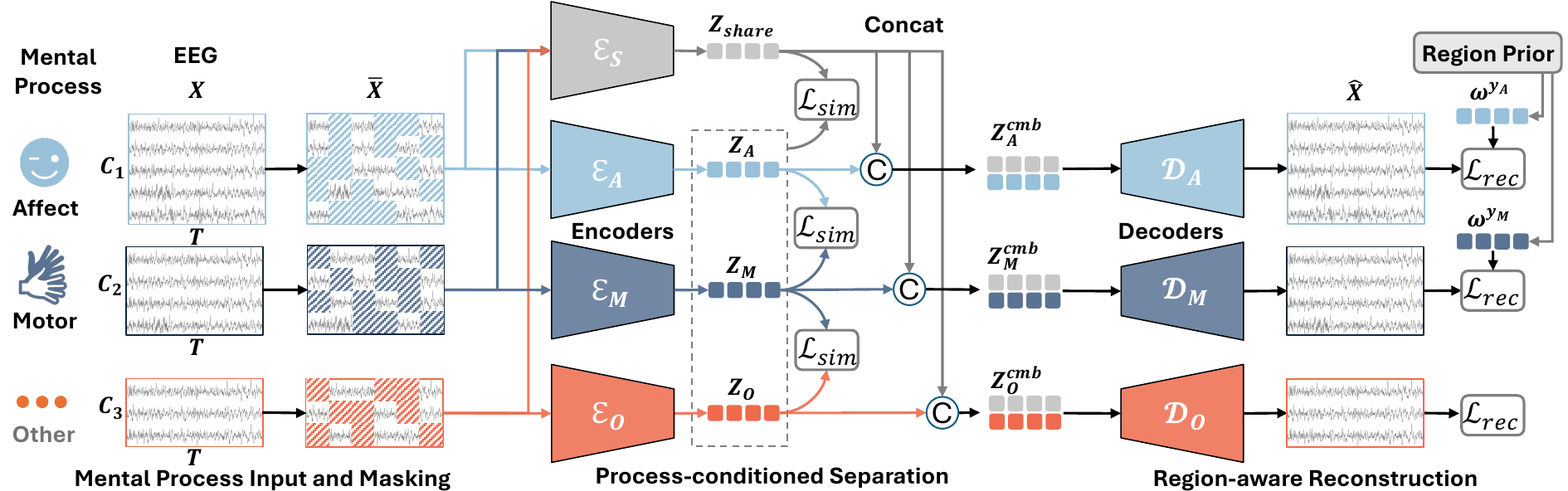}
    \caption{BrainPro framework. BrainPro consists of a shared encoder $\mathcal{E}_{S}$, affective and motor process-associated encoders ($\mathcal{E}_{A}$ and $\mathcal{E}_{M}$), and an auxiliary residual encoder $\mathcal{E}_{O}$ for heterogeneous non-affect/non-motor data. Pretraining combines process-guided region-weighted reconstruction and representation-level branch separation to learn shared and process-associated representations over a topographically aligned spatial basis.}
    \label{fig:framework}
\end{figure*}

\section{Preliminaries}
\label{sec:preliminaries}

\begin{definition}[Mental-process Cue]
A \emph{mental-process cue} denotes dataset- or protocol-derived information indicating the dominant mental process family associated with an EEG segment, such as affective or motor-related activity.
\end{definition}

\begin{remark}[Role of mental-process cues]
In BrainPro, the mental-process cue conditions the active process-associated branch and the region-weighted reconstruction prior during masked pretraining. 
\end{remark}

\begin{definition}[Overlapping Mental Processes]
Different mental processes may recruit partially overlapping cortical regions even when they arise from distinct behavioral or cognitive conditions. 
For example, affective processes can involve motor-related regions, while motor-related processes are strongly associated with sensorimotor areas~\citep{SCHNITZLER1997201,10.1093/cercor/bhaa373}. 
We say two mental processes overlap when their EEG patterns contain partially shared scalp-level spatial structure.
\end{definition}

\begin{remark}[Shared vs.\ Process-associated Representations]
Because EEG signals contain both common neural dynamics and process-related variation, it is useful to distinguish between two types of representations: 
(i) \emph{shared representations}, which capture signal structure that appears across datasets and mental processes, and 
(ii) \emph{process-associated representations}, which capture variation more closely related to affective or motor-related activity. 
Modeling both aspects can provide complementary information for downstream EEG decoding.
\end{remark}

\begin{definition}[Explicit Spatial Encoding]
\emph{Explicit spatial encoding} refers to learning spatial filters whose parameters are tied to scalp channel or region locations, instead of dataset-specific channel indices. 
Channels from different datasets are mapped to the same or nearby locations on a universal scalp-electrode template, retrieving shared spatial-filter parameters. 
This encourages spatial representations that are more consistent across heterogeneous electrode montages and provides an aligned basis for mental-process-conditioned representation learning. 
Further justification is provided in Appendix~\ref{app:Justify_spatial_learning}.
\end{definition}

\paragraph{Existing EFM Pretraining}
Most EFMs rely on masked reconstruction or related self-supervised objectives to learn general-purpose representations. 
At a high level, an encoder $f_{\theta}$ maps an input segment $\mathbf{X}$ to a latent representation $\mathbf{Z}$:
\[
\mathbf{Z} = f_{\theta}(\mathbf{X}).
\]
These approaches capture broad EEG signal structure, but typically learn a single general-purpose representation without using protocol-derived mental-process cues. 
Spatial information is also modeled implicitly, e.g., through channel embeddings or attention, which may not preserve topographic consistency across heterogeneous montages. 
Consequently, process-associated variation and montage-specific factors may become entangled in one latent space.

\paragraph{BrainPro Pretraining}
BrainPro extends this formulation by combining topographic retrieval with process-conditioned representation learning. 
First, topographic retrieval maps dataset-specific montages to a universal channel-region template, producing spatial features that are comparable across electrode layouts. 
Second, BrainPro represents each EEG segment through a shared encoder and a process-conditioned pathway, where the mental-process cue determines which additional process-associated branch contributes to the pretraining objective:
\[
\mathbf{Z} = f_{\mathrm{shared}}(\mathbf{X}) 
\;\Vert\; 
f_{y_{\mathrm{proc}}}(\mathbf{X}),
\]
where $y_{\mathrm{proc}} \in \{\mathrm{A}, \mathrm{M}, \mathrm{O}\}$ denotes the process-conditioned branch used for that segment. 
Here, $\mathrm{A}$ and $\mathrm{M}$ correspond to affective and motor-related process cues, respectively, while $\mathrm{O}$ denotes an auxiliary residual branch for heterogeneous non-affect/non-motor data. 
This formulation preserves a general EEG representation while learning additional process-associated pathways over an topographically aligned spatial basis.

\section{Method}
\label{sec:method}

\subsection{Overview of BrainPro Framework}
The overall BrainPro framework is illustrated in Figure~\ref{fig:framework}. It integrates topographic retrieval, process-conditioned branch separation, and process-guided region-weighted reconstruction into a unified pretraining pipeline (Algorithm~\ref{alg:brainpro}). It consists of three main steps.

\begin{figure*}[t!]
    \centering
    \includegraphics[width=\linewidth]{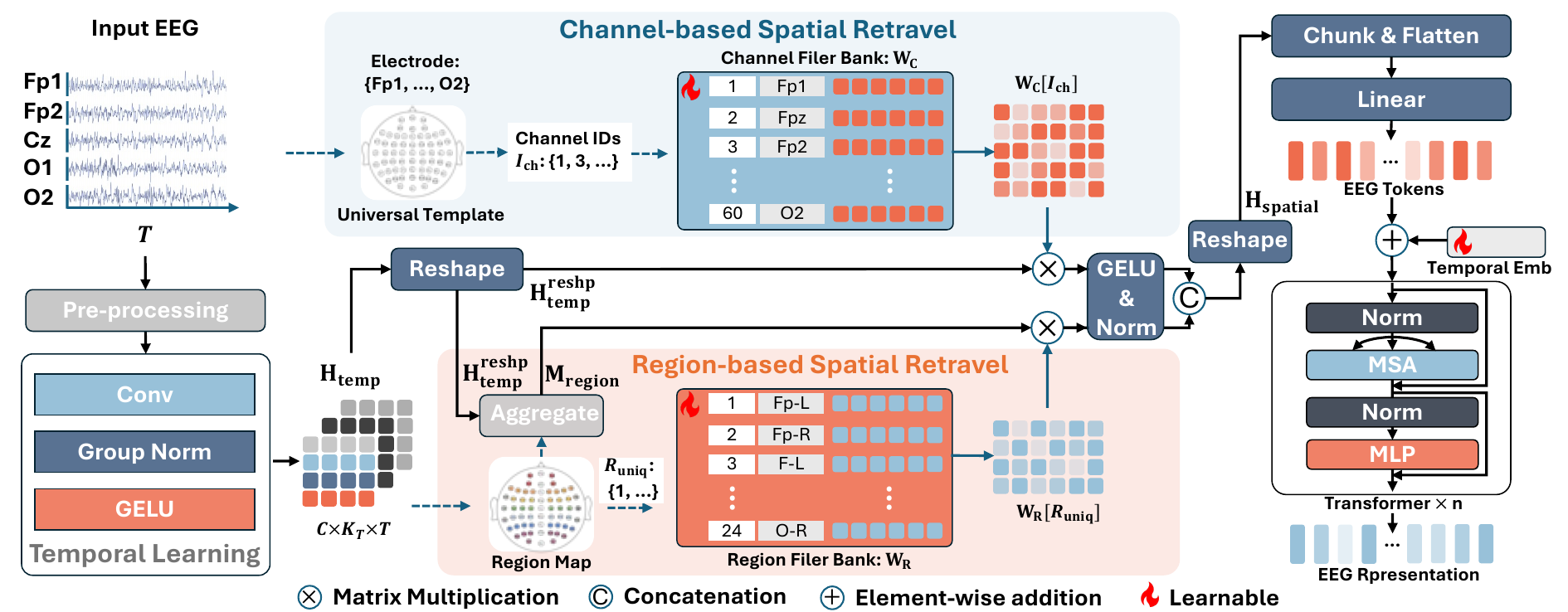}
    \caption{EEG encoder of BrainPro. Each encoder consists of three stages: (1) a temporal learning encoder based on temporal CNNs; (2) a topographic retrieval encoder that performs channel- and region-based spatial retrieval; and (3) patchification and token embedding, followed by Transformer encoders for spatiotemporal modeling.}
    \label{fig:encoder}
\end{figure*}

\textbf{Step-1.} Given an EEG segment, the model first extracts temporal features and retrieves channel- and region-level spatial filters based on a universal template, enabling explicit and montage-robust spatial learning. These features are then processed by a Transformer encoder to obtain spatiotemporal representations (Section~\ref{sec:retrieval_method} and Figure~\ref{fig:encoder}).

\textbf{Step-2.} To learn complementary shared and process-associated representations, BrainPro uses one shared encoder, affective and motor process-associated encoders, and an auxiliary residual encoder. A margin-based cosine similarity loss discourages excessive similarity among the shared, active, and inactive branch representations.

\textbf{Step-3.} To encourage process-relevant spatial structure, reconstruction is performed using both shared and active branch representations. The reconstruction loss is weighted by coarse regional priors for affective and motor-related samples, while the auxiliary residual branch uses uniform weights because no coherent regional prior is assumed.

\subsection{Topographic Retrieval}
\label{sec:retrieval_method}

\paragraph{\textit{Motivation.}}
EEG datasets use heterogeneous montages, making direct channel-index embeddings difficult to share across datasets. Process-conditioned pretraining is more meaningful when different datasets are first aligned to a common spatial basis.

\paragraph{\textit{Operation.}}
We adopt a topographic retrieval strategy. We define a universal template with $C_{\mathrm{pre}}{=}60$ electrodes and $N_\mathrm{region}{=}24$ functional regions following~\cite{10025569}. More details about the template are in Appendix~\ref{app:brian-region}. We maintain learnable spatial filter banks aligned to the positions in the universal template for both \textit{channel-} and \textit{region-level} retrieval:
\begin{align}
    \mathbf{W}_\mathrm{C} &\in \mathbb{R}^{C_\mathrm{pre} \times K_\mathrm{C}}, \quad
    \mathbf{W}_\mathrm{R} \in \mathbb{R}^{N_\mathrm{region} \times K_\mathrm{R}},
\end{align}
where $K_\mathrm{C}$ and $K_\mathrm{R}$ denote the number of channel-wise and region-wise spatial filters.  

Given an input EEG, $X \in \mathbb{R}^{C \times T}$, a temporal encoder, $\mathcal{F}_{temp}$, learns temporal dynamics, $\mathbf{H}_\mathrm{temp}$. More details about $\mathcal{F}_{temp}$ can be found in Appendix~\ref{app:more_details_tcnn}.
We reshape the $\mathbf{H}_\mathrm{temp}$ into $\mathbf{H}_\mathrm{temp}^{\mathrm{reshp}}\in \mathbb{R}^{C \times K_T *T}$, and then extract channel- and region-level features in parallel. 

For \textit{channel}-level features, Let $I_\mathrm{ch} \subseteq \{1,\dots,C_{\mathrm{pre}}\}$ be the set of indices in the universal template that are present in the current sample. This set is obtained by nearest-neighbor mapping in standardized 3D head coordinates. Unmatched channels are discarded (More details in Appendix~\ref{app:new_mapping}). We then retrieve the corresponding filters from the channel filter bank and apply them to the temporal features:
\begin{equation}
    \mathbf{H}_\mathrm{C} = \sigma\!\left(\mathbf{W}_\mathrm{C}[I_\mathrm{ch}]^\top \mathbf{H}_\mathrm{temp}^{\mathrm{reshp}}\right)
    \in \mathbb{R}^{K_\mathrm{C} \times K_{T}* T},
\end{equation}
where $\sigma(\cdot)$ denotes nonlinear operations (GELU and GroupNorm).

For \textit{region}-level features, we aggregate temporal features within each functional region present in the current sample and apply the region filter bank. Let $\mathcal{R}_\mathrm{uniq}$ be the set of functional regions covered by the current sample, and let $C_j$ be the channels belonging to region $j$. We first aggregate temporal features within each region:

\begin{equation}
    \mathbf{M}_\mathrm{region}[j] = \frac{1}{|\mathcal{C}_j|} 
    \sum_{c \in \mathcal{C}_j} \mathbf{H}_\mathrm{temp}^{\mathrm{reshp}}[c, :] 
    \in \mathbb{R}^{1 \times K_T*T}.
\end{equation}
Then we retrieve the corresponding region spatial filters using $\mathcal{R}_\mathrm{uniq}$. The region-wise representations are processed by the region filter bank:
\begin{equation}
    \mathbf{H}_\mathrm{R} = \sigma\!\left(
    \mathbf{W}_\mathrm{R}[\mathcal{R}_\mathrm{uniq}]^\top 
    \mathbf{M}_\mathrm{region}\right) 
    \in \mathbb{R}^{K_\mathrm{R} \times K_{T}*T}.
\end{equation}
Finally, the channel- and region-level features are concatenated and reshaped to form the spatial representation:
\begin{equation}
    \mathbf{H}_\mathrm{spatial} = \mathrm{Reshape}(\mathrm{Concat}\!\left(\mathbf{H}_\mathrm{C},\, \mathbf{H}_\mathrm{R}\right)) 
    \in \mathbb{R}^{K_{T}*K_\mathrm{total} \times T},
\end{equation}
where $\quad K_\mathrm{total} = K_\mathrm{C} + K_\mathrm{R}$. 
This $\mathbf{H}_\mathrm{spatial}$ is further processed by a Transformer encoder to obtain spatiotemporal representations for mental-process-conditioned separation and process-guided region-weighted reconstruction. Details on the tokenization and Transformer encoder are in Appendix~\ref{app:more_details_transformer}.

\paragraph{\textit{Effect.}}
This retrieval mechanism produces montage-consistent spatial representations and allows spatial parameters to be shared according to scalp location.

\subsection{Process-conditioned Representation Separation}
\label{sec:decoupling_method}

\paragraph{\textit{Motivation.}}
EEG contains both shared signal structure and process-associated variation, but standard masked pretraining may mix them in one representation space. BrainPro uses protocol-derived process cues to separate shared and process-associated representations, with the latter potentially capturing complementary information beyond the shared representation.

\paragraph{\textit{Operation.}}
As shown in Figure~\ref{fig:framework}, BrainPro contains one shared encoder $\mathcal{E}_\mathrm{S}$, two primary process-associated encoders $\mathcal{E}_\mathrm{A}$ and $\mathcal{E}_\mathrm{M}$ for affective and motor-related EEG as they have relatively stable scalp-level priors and are consistently available across large-scale public EEG datasets~\citep{7946165,PFURTSCHELLER2006153}, and an auxiliary residual encoder $\mathcal{E}_\mathrm{O}$ for heterogeneous non-affect/non-motor data. Given an EEG segment $\mathbf{X}$, the encoders produce:
\begin{equation}
  \mathbf{Z}_\mathrm{shared} = \mathcal{E}_\mathrm{S}(\mathbf{X}), 
  \mathbf{Z}_k = \mathcal{E}_k(\mathbf{X}), \quad k\in\{\mathrm{A, M, O}\} .
\end{equation}

For an EEG sample with mental-process cue $y_\mathrm{proc}$, BrainPro updates the shared encoder $\mathcal{E}_\mathrm{S}$ and the active branch $\mathcal{E}_{y_\mathrm{proc}}$ only. Inactive branch outputs are detached, so they can serve as separation targets without receiving gradients from samples outside their assigned process.

To enforce representation separation, we define a separation loss $\mathcal{L}_{\mathrm{sep}}$ consisting of two margin-based cosine similarity losses, denoted as $\mathcal{L}_{\mathrm{sim}}$. The similarity loss is:
\begin{equation}
\mathcal{L}_{\mathrm{sim}}(\mathbf{a}, \mathbf{b})
= \max\!\big(\mathrm{sim}_{\mathrm{cos}}(\mathbf{a}, \mathbf{b}) - m, 0\big),
\end{equation}
where $m{=}0.1$ is a margin. The first similarity term penalizes similarity between shared and active mental-process representations, and the second penalizes similarity between the active process encoder and all inactive encoders:
\begin{equation}
\mathcal{L}_{\mathrm{sep}}
= \underbrace{\mathcal{L}_{\mathrm{sim}}(\mathbf{Z}_\mathrm{shared}, \mathbf{Z}_{y_\mathrm{proc}})}_{\text{shared-active}}
\;+\;
\underbrace{\sum_{k\neq y_\mathrm{proc}} 
\mathcal{L}_{\mathrm{sim}}(\mathbf{Z}_{y_\mathrm{proc}}, \mathbf{Z}_k)}_{\text{active-inactive}}.
\end{equation}

\paragraph{\textit{Effect.}}
The shared encoder captures general EEG patterns across datasets, while the active branch learns complementary process-associated variation.

\subsection{Masking and Region-aware Reconstruction}
\label{sec:reconstruction_method}

\paragraph{\textit{Motivation.}}
Masked reconstruction provides a self-supervised learning guidance, but treating all channels uniformly may underuse process-related spatial priors. BrainPro therefore conditions reconstruction on protocol-derived process cues and a shared topographic template.

\paragraph{\textit{Operation.}}
We mask the input patches with a ratio of $\rho{=}0.5$~\citep{jiang2024large} and pass the masked signal $\bar{\mathbf{X}}$ through both the shared encoder and the active process-associated encoder. Their outputs are concatenated and decoded:
\begin{equation}
    \hat{\mathbf{X}} = \mathcal{D}_{y_{\mathrm{proc}}}\!\left(\mathcal{E}_\mathrm{S}(\bar{\mathbf{X}}) \,\|\, \mathcal{E}_{y_{\mathrm{proc}}}(\bar{\mathbf{X}})\right),
\end{equation}
where masked positions in $\bar{\mathbf{X}}$ are replaced with learned \texttt{[MASK]} embeddings and $\hat{\mathbf{X}}\in\mathbb{R}^{C_{\mathrm{pre}}\times T}$ is the reconstructed signal in the 60-channel template space.

We assign each active branch $y_{\mathrm{proc}}$ a channel-importance vector $w^{(y_{\mathrm{proc}})} \in [0,1]^{C_{\mathrm{pre}}}$ to emphasize coarse process-relevant regions. Based on neuroscience priors, frontal, temporal, and central channels are emphasized for affective samples~\citep{7946165,gao2021novel}, while central and parietal channels are emphasized for motor-related samples~\citep{PFURTSCHELLER2006153}. For the auxiliary residual branch encoder, all channels receive equal importance. We apply a smooth weighting function:
\begin{equation}
    \mathrm{weights}(w) = 0.5 + \sigma\!\big(\tau \cdot (w - 0.5)\big),
\end{equation}
with sharpness parameter $\tau{=}\text{epoch}$ to gradually increase weighting strength.

Let $S(\mathbf{X})\subseteq\{1,\dots,C_{\mathrm{pre}}\}$ denote the retained template-channel indices observed in sample $\mathbf{X}$. The reconstruction loss is computed on these available mapped channels:
\begin{equation}
    \label{eq:rec}
    \mathcal{L}_{\mathrm{rec}}
    =
    \frac{1}{|S(\mathbf{X})|T}
    \sum_{i\in S(\mathbf{X})}
    \sum_{t=1}^{T}
    \mathrm{weights}\!\big(w_i^{(y_\mathrm{proc})}\big) \cdot 
    \left(
    \mathbf{X}_{i,t}
    -
    \hat{\mathbf{X}}_{i,t}
    \right)^2.
\end{equation}
The weighted MSE is computed over all available retained channels and time points, while absent or unmapped template channels are excluded.

The overall pretraining objective combines reconstruction and separation:
\begin{equation}
    \mathcal{L}_{\mathrm{BrainPro}} = \mathcal{L}_{\mathrm{rec}} + \mathcal{L}_{\mathrm{sep}}.
\end{equation}

\paragraph{\textit{Effect.}}
This design preserves a unified topographic reconstruction space, avoids forcing the model to hallucinate unobserved channels, and injects coarse affective or motor-related spatial priors into self-supervised pretraining.

\section{Experiment}
\subsection{Datasets and Pre-processing}
\paragraph{Pretraining.}
We construct a large-scale pretraining corpus from diverse EEG datasets. Affective and motor-related data are used as the two primary mental-process groups because they have relatively stable scalp-level priors.
All remaining datasets are assigned to an auxiliary residual group, which absorbs heterogeneous non-affect/non-motor variability during pretraining. 
The resulting corpus contains about \textbf{2,400 hours} of EEG recordings, with a complete dataset list and grouping provided in Appendix~\ref{app:pretrainingdatasets}. 
All recordings are segmented into 10-second clips, resampled to 200 Hz, and scaled to 0.1 mV following~\cite{jiang2024large}. 
Noisy segments with absolute values greater than 10 are removed, yielding approximately \textbf{2,180 hours} of clean data. 
Further implementation details are provided in Appendix~\ref{app:experiment_settings_pretraining}.

\newcommand{\std}[1]{%
  {\raisebox{0.15ex}{\fontsize{7}{7}\selectfont\textpm\hspace{0.06em}#1}}%
}

\begin{table*}[t]
\centering
\caption{Comparison results of different methods on downstream tasks.}
\label{tab:main}

{\setlength{\tabcolsep}{5pt}
\renewcommand{\arraystretch}{1.2}

\resizebox{\textwidth}{!}{
\begin{threeparttable}
\begin{tabular}{l ccc ccc ccc}
\toprule
\multirow{2}{*}{\textbf{Methods}}
& \multicolumn{3}{c}{\textbf{FACED (9-Class)}}
& \multicolumn{3}{c}{\textbf{SEED-V (5-Class)}}
& \multicolumn{3}{c}{\textbf{BCI-IV-2a (4-Class)}} \\
\cmidrule(lr){2-4} \cmidrule(lr){5-7} \cmidrule(lr){8-10}
& ACC-B & Kappa & F1-W
& ACC-B & Kappa & F1-W
& ACC-B & Kappa & F1-W \\
\midrule

EEGNet
& 0.3692 \std{0.0103}
& 0.2880 \std{0.0113}
& 0.3693 \std{0.0111}
& 0.2408 \std{0.0031}
& 0.0536 \std{0.0042}
& 0.1908 \std{0.0088}
& 0.5521 \std{0.0183}
& 0.4028 \std{0.0244}
& 0.5346 \std{0.0228} \\

Conformer
& 0.4566 \std{0.0108}
& 0.3849 \std{0.0121}
& 0.4555 \std{0.0098}
& 0.3087 \std{0.0095}
& 0.1294 \std{0.0134}
& 0.2837 \std{0.0229}
& 0.4879 \std{0.0183}
& 0.3171 \std{0.0244}
& 0.4561 \std{0.0206} \\

BIOT
& 0.2992 \std{0.0119}
& 0.2105 \std{0.0147}
& 0.2954 \std{0.0149}
& 0.3575 \std{0.0052}
& 0.1971 \std{0.0081}
& 0.3636 \std{0.0078}
& 0.2392 \std{0.0280}
& -0.0144 \std{0.0373}
& 0.1063 \std{0.0060} \\

EEGPT
& 0.2809 \std{0.0116}
& 0.1912 \std{0.0129}
& 0.2810 \std{0.0120}
& 0.2421 \std{0.0048}
& 0.0557 \std{0.0077}
& 0.2438 \std{0.0055}
& 0.3849 \std{0.0226}
& 0.1799 \std{0.0301}
& 0.3272 \std{0.0336} \\

LaBraM
& 0.5224 \std{0.0116}
& 0.4610 \std{0.0126}
& 0.5259 \std{0.0108}
& 0.3986 \std{0.0200}
& 0.2491 \std{0.0256}
& 0.4040 \std{0.0197}
& 0.5255 \std{0.0329}
& 0.3674 \std{0.0439}
& 0.5095 \std{0.0405} \\

CBraMod
& 0.5669 \std{0.0094}
& 0.5112 \std{0.0110}
& 0.5729 \std{0.0105}
& 0.3960 \std{0.0033}
& 0.2521 \std{0.0048}
& 0.4050 \std{0.0052}
& 0.5148 \std{0.0402}
& 0.3530 \std{0.0536}
& 0.5032 \std{0.0512} \\

CodeBrain
& 0.5801 \std{0.0089}
& 0.5242 \std{0.0114}
& 0.5821 \std{0.0115}
& 0.4025 \std{0.0104}
& 0.2564 \std{0.0123}
& 0.4095 \std{0.0101}
& 0.5024 \std{0.0109}
& 0.3366 \std{0.0145}
& 0.4952 \std{0.0120} \\

\cellcolor{cyan!15}BrainPro
& \cellcolor{cyan!15}\textbf{0.5937} \std{0.0087}
& \cellcolor{cyan!15}\textbf{0.5418} \std{0.0092}
& \cellcolor{cyan!15}\textbf{0.6023} \std{0.0061}
& \cellcolor{cyan!15}\textbf{0.4078} \std{0.0075}
& \cellcolor{cyan!15}\textbf{0.2612} \std{0.0089}
& \cellcolor{cyan!15}\textbf{0.4115} \std{0.0071}
& \cellcolor{cyan!15}\textbf{0.5674} \std{0.0148}
& \cellcolor{cyan!15}\textbf{0.4232} \std{0.0198}
& \cellcolor{cyan!15}\textbf{0.5653} \std{0.0169} \\

\midrule

\multirow{2}{*}{\textbf{Methods}}
& \multicolumn{3}{c}{\textbf{SHU (2-Class)}}
& \multicolumn{3}{c}{\textbf{Mental Arithmetic (2-Class)}}
& \multicolumn{3}{c}{\textbf{Attention (2-Class)}} \\
\cmidrule(lr){2-4} \cmidrule(lr){5-7} \cmidrule(lr){8-10}
& ACC-B & AUC-PR & AUROC
& ACC-B & AUC-PR & AUROC
& ACC-B & AUC-PR & AUROC \\
\midrule

EEGNet
& 0.5664 \std{0.0522}
& 0.6609 \std{0.0099}
& 0.6791 \std{0.0085}
& 0.5533 \std{0.0280}
& 0.5702 \std{0.0539}
& 0.5717 \std{0.0456}
& 0.6004 \std{0.0123}
& 0.6294 \std{0.0288}
& 0.6647 \std{0.0180} \\

Conformer
& 0.6167 \std{0.0172}
& 0.6763 \std{0.0054}
& 0.6927 \std{0.0072}
& 0.6867 \std{0.0492}
& 0.7691 \std{0.0286}
& 0.7039 \std{0.0177}
& 0.7198 \std{0.0158}
& 0.7819 \std{0.0218}
& 0.7992 \std{0.0283} \\

BIOT
& 0.4981 \std{0.0044}
& 0.5077 \std{0.0025}
& 0.5069 \std{0.0060}
& 0.5583 \std{0.0306}
& 0.5512 \std{0.0516}
& 0.5662 \std{0.0250}
& 0.6111 \std{0.0411}
& 0.7273 \std{0.0132}
& 0.7367 \std{0.0162} \\

EEGPT
& 0.5481 \std{0.0139}
& 0.5631 \std{0.0165}
& 0.5673 \std{0.0197}
& 0.5650 \std{0.0341}
& 0.5860 \std{0.0805}
& 0.5937 \std{0.0762}
& 0.6674 \std{0.0560}
& 0.8015 \std{0.0372}
& 0.8103 \std{0.0303} \\

LaBraM
& 0.6175 \std{0.0172}
& 0.6821 \std{0.0214}
& 0.6720 \std{0.0253}
& 0.6688 \std{0.0279}
& 0.7504 \std{0.0649}
& 0.7168 \std{0.0452}
& 0.6785 \std{0.0223}
& 0.7838 \std{0.0307}
& 0.7994 \std{0.0198} \\

CBraMod
& 0.6108 \std{0.0233}
& 0.6632 \std{0.0277}
& 0.6668 \std{0.0338}
& 0.7354 \std{0.0410}
& 0.8237 \std{0.0225}
& 0.7654 \std{0.0203}
& 0.6478 \std{0.0258}
& 0.7417 \std{0.0175}
& 0.7468 \std{0.0198} \\

CodeBrain
& 0.6205 \std{0.0089}
& 0.6819 \std{0.0154}
& 0.6833 \std{0.0146}
& 0.7111 \std{0.0210}
& 0.7984 \std{0.0112}
& 0.7510 \std{0.0139}
& 0.6988 \std{0.0134}
& 0.7518 \std{0.0202}
& 0.7782 \std{0.0080} \\

\cellcolor{cyan!15}BrainPro
& \cellcolor{cyan!15}\textbf{0.6319} \std{0.0107}
& \cellcolor{cyan!15}\textbf{0.7102} \std{0.0076}
& \cellcolor{cyan!15}\textbf{0.7105} \std{0.0076}
& \cellcolor{cyan!15}\textbf{0.8083} \std{0.0156}
& \cellcolor{cyan!15}\textbf{0.8980} \std{0.0052}
& \cellcolor{cyan!15}\textbf{0.8512} \std{0.0083}
& \cellcolor{cyan!15}\textbf{0.7222} \std{0.0291}
& \cellcolor{cyan!15}\textbf{0.7975} \std{0.0392}
& \cellcolor{cyan!15}\textbf{0.8064} \std{0.0259} \\

\bottomrule
\end{tabular}

\begin{tablenotes}
\small
\item Note: \textbf{Bold} indicates the best performance. \cellcolor{cyan!15}{Cyan highlight} marks our BrainPro.
\end{tablenotes}
\end{threeparttable}
}
}
\end{table*}

\paragraph{Downstream Evaluation.}
To comprehensively evaluate our method, we consider six representative downstream BCI tasks using nine publicly available datasets: emotion recognition, motor imagery, imagined speech, mental disorder diagnosis, mental stress detection, and attention detection. 
These tasks and their corresponding datasets are summarized in Table~\ref{tab:bci_tasks}, with additional dataset-specific details in Appendix~\ref{app:downstream_dataset}. 
For consistency with pretraining, all EEG signals are resampled to 200 Hz. 
Further preprocessing details are provided in Appendix~\ref{app:downstream_dataset}. 

\subsection{Baselines and Evaluation Metrics}
Following prior studies~\cite{jiang2024large,wang2025cbramod}, we compare BrainPro against both non-foundation and foundation model baselines. All baselines share the same fine-tuning and evaluation setting. 
Among non-foundation approaches, we include EEGNet~\citep{Lawhern_2018} and Conformer~\citep{9991178}.   
For foundation-model baselines, we consider:  
BIOT~\citep{NEURIPS2023_f6b30f3e}, LaBraM~\citep{jiang2024large}, EEGPT~\citep{NEURIPS2024_4540d267}, CBraMod~\citep{wang2025cbramod} and CodeBrain~\citep{ma2026codebrain}.
Further baseline details are in Appendix~\ref{app:baseline}.
For evaluation metrics, we adopt the ones suitable for binary and multi-class tasks.  
For binary classification, we report Balanced Accuracy (ACC-B), AUC-PR, and AUROC.  
For multi-class classification, following prior studies~\citep{jiang2024large,wang2025cbramod}, we report  
Balanced Accuracy (ACC-B), Cohen’s Kappa, and Weighted F1 (F1-W).

\begin{figure*}[t]
    \centering
    \begin{subfigure}{0.13\textwidth}
        \includegraphics[width=\linewidth]{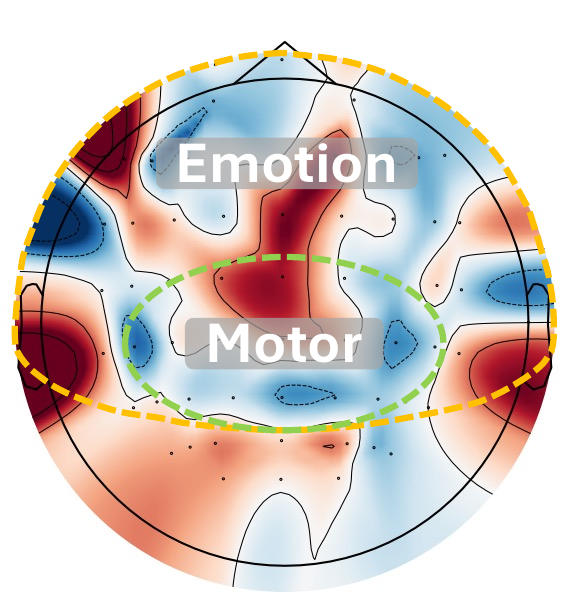}
        \caption{Affect}
    \end{subfigure}
    \begin{subfigure}{0.13\textwidth}
        \includegraphics[width=\linewidth]{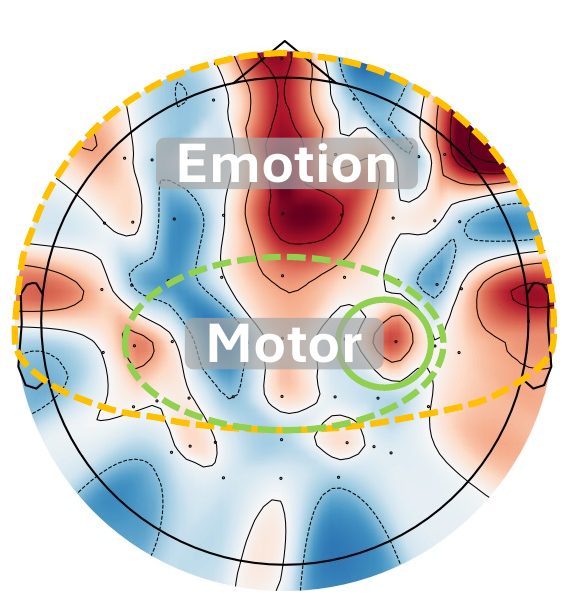}
        \caption{Affect}
    \end{subfigure}
    \begin{subfigure}{0.13\textwidth}
        \includegraphics[width=\linewidth]{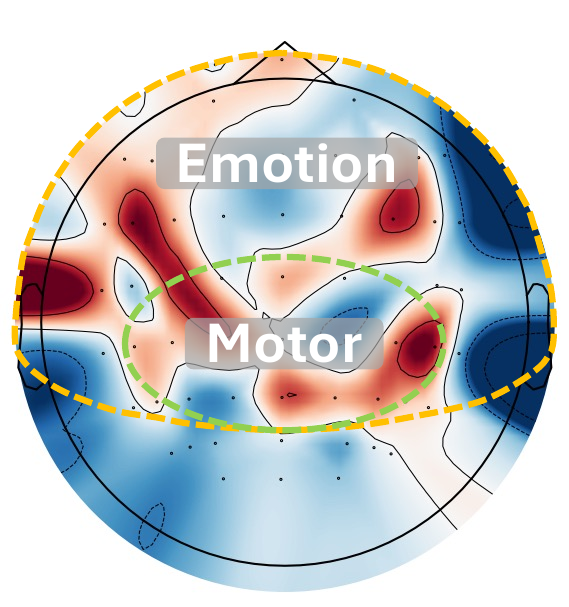}
        \caption{Motor}
    \end{subfigure}
    \begin{subfigure}{0.13\textwidth}
        \includegraphics[width=\linewidth]{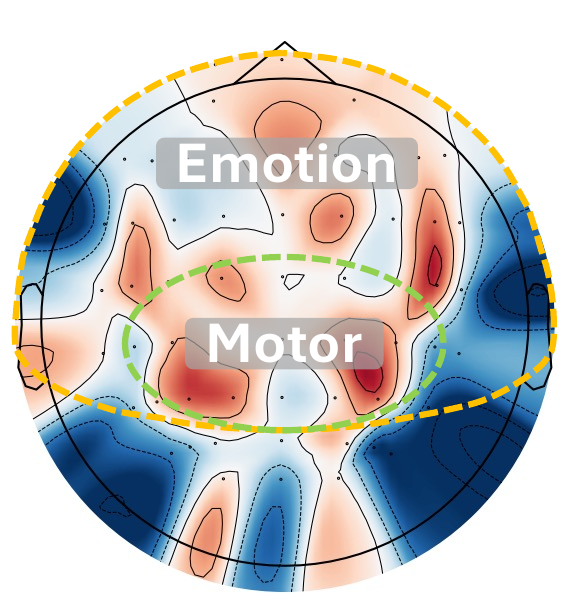}
        \caption{Motor}
    \end{subfigure}
    \begin{subfigure}{0.13\textwidth}
        \includegraphics[width=\linewidth]{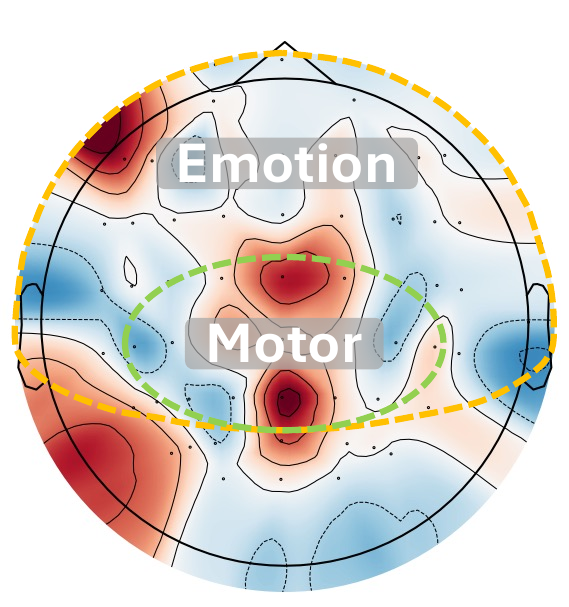}
        \caption{Shared}
    \end{subfigure}
    \begin{subfigure}{0.13\textwidth}
        \includegraphics[width=\linewidth]{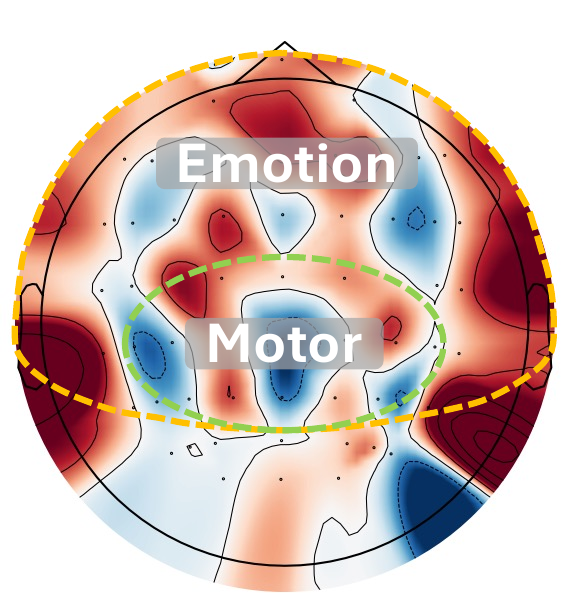}
        \caption{Shared}
    \end{subfigure}
    \caption{Visualization of the learned spatial filters for the Affect, Motor, and Shared encoders after pretraining. Each subplot shows a 2D scalp topography in which warmer colors indicate larger filter weights and cooler colors indicate smaller weights, reflecting the relative contribution of each scalp region to the encoder’s representation. Emotion- and motor-related regions are outlined.}

    \label{fig:spatial_filters_main_manuscript}
\end{figure*}

\begin{table*}[t]
\centering
\scriptsize
\caption{Ablation studies of the main components on affective and motor-related downstream tasks.}
\label{tab:ablation}
\resizebox{\textwidth}{!}{
\begin{threeparttable}
\begin{tabular}{l ccc ccc}
\toprule
\multirow{2}{*}{\textbf{Methods}}
& \multicolumn{3}{c}{\textbf{FACED (9-Class)}} 
& \multicolumn{3}{c}{\textbf{BCI-IV-2A (4-Class)}} \\
\cmidrule(lr){2-4} \cmidrule(lr){5-7}
& ACC-B & Kappa & F1-W 
& ACC-B & Kappa & F1-W \\
\midrule

w/o reconstruction
& 0.5391 {\raisebox{0.25ex}{\fontsize{5}{5}\selectfont\textpm\hspace{0.08em}0.0151}}
& 0.3854 {\raisebox{0.25ex}{\fontsize{5}{5}\selectfont\textpm\hspace{0.08em}0.0202}}
& 0.5401 {\raisebox{0.25ex}{\fontsize{5}{5}\selectfont\textpm\hspace{0.08em}0.0177}}
& 0.4354 {\raisebox{0.25ex}{\fontsize{5}{5}\selectfont\textpm\hspace{0.08em}0.0329}}
& 0.2472 {\raisebox{0.25ex}{\fontsize{5}{5}\selectfont\textpm\hspace{0.08em}0.0439}}
& 0.3758 {\raisebox{0.25ex}{\fontsize{5}{5}\selectfont\textpm\hspace{0.08em}0.0548}} \\

w/o separation
& 0.5638 {\raisebox{0.25ex}{\fontsize{5}{5}\selectfont\textpm\hspace{0.08em}0.0138}}
& 0.4184 {\raisebox{0.25ex}{\fontsize{5}{5}\selectfont\textpm\hspace{0.08em}0.0185}}
& 0.5603 {\raisebox{0.25ex}{\fontsize{5}{5}\selectfont\textpm\hspace{0.08em}0.0107}}
& 0.4870 {\raisebox{0.25ex}{\fontsize{5}{5}\selectfont\textpm\hspace{0.08em}0.0184}}
& 0.3160 {\raisebox{0.25ex}{\fontsize{5}{5}\selectfont\textpm\hspace{0.08em}0.0246}}
& 0.4725 {\raisebox{0.25ex}{\fontsize{5}{5}\selectfont\textpm\hspace{0.08em}0.0219}} \\

w/ random retrieval
& 0.5747 {\raisebox{0.25ex}{\fontsize{5}{5}\selectfont\textpm\hspace{0.08em}0.0098}}
& 0.4329 {\raisebox{0.25ex}{\fontsize{5}{5}\selectfont\textpm\hspace{0.08em}0.0130}}
& 0.5674 {\raisebox{0.25ex}{\fontsize{5}{5}\selectfont\textpm\hspace{0.08em}0.0068}}
& 0.5125 {\raisebox{0.25ex}{\fontsize{5}{5}\selectfont\textpm\hspace{0.08em}0.0308}}
& 0.3500 {\raisebox{0.25ex}{\fontsize{5}{5}\selectfont\textpm\hspace{0.08em}0.0410}}
& 0.5052 {\raisebox{0.25ex}{\fontsize{5}{5}\selectfont\textpm\hspace{0.08em}0.0379}} \\

w/o spatial prior
& 0.5634 {\raisebox{0.25ex}{\fontsize{5}{5}\selectfont\textpm\hspace{0.08em}0.0271}}
& 0.5060 {\raisebox{0.25ex}{\fontsize{5}{5}\selectfont\textpm\hspace{0.08em}0.0319}}
& 0.5672 {\raisebox{0.25ex}{\fontsize{5}{5}\selectfont\textpm\hspace{0.08em}0.0330}}
& 0.5496 {\raisebox{0.25ex}{\fontsize{5}{5}\selectfont\textpm\hspace{0.08em}0.0143}}
& 0.3994 {\raisebox{0.25ex}{\fontsize{5}{5}\selectfont\textpm\hspace{0.08em}0.0191}}
& 0.5495 {\raisebox{0.25ex}{\fontsize{5}{5}\selectfont\textpm\hspace{0.08em}0.0141}} \\

w/ random process cue
& 0.5269 {\raisebox{0.25ex}{\fontsize{5}{5}\selectfont\textpm\hspace{0.08em}0.0288}}
& 0.4682 {\raisebox{0.25ex}{\fontsize{5}{5}\selectfont\textpm\hspace{0.08em}0.0341}}
& 0.5330 {\raisebox{0.25ex}{\fontsize{5}{5}\selectfont\textpm\hspace{0.08em}0.0329}}
& 0.5232 {\raisebox{0.25ex}{\fontsize{5}{5}\selectfont\textpm\hspace{0.08em}0.0065}}
& 0.3642 {\raisebox{0.25ex}{\fontsize{5}{5}\selectfont\textpm\hspace{0.08em}0.0087}}
& 0.5104 {\raisebox{0.25ex}{\fontsize{5}{5}\selectfont\textpm\hspace{0.08em}0.0127}} \\

w/ single encoder
& 0.5683 {\raisebox{0.25ex}{\fontsize{5}{5}\selectfont\textpm\hspace{0.08em}0.0241}}
& 0.5133 {\raisebox{0.25ex}{\fontsize{5}{5}\selectfont\textpm\hspace{0.08em}0.0281}}
& 0.5735 {\raisebox{0.25ex}{\fontsize{5}{5}\selectfont\textpm\hspace{0.08em}0.0272}}
& 0.5554 {\raisebox{0.25ex}{\fontsize{5}{5}\selectfont\textpm\hspace{0.08em}0.0142}}
& 0.4072 {\raisebox{0.25ex}{\fontsize{5}{5}\selectfont\textpm\hspace{0.08em}0.0189}}
& 0.5500 {\raisebox{0.25ex}{\fontsize{5}{5}\selectfont\textpm\hspace{0.08em}0.0176}} \\

w/o pretraining
& 0.4244 {\raisebox{0.25ex}{\fontsize{5}{5}\selectfont\textpm\hspace{0.08em}0.2751}}
& 0.3512 {\raisebox{0.25ex}{\fontsize{5}{5}\selectfont\textpm\hspace{0.08em}0.3088}}
& 0.4028 {\raisebox{0.25ex}{\fontsize{5}{5}\selectfont\textpm\hspace{0.08em}0.3202}}
& 0.4479 {\raisebox{0.25ex}{\fontsize{5}{5}\selectfont\textpm\hspace{0.08em}0.1066}}
& 0.2639 {\raisebox{0.25ex}{\fontsize{5}{5}\selectfont\textpm\hspace{0.08em}0.1421}}
& 0.4224 {\raisebox{0.25ex}{\fontsize{5}{5}\selectfont\textpm\hspace{0.08em}0.1454}} \\

\cellcolor{cyan!15}BrainPro
& \cellcolor{cyan!15}\textbf{0.5937} {\raisebox{0.25ex}{\fontsize{5}{5}\selectfont\textpm\hspace{0.08em}0.0087}}
& \cellcolor{cyan!15}\textbf{0.5418} {\raisebox{0.25ex}{\fontsize{5}{5}\selectfont\textpm\hspace{0.08em}0.0092}}
& \cellcolor{cyan!15}\textbf{0.6023} {\raisebox{0.25ex}{\fontsize{5}{5}\selectfont\textpm\hspace{0.08em}0.0061}}
& \cellcolor{cyan!15}\textbf{0.5674} {\raisebox{0.25ex}{\fontsize{5}{5}\selectfont\textpm\hspace{0.08em}0.0148}}
& \cellcolor{cyan!15}\textbf{0.4232} {\raisebox{0.25ex}{\fontsize{5}{5}\selectfont\textpm\hspace{0.08em}0.0198}}
& \cellcolor{cyan!15}\textbf{0.5653} {\raisebox{0.25ex}{\fontsize{5}{5}\selectfont\textpm\hspace{0.08em}0.0169}} \\

\bottomrule
\end{tabular}

\begin{tablenotes}
\tiny
\item Note: \textbf{Bold} indicates the best performance. \cellcolor{cyan!15}{Cyan highlight} marks BrainPro.
\end{tablenotes}
\end{threeparttable}
}
\end{table*}

\subsection{Implementation Details}
\paragraph{Pretraining.}
Pretraining is conducted end-to-end. 
We train for 30 epochs using the AdamW optimizer with a cosine learning rate schedule, warmup, and gradient clipping. 
The model architecture consists of temporal and spatial encoders, patch makers, and transformer-based encoder-decoder modules. 
Hyperparameters are chosen to balance efficiency and representational capacity, with the full configuration provided in Appendix~\ref{app:pre_training_hyper_parameters} (Table~\ref{tab:pretraining_hyper_parameters}). 

\paragraph{Downstream Evaluation.}
We split each dataset into non-overlapping train/validation/test sets to avoid potential data leakage~\citep{wang2025cbramod,10.1145/3524499,9762054}. 
Models are trained on the training split, selected using the validation split, and evaluated once on the test split. 
We report mean$\pm$std over five seeds $\{0,1,2,3,4\}$. 
Experiments run on 5 NVIDIA A800 GPUs. 
Hyperparameters are provided in Appendix~\ref{app:downstream_hyper_parameters}. For downstream adaptation the shared encoder is always activated, and the affective, motor-related, or auxiliary residual branch is activated according to the downstream task family. Details are provided in Appendix~\ref{app:flexible_configuration}.
After loading pre-trained weights, we re-initialize temporal position embeddings with Xavier uniform initialization. An ablation is provided in Appendix~\ref{app:reset_t_pos_embd}.

\subsection{Primary Results}
\paragraph{BrainPro improves downstream decoding across diverse BCI tasks.}
Table~\ref{tab:main} compares BrainPro with non-foundation and foundation-model baselines across six downstream BCI task families. 
EEGNet and Conformer perform reasonably on some binary tasks but degrade on more complex multi-class settings, particularly in Kappa and weighted F1. 
BIOT and EEGPT show weaker transferability on several datasets, while LaBraM and CBraMod benefit from larger-scale pretraining. 
BrainPro achieves strong overall performance among the evaluated baselines across the reported benchmarks. 
For example, BrainPro achieves 0.5937/0.5418/0.6023 (ACC-B/Kappa/F1-W) on FACED and 0.8083/0.8980/0.8512 (ACC-B/AUC-PR/AUROC) on Mental Arithmetic. 
More results are provided in Appendix~\ref{app:more_results}. 
These results suggest that BrainPro improves downstream EEG decoding under the reported evaluation protocols. 

\subsection{Interpretation of Learned Spatial Filters}
\paragraph{Spatial filters show process-consistent topographic patterns.}
Figure~\ref{fig:spatial_filters_main_manuscript} visualizes the spatial filters learned by the Affect, Motor, and Shared encoders after pretraining. 
The learned filters show qualitatively distinct scalp patterns. 
The Affect encoder emphasizes frontal and temporal regions, consistent with affective EEG literature~\citep{7946165,gao2021novel}, while the Motor encoder highlights central sensorimotor areas~\citep{PFURTSCHELLER2006153}. 
In contrast, the Shared encoder produces more distributed patterns that may reflect global EEG structure. 
These visualizations provide qualitative evidence that BrainPro learns mental-process-associated spatial structure over an topographically aligned basis. 
More details are provided in Appendix~\ref{app:visual_spatial_filters}.

\subsection{Ablation Studies}\label{sec:ablation}
\noindent\textbf{Ablations support the contribution of major components.}
(1) Removing reconstruction degrades performance, showing that masked reconstruction provides useful self-supervised EEG structure.
(2) Removing separation lowers decoding results, indicating that branch decoupling improves shared and process-associated representation learning.
(3) Random retrieval performs worse than BrainPro, supporting the benefit of topographically aligned spatial filters.
(4) Removing the spatial prior reduces performance, showing that process-guided region weighting adds value beyond retrieval alone.
(5) Random process cues cause clear drops, suggesting that correct protocol-derived cues matter and that the gains are not from arbitrary branching.
(6) The matched-size single encoder remains below BrainPro, indicating that process-conditioned multi-branch specialization is beneficial beyond simply increasing model size.
(7) Removing pretraining leads to large drops and high variance, confirming the importance of large-scale self-supervised pretraining.
Overall, all major components contribute to BrainPro's performance on affective and motor-related downstream tasks.

\begin{figure*}[t]
    \centering
    \begin{subfigure}[b]{0.20\textwidth}
        \centering
        \includegraphics[width=\textwidth]{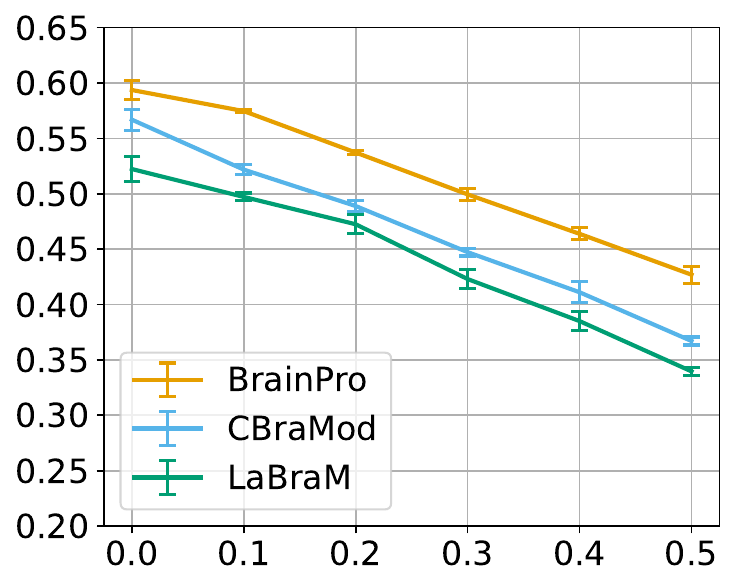}
        \caption{ACC-B}
        \label{fig:acc_noise}
    \end{subfigure}
    \hfill
    \begin{subfigure}[b]{0.20\textwidth}
        \centering
        \includegraphics[width=\textwidth]{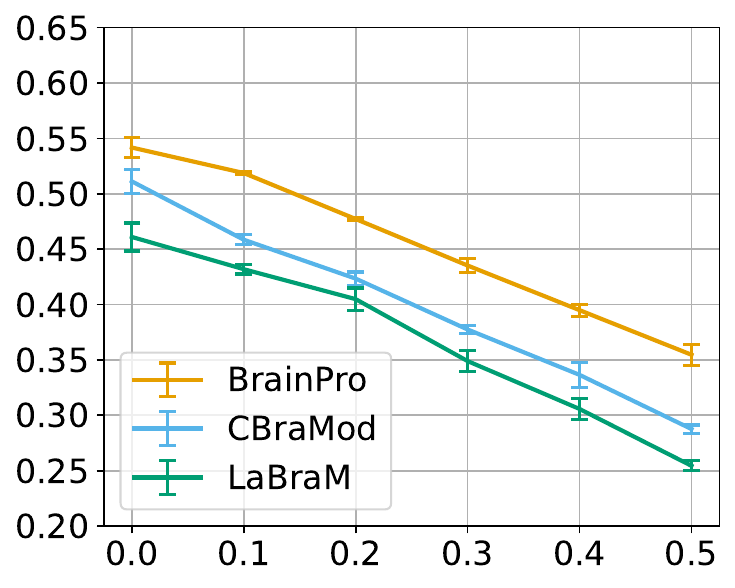}
        \caption{Kappa}
        \label{fig:kappa_noise}
    \end{subfigure}
    \hfill
    \begin{subfigure}[b]{0.20\textwidth}
        \centering
        \includegraphics[width=\textwidth]{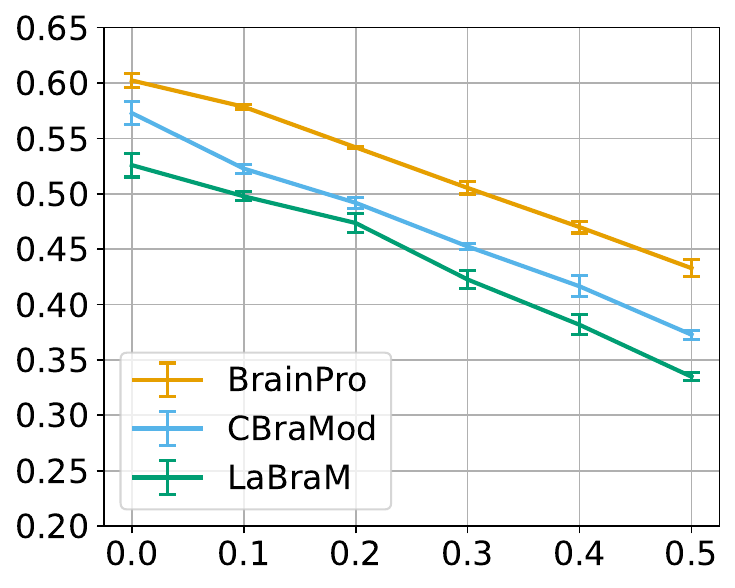}
        \caption{F1-W}
        \label{fig:f1_noise}
    \end{subfigure}
    \hfill
    \begin{subfigure}[b]{0.21\textwidth}
        \centering
        \includegraphics[width=\textwidth]{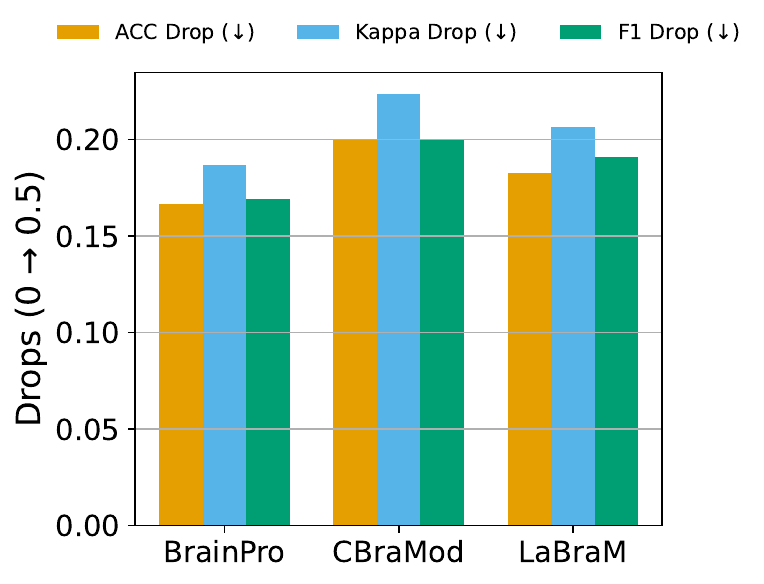}
        \caption{Drop from 0 to 0.5}
        \label{fig:drop_barchart}
    \end{subfigure}

    \caption{Performance of BrainPro, CBraMod, and LaBraM under different levels of channel-drop rate.}
    \label{fig:channel_drop}
\end{figure*}

\begin{table*}[t]
\centering
\scriptsize
\caption{Effects of different encoder configurations on downstream performance.}
\label{tab:encoder_config}

{\setlength{\tabcolsep}{5pt}
\renewcommand{\arraystretch}{1.0}

\resizebox{\textwidth}{!}{
\begin{threeparttable}
\begin{tabular}{l ccc ccc}
\toprule
& \multicolumn{3}{c}{\textbf{FACED (9-Class)}}
& \multicolumn{3}{c}{\textbf{BCI-IV-2A (4-Class)}} \\
\cmidrule(lr){2-4} \cmidrule(lr){5-7}
\textbf{Method}
& ACC-B & Kappa & F1-W
& ACC-B & Kappa & F1-W \\
\midrule

S
& 0.5635 {\raisebox{0.25ex}{\fontsize{5}{5}\selectfont\textpm\hspace{0.08em}0.0031}}
& 0.5088 {\raisebox{0.25ex}{\fontsize{5}{5}\selectfont\textpm\hspace{0.08em}0.0050}}
& 0.5710 {\raisebox{0.25ex}{\fontsize{5}{5}\selectfont\textpm\hspace{0.08em}0.0049}}
& 0.5527 {\raisebox{0.25ex}{\fontsize{5}{5}\selectfont\textpm\hspace{0.08em}0.0218}}
& 0.4036 {\raisebox{0.25ex}{\fontsize{5}{5}\selectfont\textpm\hspace{0.08em}0.0290}}
& 0.5504 {\raisebox{0.25ex}{\fontsize{5}{5}\selectfont\textpm\hspace{0.08em}0.0236}} \\

\cellcolor{cyan!15}S + A
& \cellcolor{cyan!15}0.5937 {\raisebox{0.25ex}{\fontsize{5}{5}\selectfont\textpm\hspace{0.08em}0.0087}}
& \cellcolor{cyan!15}0.5418 {\raisebox{0.25ex}{\fontsize{5}{5}\selectfont\textpm\hspace{0.08em}0.0092}}
& \cellcolor{cyan!15}0.6023 {\raisebox{0.25ex}{\fontsize{5}{5}\selectfont\textpm\hspace{0.08em}0.0061}}
& 0.5499 {\raisebox{0.25ex}{\fontsize{5}{5}\selectfont\textpm\hspace{0.08em}0.0044}}
& 0.3999 {\raisebox{0.25ex}{\fontsize{5}{5}\selectfont\textpm\hspace{0.08em}0.0059}}
& 0.5469 {\raisebox{0.25ex}{\fontsize{5}{5}\selectfont\textpm\hspace{0.08em}0.0060}} \\

S + M
& 0.5828 {\raisebox{0.25ex}{\fontsize{5}{5}\selectfont\textpm\hspace{0.08em}0.0142}}
& 0.5296 {\raisebox{0.25ex}{\fontsize{5}{5}\selectfont\textpm\hspace{0.08em}0.0143}}
& 0.5904 {\raisebox{0.25ex}{\fontsize{5}{5}\selectfont\textpm\hspace{0.08em}0.0144}}
& \cellcolor{cyan!15}0.5674 {\raisebox{0.25ex}{\fontsize{5}{5}\selectfont\textpm\hspace{0.08em}0.0148}}
& \cellcolor{cyan!15}0.4232 {\raisebox{0.25ex}{\fontsize{5}{5}\selectfont\textpm\hspace{0.08em}0.0198}}
& \cellcolor{cyan!15}0.5653 {\raisebox{0.25ex}{\fontsize{5}{5}\selectfont\textpm\hspace{0.08em}0.0169}} \\

S + O
& 0.5821 {\raisebox{0.25ex}{\fontsize{5}{5}\selectfont\textpm\hspace{0.08em}0.0144}}
& 0.5292 {\raisebox{0.25ex}{\fontsize{5}{5}\selectfont\textpm\hspace{0.08em}0.0150}}
& 0.5881 {\raisebox{0.25ex}{\fontsize{5}{5}\selectfont\textpm\hspace{0.08em}0.0133}}
& 0.5525 {\raisebox{0.25ex}{\fontsize{5}{5}\selectfont\textpm\hspace{0.08em}0.0173}}
& 0.4033 {\raisebox{0.25ex}{\fontsize{5}{5}\selectfont\textpm\hspace{0.08em}0.0231}}
& 0.5491 {\raisebox{0.25ex}{\fontsize{5}{5}\selectfont\textpm\hspace{0.08em}0.0192}} \\

S + A + M
& 0.5988 {\raisebox{0.25ex}{\fontsize{5}{5}\selectfont\textpm\hspace{0.08em}0.0057}}
& 0.5477 {\raisebox{0.25ex}{\fontsize{5}{5}\selectfont\textpm\hspace{0.08em}0.0058}}
& 0.6098 {\raisebox{0.25ex}{\fontsize{5}{5}\selectfont\textpm\hspace{0.08em}0.0043}}
& 0.5797 {\raisebox{0.25ex}{\fontsize{5}{5}\selectfont\textpm\hspace{0.08em}0.0229}}
& 0.4396 {\raisebox{0.25ex}{\fontsize{5}{5}\selectfont\textpm\hspace{0.08em}0.0305}}
& 0.5764 {\raisebox{0.25ex}{\fontsize{5}{5}\selectfont\textpm\hspace{0.08em}0.0245}} \\

S + A + M + O
& 0.5824 {\raisebox{0.25ex}{\fontsize{5}{5}\selectfont\textpm\hspace{0.08em}0.0050}}
& 0.5272 {\raisebox{0.25ex}{\fontsize{5}{5}\selectfont\textpm\hspace{0.08em}0.0054}}
& 0.5841 {\raisebox{0.25ex}{\fontsize{5}{5}\selectfont\textpm\hspace{0.08em}0.0056}}
& 0.5660 {\raisebox{0.25ex}{\fontsize{5}{5}\selectfont\textpm\hspace{0.08em}0.0067}}
& 0.4213 {\raisebox{0.25ex}{\fontsize{5}{5}\selectfont\textpm\hspace{0.08em}0.0089}}
& 0.5631 {\raisebox{0.25ex}{\fontsize{5}{5}\selectfont\textpm\hspace{0.08em}0.0080}} \\
\bottomrule
\end{tabular}

\begin{tablenotes}
\tiny
\item Note: \cellcolor{cyan!15}{Cyan highlight} marks configurations used in BrainPro.
\end{tablenotes}

\end{threeparttable}}
}
\end{table*}

\subsection{Channel-Drop Analysis}
\paragraph{BrainPro exhibits smaller performance degradation under missing-channel conditions.}
Figure~\ref{fig:channel_drop} evaluates decoding performance on FACED under progressively increasing channel-drop rates during inference, where a random subset of channels is removed at each drop level. Detailed results are in Table~\ref{tab:drop_channel_details}. This provides an assessment of how models behave when electrode coverage is incomplete. All metrics decrease as more channels are removed, and the differences between models become more pronounced at higher drop ratios. BrainPro consistently maintains higher performance than LaBraM and CBraMod across the entire drop spectrum, and its total ACC-B drop at 50\% removal is the smallest among all models shown in Figure~\ref{fig:channel_drop} (d). These results provide empirical evidence that the structured retrieval mechanism contributes to more stable downstream performance when electrodes are missing, supporting the intended behavior of BrainPro under montage variability.

\subsection{Encoder-Combination Analysis}

Table~\ref{tab:encoder_config} reports decoding performance on FACED and BCI-IV-2A under different encoder combinations. 
Task-relevant process branches improve over the shared-only encoder: $\mathrm{S{+}A}$ performs best among single-branch variants on FACED, while $\mathrm{S{+}M}$ performs best on BCI-IV-2A. 
This suggests that the affective and motor-related encoders learn task-aligned mental-process-associated representations. 
Combining affective and motor branches ($\mathrm{S{+}A{+}M}$) can further improve performance, indicating complementary EEG information across branches. 
However, adding the auxiliary residual branch to all encoders does not always help, suggesting that this branch is a pragmatic auxiliary pathway instead of a coherent mental-process-associated encoder. 
An extended analysis further shows that the gains are not explained by MLP parameter changes alone (Appendix~\ref{app:improvement_is_not_MLP}).

\section{Conclusion}
\label{sec:conclusion}

We introduced BrainPro, a process-conditioned pretraining framework with topographic retrieval for large EEG models. 
BrainPro couples topographic retrieval with shared and process-associated encoders, an auxiliary residual branch, and process-guided region-weighted reconstruction to learn montage-consistent and process-sensitive EEG representations. 
Across nine public BCI benchmarks, BrainPro improves over evaluated baselines under the reported protocols. 
Spatial-filter visualizations, channel-drop tests, encoder-combination studies, and ablations suggest that the gains are associated with montage-consistent spatial modeling and process-conditioned specialization. 
BrainPro provides a practical framework for using coarse mental-process cues and topographic priors in large-scale EEG pretraining.

\begin{ack}
This research is supported by the Ministry of Education, Singapore, under its Academic Research Fund Tier 2 (Grant No. MOE-T2EP20124-0001)
\end{ack}

\bibliographystyle{unsrtnat}
\bibliography{mybib}


\appendix
\newpage

\etocdepthtag.toc{appendix}
\clearpage
\begingroup
\setlength{\parskip}{0pt}
\etocsetnexttocdepth{subsection}
\etocsettagdepth{main}{none}
\etocsettagdepth{appendix}{subsection}
\etocsettocstyle{\section*{Appendix Contents}}{}
\makeatletter
\renewcommand*{\l@section}[2]{%
  \addvspace{3pt}%
  \@dottedtocline{1}{0em}{2.2em}
    {\bfseries #1}{\bfseries #2}%
}
\makeatother

\tableofcontents
\endgroup
\clearpage

\section{More details about the temporal encoder in BrainPro}
\label{app:more_details_tcnn}
Following~\cite{jiang2024large}, a temporal CNN, $\mathcal{F}_{temp}$, extracts temporal dynamics along the time axis from the input EEG, $X\in \mathbb{R}^{C \times T}$, independently for each channel. Each convolutional block consists of a 1D convolution followed by Group Normalization (GroupNorm), which normalizes activations within groups of channels to stabilize training with small batch sizes, and the Gaussian Error Linear Unit (GELU), a smooth nonlinear activation that weights inputs by their Gaussian cumulative distribution for improved expressivity. It can be formulated as
\begin{align}
\mathbf{Z}^{(\ell)} 
    &= \mathcal{F}_{temp}(\mathbf{Z}^{(\ell-1)}) 
     = \mathrm{GELU}\!\left(
        \mathrm{GroupNorm}\!\left(
        \mathrm{Conv1D}^{(\ell)}(\mathbf{Z}^{(\ell-1)})
        \right)\right), \quad \ell = 1,\dots,L_T.
\end{align}

where $\mathbf{Z}^{(0)} = \mathbf{X}$ and each $\mathrm{Conv1D}^{(\ell)}$ applies $K_T$ filters (kernel size $k_\ell$, stride $s_\ell$) along time. Padding is applied to keep the output size the same. The output
$\mathbf{H}_\mathrm{temp} = \mathbf{Z}^{(L_T)} \in \mathbb{R}^{K_T \times C \times T}$
preserves temporal resolution ($T$) while expanding the per-channel feature depth to $K_T$.

\begin{table}[htp]
\centering
\caption{Overview of downstream BCI tasks and datasets.}
\label{tab:bci_tasks}
\resizebox{\textwidth}{!}{
\begin{threeparttable}
\begin{tabular}{l l c c c c c}
\toprule
\textbf{BCI Task} & \textbf{Dataset} & \textbf{$f_{s}$} & \textbf{\# Ch.} & \textbf{T} & \textbf{\# Samples} & \textbf{Label} \\
\midrule
I.Emotion Recognition & FACED        & 250 Hz  & 32 & 10 s & 10{,}332  & 9-class \\
 & SEED-V       & 1000 Hz & 62 & 1 s  & 117{,}744 & 5-class \\ 
 & SEED-VII     & 1000 Hz & 62 & 1 s  & 281{,}679 & 7-class \\ 
II.Motor Imagery & BCI-IV-2A    & 250 Hz  & 22 & 4 s & 5{,}088   & 4-class \\
 & SHU-MI       & 250 Hz  & 32 & 4 s & 11{,}988  & 2-class \\ 
III.Imagined Speech 
 & BCIC2020-3   & 256 Hz  & 64 & 4 s & 5{,}250   & 5-class \\
IV.Mental Disorder Diagnosis 
 & Mumtaz2016   & 256 Hz  & 19 & 4 s & 8{,}860   & 2-class \\
V.Mental Stress Detection 
 & MentalArithmetic & 500 Hz  & 20 & 4 s & 1{,}080 & 2-class \\ 
VI.Mental Attention Detection 
 & ATTEN        & 500 Hz  & 28 & 4 s & 4{,}680   & 2-class \\ 
\bottomrule
\end{tabular}

\begin{tablenotes}
\scriptsize
\item \textbf{Note:} All datasets are preprocessed into fixed-length EEG segments following their official protocols.
\end{tablenotes}
\end{threeparttable}
}
\end{table}

\section{More Details on Tokenization and the Transformer Used in This Paper}
\label{app:more_details_transformer}

Given the spatial representation from the retrieval-based spatial encoder, $\mathbf{H}_\mathrm{spatial}$, we divide it into $N_p$ (possibly strided with the step being $s_{P}$) temporal patches of length $P$:
\begin{equation}
  \mathbf{Z}_\mathrm{patch} \in \mathbb{R}^{N_p \times K_{T} \times K_\mathrm{total} \times P}, \quad
  N_p = \Big\lfloor \frac{T - P}{s_P} \Big\rfloor + 1.
\end{equation}
Each patch is flattened and projected into a $d$-dimensional token using a learnable embedding matrix $\mathbf{E}\in\mathbb{R}^{d \times (K_T K_\mathrm{total} P)}$, followed by adding a positional embedding $\mathbf{p}_i\in\mathbb{R}^d$ onto the flattened and linearly projected tokens:
\begin{equation}
    \mathbf{z}_i = \mathbf{E}\,\mathrm{Flatten}(\mathbf{Z}_\mathrm{patch}[i]) + \mathbf{p}_i,
    \quad i=1,\dots,N_p.
\end{equation}
This yields a sequence of tokens 
$\mathbf{Z}=\{\mathbf{z}_i\}_{i=1}^{N_p} \in \mathbb{R}^{N_p \times d}$, which serves as the input to the Transformer layers for modeling temporal contextual information.

To capture long-range spatiotemporal dependencies, each encoder applies $L$ Transformer blocks. Each block follows a pre-norm architecture consisting of multi-head self-attention (MSA) and a position-wise multilayer perceptron (MLP):
\begin{align}
    \mathbf{Z}'^{(\ell)} &= \mathbf{Z}^{(\ell)}
    + \mathrm{MSA}\!\left(\mathrm{LN}(\mathbf{Z}^{(\ell)})\right), \\
    \mathbf{Z}^{(\ell+1)} &= \mathbf{Z}'^{(\ell)}
    + \mathrm{MLP}\!\left(\mathrm{LN}(\mathbf{Z}'^{(\ell)})\right),
    \quad \ell = 0,\dots,L-1,
\end{align}
where $\mathbf{Z}^{(0)}=\mathbf{Z}$ denotes the input token embeddings and $\mathrm{LN}$ is layer normalization.

The MSA module models pairwise interactions among all patches by projecting $\mathbf{Z}^{(\ell)}$ into queries ($Q$), keys ($K$), and values ($V$), and computing
\begin{equation}
    \mathrm{Attention}(Q,K,V) =
    \mathrm{softmax}\!\left(\frac{QK^\top}{\sqrt{d_k}}\right)V,
\end{equation}
where $d_k$ is the key dimension. Multiple attention heads operate in parallel and are concatenated to form richer contextual representations.

The MLP module consists of two fully connected layers with hidden dimension $d_\mathrm{ff}$ and a nonlinear GELU activation, applied independently to each token. Residual connections around both the MSA and MLP components stabilize training.

After $L$ blocks, we obtain the final token sequence
$\mathbf{Z}^{(L)} \in \mathbb{R}^{N_p \times d}$.

\section{Downstream Adaptation}
\label{app:flexible_configuration}

Downstream adaptation follows a fixed branch-activation rule in all main experiments. 
The shared encoder is always activated. 
For affective downstream tasks, we activate the affective branch; for motor-imagery tasks, we activate the motor-related branch; for imagined-speech tasks, we activate the auxiliary residual branch. 
Only the activated encoders are fine-tuned, and their outputs are concatenated and fed to an MLP classifier~\citep{wang2025cbramod}. For future applications where the process category is ambiguous or mixed, one possible strategy is to select from a fixed candidate set $\{\mathrm{S}, \mathrm{S{+}A}, \mathrm{S{+}M}, \mathrm{S{+}O}, \mathrm{S{+}A{+}M}\}$ using the validation split and choose the smallest near-best configuration.

\begin{algorithm}[!ht]
\caption{BrainPro Pretraining Pipeline}
\label{alg:brainpro}
\begin{algorithmic}[1]
\State \textbf{Input:} EEG segment
$\mathbf{X}\in\mathbb{R}^{C\times T}$,
montage mapping, mental-process cue $y_{\mathrm{proc}}$
\State \textbf{Params:} Shared encoder $\mathcal{E}_{\mathrm{S}}$,
mental-process encoders
$\{\mathcal{E}_k\}_{k\in\{\mathrm{A,M,O}\}}$,
mental-process decoders
$\{\mathcal{D}_k\}_{k\in\{\mathrm{A,M,O}\}}$,
spatial filter banks $\mathbf{W}_{\mathrm{C}},\mathbf{W}_{\mathrm{R}}$,
channel priors $w^{(y_{\mathrm{proc}})}$,
learnable input mask, mask ratio $\rho$
\Statex

\State \textbf{Step 1: Mask input}
\State Replace a randomly selected fraction $\rho$ of
patches in $\mathbf{X}$ with learnable
masks to obtain $\tilde{\mathbf{X}}$

\State \textbf{Step 2: Representation extraction for separation}
\For{each encoder
$\mathcal{E}_j\in
\{\mathcal{E}_{\mathrm{S}},\mathcal{E}_{\mathrm{A}},
\mathcal{E}_{\mathrm{M}},\mathcal{E}_{\mathrm{O}}\}$}
    \State Temporal encoding:
    $\mathbf{H}_{\mathrm{temp}}
    \gets \mathcal{F}_{\mathrm{temp}}(\tilde{\mathbf{X}})$
    \State Spatial retrieval:
    $\mathbf{H}_{\mathrm{spatial}}
    \gets \mathcal{F}_{\mathrm{spatial}}
    (\mathbf{H}_{\mathrm{temp}},
    \mathbf{W}_{\mathrm{C}},\mathbf{W}_{\mathrm{R}})$
    \State Patchify \& embed:
    $\mathbf{Z}
    \gets \mathrm{PatchifyEmbed}(\mathbf{H}_{\mathrm{spatial}})$
    \State Transformer encoding:
    $\mathbf{Z}_j \gets \mathrm{Transformer}(\mathbf{Z})$
    \If{$j\neq y_{\mathrm{proc}}$ and $j\neq\mathrm{S}$}
        \State Detach $\mathbf{Z}_j$
        (inactive encoders, used only for separation loss)
    \EndIf
\EndFor

\State \textbf{Step 3: Reconstruction}
\State Concatenate outputs:
$\mathbf{Z}_{y_{\mathrm{proc}}}^{\mathrm{comb}}
\gets \mathrm{Concat}
(\mathbf{Z}_{\mathrm{S}},\mathbf{Z}_{y_{\mathrm{proc}}})$
\State Reconstruct EEG:
$\hat{\mathbf{X}}
\gets \mathcal{D}_{y_{\mathrm{proc}}}
(\mathbf{Z}_{y_{\mathrm{proc}}}^{\mathrm{comb}})$

\State \textbf{Step 4: Loss computation}
\State Compute $\mathcal{L}_{\mathrm{rec}}$
against the unmasked target on retained
channels using Eq.~\eqref{eq:rec}
\State Mental-process separation loss:
$\mathcal{L}_{\mathrm{sep}}
\gets \text{margin-based cosine loss}$
\State Total loss:
$\mathcal{L}_{\mathrm{BrainPro}}
\gets \mathcal{L}_{\mathrm{rec}}+\mathcal{L}_{\mathrm{sep}}$

\State \textbf{Step 5: Parameter update}
\State Update $\mathcal{E}_{\mathrm{S}}$,
$\mathcal{E}_{y_{\mathrm{proc}}}$,
$\mathcal{D}_{y_{\mathrm{proc}}}$, and learnable mask
parameters using $\mathcal{L}_{\mathrm{BrainPro}}$
\end{algorithmic}
\end{algorithm}

\section{Theoretical Justification}
\label{app:Justify_spatial_learning}
In this section, we provide a theoretical analysis showing that
(1) retrieval-based spatial filtering enables \emph{explicit} encoding of spatial information across heterogeneous EEG channel montages, whereas
(2) self-attention yields \emph{implicit} spatial representations that do not correspond to stable channel identities.

\subsection{Preliminaries}
Let an EEG input be represented as $X \in \mathbb{R}^{C \times T}, \qquad x(t) = X_{:,t}$, where $C$ is the number of channels and $T$ is the number of time samples.
Each channel $c$ corresponds to a fixed scalp location $p_c \in \mathbb{R}^3$.

A spatial filter is parameterized by $w = (w_1, \dots, w_C)^\top \in \mathbb{R}^{C}$,
and produces the spatially filtered signal 
\[
y(t) = w^\top x(t) = \sum_{c=1}^C w_c X_{c,t}.
\]
Self-attention operates on channel embeddings $H \in \mathbb{R}^{C \times d}$ via $Q = HW_Q,\quad K = HW_K,\quad V = HW_V$, and produces
\[
    Z = \mathrm{softmax}\!\left( \frac{QK^\top}{\sqrt{d_k}} \right)V,
\]
where $W_Q, W_K, W_V$ are shared projection matrices.

\subsection{Explicit Spatial Encoding via Spatial Filters}

\begin{proposition}[Explicit Spatial Encoding]
If a model contains a learnable parameter $w_c$ that always multiplies the same channel $c$, then the model explicitly encodes spatial information because each parameter corresponds to a fixed scalp location $p_c$.
\end{proposition}

\begin{proof}
The mapping channel $\to$ scalp location is fixed: $c \mapsto p_c$.
Since $w_c$ always multiplies $X_{c,t}$, the contribution of scalp location $p_c$ to the output $y(t)$ is governed by a dedicated parameter.
Thus spatial information is tied directly to parameters, yielding an explicit spatial representation.
\end{proof}

\subsection{Lack of Explicit Spatial Encoding in Self-Attention}

\begin{proposition}[Self-Attention Is Not Spatially Explicit]
Self-attention cannot explicitly encode spatial structure because none of its learnable parameters are tied to channel identity or scalp location. The attention weights $a_{ij}$ depend on input content rather than spatial location.
\end{proposition}

\begin{proof}
Self-attention uses shared parameters $W_Q, W_K, W_V$ for all channels.
For channels $i$ and $j$, the interaction weight
\[
    a_{ij} = \frac{\exp(q_i k_j^\top)}{\sum_{j'=1}^C \exp(q_i k_{j'}^\top)}
\]
depends on the embeddings $h_i$ and $h_j$, which depend on the input signal and not on fixed locations $p_i$ or $p_j$.
Thus $a_{ij}$ does not represent a stable spatial relationship across datasets or subjects.
Since no parameter is associated with a fixed channel index, self-attention lacks explicit spatial semantics.
\end{proof}

\subsection{Retrieval-Based Alignment Restores Explicit Spatial Structure}

Suppose dataset $A$ has channels $\{1,\dots,C_A\}$ with locations $\{p^{(A)}_c\}$,
and dataset $B$ has channels $\{1,\dots,C_B\}$ with locations $\{p^{(B)}_c\}$.
A retrieval function
\[
    \phi : \{1,\dots,C_B\} \rightarrow \{1,\dots,C_A\}
\]
selects, for each channel $c$ in dataset $B$, the most similar channel in dataset $A$:
\[
    \phi(c) = \arg\min_{j \in \{1,\dots,C_A\}} d(p^{(B)}_c, p^{(A)}_j),
\]
for distance measure $d(\cdot,\cdot)$.

The spatial filter learned on dataset $A$, $w^{(A)} \in \mathbb{R}^{C_A}$, is transferred to dataset $B$ via
\[
    w^{(B)}_c = w^{(A)}_{\phi(c)}.
\]

\begin{proposition}[Retrieval Restores Explicit Spatial Encoding]
If $\phi$ maps channels in dataset $B$ to their spatially closest counterparts in dataset $A$, then the transferred filter $w^{(B)}$ preserves explicit spatial semantics across datasets.
\end{proposition}

\begin{proof}
The retrieval function $\phi$ associates each channel $c$ in dataset $B$ with a channel $\phi(c)$ in dataset $A$ that corresponds to the closest spatial location.
Thus $w^{(B)}_c$ inherits the spatial meaning of $w^{(A)}_{\phi(c)}$.
By Proposition~1, $w^{(A)}_{\phi(c)}$ explicitly encodes the spatial contribution of a fixed location.
Therefore, $w^{(B)}_c$ is tied to a stable spatial meaning, yielding an explicit spatial representation across heterogeneous channel configurations.
\end{proof}

\subsection{Conclusion}
Spatial filters produce explicit spatial representations because each parameter corresponds to a fixed topographic location. Self-attention produces only implicit spatial structure because its interaction weights depend on sample-dependent feature similarity. Retrieval-based alignment allows spatial filters to retain explicit spatial semantics even when aggregating data across heterogeneous EEG montages.

\section{Related Work}
\subsection{Previous Methods for EEG Decoding}
Traditional EEG decoding relied on handcrafted features and classical classifiers such as common spatial patterns (CSP), linear discriminant analysis (LDA), and support vector machines (SVMs), which required extensive domain expertise and were often task-specific~\citep{Lotte_2007}. With the rise of deep learning, neural architectures emerged to learn spatiotemporal EEG representations directly from raw signals~\citep{https://doi.org/10.1002/hbm.23730}. CNN-based models captured spatial correlations between electrodes~\citep{Lawhern_2018,9762054}, while recurrent networks such as LSTMs modeled temporal dynamics~\citep{8496885,MICHIELLI201971}. Hybrid CNN-LSTM frameworks~\citep{LI2022103342} and attention-based architectures~\citep{9991178,10763464} further advanced decoding performance, and graph neural networks were introduced to exploit functional brain connectivity~\citep{8320798,9091308,10025569}. These approaches achieved strong results in tasks including motor imagery classification, seizure detection, emotion recognition, and sleep staging. However, they typically relied on supervised learning tailored to specific datasets, limiting scalability in the face of EEG’s heterogeneous channel configurations, variable sampling rates, and low signal-to-noise ratios.

\subsection{EEG Foundation Models}
Inspired by the success of large language and vision models, recent studies have proposed \textit{EEG foundation models} (EEG-FMs) pre-trained on large unlabeled corpora with self-supervised objectives~\citep{zhou2025brainfoundationmodelssurvey}. These models aim to learn universal neural representations that can be adapted to diverse downstream BCI tasks. Brant~\citep{NEURIPS2023_535915d2} and NeuroGPT~\citep{10635453}, explored masked modeling and contrastive learning strategies to enhance cross-task generalization. BIOT~\citep{NEURIPS2023_f6b30f3e} introduced a unified encoder that tokenizes heterogeneous biosignals into a sentence-like representation to support cross-dataset pretraining. LaBraM~\citep{jiang2024large} introduced a neural tokenizer and masked EEG modeling trained on 2,500 hours of data, achieving state-of-the-art results in abnormal detection, event classification, and emotion recognition. EEGPT~\citep{NEURIPS2024_4540d267} developed a 10-million-parameter pretrained transformer with a dual mask-based self-supervised framework and spatio-temporal representation alignment to improve robustness under low SNR. CBraMod~\citep{wang2025cbramod} employed a criss-cross transformer with asymmetric positional encoding to separately model spatial and temporal dependencies, achieving strong generalizability across diverse BCI datasets. 

Despite recent progress, EEG-FMs still face important limitations. Many existing models capture spatial interactions only implicitly or assume fixed channel configurations, limiting adaptability across heterogeneous montages. They also typically learn a single general-purpose representation, without explicitly modeling shared EEG structure and variation associated with different mental processes, such as affective and motor-related activity. These constraints motivate models that explicitly encode spatial dependencies while learning both shared and process-associated representations.

\section{More details for the Brain Region Definition}
\label{app:brian-region}
\begin{figure}[!ht]
    \centering
    \includegraphics[width=0.4\textwidth]{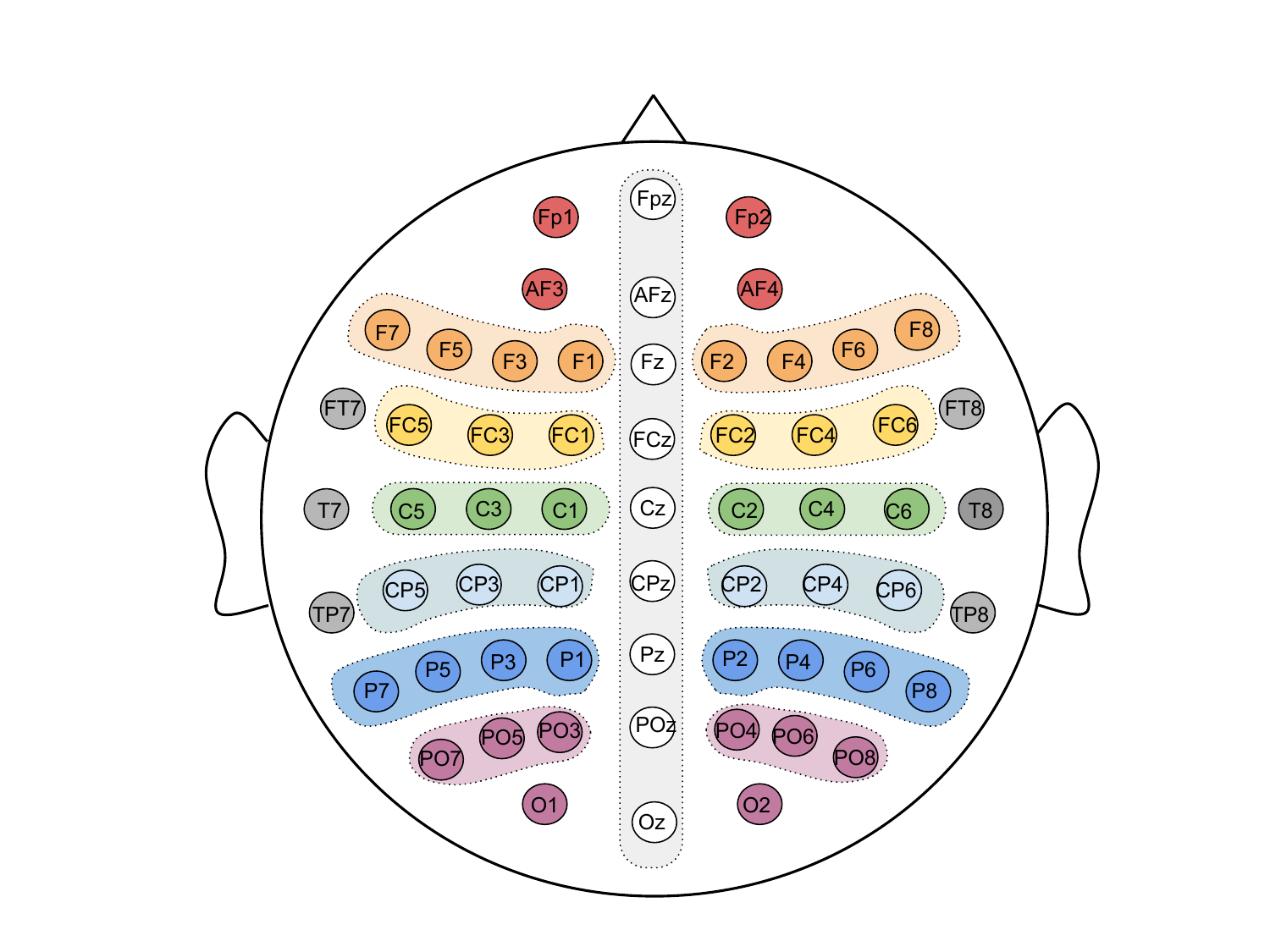}
    \vspace{-15pt}
    \caption{The brain region definition used in spatial learning block.}
    \label{fig:brain_region}
\end{figure}

EEG captures activity from multiple functional areas of the brain. Treating electrodes as isolated nodes or relying only on global connections overlooks localized cooperative activity. LGGNet~\citep{10025569} addresses this by defining local brain regions and modeling both within-area activity (local graphs) and between-area interactions (global graphs), thereby capturing the brain’s hierarchical organization. The hemisphere LGG further extends this structure by introducing symmetric subgraphs across hemispheres, motivated by evidence of inter-hemispheric asymmetries in cognitive and emotional processes. This enables more effective modeling of bilateral patterns and asymmetries, which is beneficial for tasks such as emotion recognition and preference prediction. Following LGGNet, BrainPro adopts the hemisphere-based LGG region definition with minor adjustments: Fp1/Fp2 are separated from the AF regions, O1/O2 from the PO regions, and the FT, T, and TP regions are split into distinct areas. The brain regions are shown in Figure~\ref{fig:brain_region}.

\section{More Details For Pretraining}
\subsection{Main Modules}
The main modules in BrainPro framework are show in Table~\ref{tab:module_overview}. The process-cue column indicates whether the operation depends on the sample's process label. Retrieval indices depend on the montage, while spatial filter parameters are encoder-specific.

\begin{table}[!ht]
\centering
\small
\caption{Overview of the main modules in BrainPro.}
\label{tab:module_overview}
\setlength{\tabcolsep}{4pt}
\renewcommand{\arraystretch}{1.15}
\begin{tabularx}{\linewidth}{
    @{}
    >{\raggedright\arraybackslash}p{0.15\linewidth}
    >{\raggedright\arraybackslash}p{0.17\linewidth}
    >{\raggedright\arraybackslash}X
    >{\raggedright\arraybackslash}p{0.10\linewidth}
    >{\raggedright\arraybackslash}p{0.21\linewidth}
    @{}
}
\toprule
\textbf{Module}
& \textbf{Input}
& \textbf{Operation}
& \textbf{Process cue}
& \textbf{Purpose} \\
\midrule

Topographic mapping
& Dataset montage
& Map supported channels to the 60-channel template
& $\times$
& Establish a common spatial basis \\

\addlinespace
Channel and region retrieval
& Temporal features and template indices
& Retrieve spatial filters from each encoder using
montage-defined indices
& $\times$
& Associate spatial parameters with scalp locations \\

\addlinespace
Branch routing
& EEG segment and process label
& Route gradients to the shared and assigned process
encoders
& $\checkmark$
& Learn shared and process-associated representations \\

\addlinespace
Separation loss
& Branch representations
& Apply margin-based cosine regularization with
detached inactive outputs
& $\checkmark$
& Discourage excessive similarity between branches \\

\addlinespace
Reconstruction
& Shared and active branch outputs
& Decode the unmasked EEG target in template space
using the active decoder
& $\checkmark$
& Provide a masked reconstruction objective \\

\addlinespace
Region weighting
& Reconstruction errors, process priors, and epoch
& Apply epoch-dependent channel weights,
with uniform weights for the auxiliary group
& $\checkmark$
& Emphasize coarse process-relevant regions \\

\bottomrule
\end{tabularx}
\end{table}
\subsection{Pretraining Datasets}
\label{app:pretrainingdatasets}
A detailed description of the pretraining datasets for BrainPro is provided here. Most of these datasets overlap with those used for LaBraM, but we excluded some to ensure that the pretraining data represent meaningful mental processes. In addition, we incorporated two extra motor-related datasets to provide sufficient data for training each brain process encoder. In total, the datasets comprise around \textbf{2400 hours}.

\begin{itemize}[leftmargin=10pt]
    \item \textbf{Emobrain} \citep{savran2006emotion}: Multimodal emotion dataset with EEG (64 channels, 1024 Hz) and fNIRS from 16 subjects, collected via Biosemi Active 2. Emotions elicited with a subset of IAPS stimuli. (4.94 h)
    \item \textbf{Grasp and Lift EEG Challenge} \citep{luciw2014multi}: EEG from 12 subjects (32 channels, 500 Hz) performing grasp-and-lift trials with a BrainAmp EEG amplifier. (11.72 h)
    \item \textbf{EEG Motor Movement/Imagery} \citep{schalk2004bci2000}: Recordings from 109 volunteers (64 channels, 160 Hz) performing baseline (eyes open/closed), movement, and imagery (both fists/feet) tasks using the BCI2000 system. (47.3 h)
    \item \textbf{KU} \citep{10.1093/gigascience/giz002}: Data from 54 subjects conducting motor imagery tasks. It was recorded with 62 channels, 1000 Hz. (24.00 h)
    \item \textbf{HGD} \citep{https://doi.org/10.1002/hbm.23730}: Data from over 14 subjects, recorded 128 channels, 500 Hz. (7.49 h)
    \item \textbf{Raw EEG Data} \citep{T8/SS2NHB_2020}: EEG (64 channels, 256 Hz) collected during information-integration and multidimensional rule-based categorization tasks. (34.35 h)
    \item \textbf{Resting State EEG Data} \citep{trujillo2017effect}: EEG from 22 subjects during 8-minute resting (4 minutes eyes closed, 4 minutes eyes open), recorded with 64 channels at 256 Hz using BioSemi active Ag/AgCl electrodes. (3.04 h)
    \item \textbf{SEED Series} \citep{zheng2015investigating,8283814,liu2022identifying}: Emotional EEG datasets including SEED (15 subjects), SEED-IV (15 subjects), SEED-GER (8 subjects), and SEED-FRA (8 subjects). Signals recorded at 1000 Hz from 62 channels with the ESI NeuroScan System while subjects viewed videos. (166.75 h)
    \item \textbf{SPIS Resting State} \citep{torkamani2020prediction}: EEG from 10 subjects (64 channels, 2048 Hz) with 2.5-minute eyes-open/closed sessions before a 105-minute sustained attention task. (0.83 h)
    \item \textbf{TUEP} \citep{veloso2017big}: Subset of TUEG with 100 epilepsy and 100 control subjects, verified by a neurologist; EEG recorded with 19-23 channels at 256 Hz. (591.22 h)
    \item \textbf{TUSZ} \citep{shah2018temple}: Seizure-annotated EEG corpus (19-23 channels, 256 Hz) including onset, offset, channel, and seizure type information. (1138.53 h)
    \item \textbf{TUSL} \citep{von2017electroencephalographic}: TUEG subset with slowing event annotations (23 channels, 256 Hz), used in seizure detection error analysis. (20.59 h)
    \item \textbf{SHJT EEG Data} \citep{10.1145/3581783.3613797,9669637,9871630,9441086,9206957}: Data from over 140 subjects, recorded with the ESI NeuroScan System (62 channels, 1000 Hz). (342.23 h)
   
\end{itemize}

\subsection{Experiment Settings}
\label{app:experiment_settings_pretraining}
We describe the experimental setting for pretraining as follows. All EEG recordings are segmented into 10-second clips and resampled to 200 Hz. We predefine a 60-channel montage based on the SEED-series datasets (excluding two reference channels) as the reference configuration. Channels from different pretraining datasets are mapped to this predefined montage according to their spatial distance on the scalp. Channels that are absent from the predefined set or cannot be reliably mapped are discarded. To enable process-associated representation learning, we categorize the datasets into three groups, \emph{affect}, \emph{motor} and \emph{other}, for brain-process decoupling training (see Table~\ref{tab:pre-train_dataset_category} for details). To make the pretraining stable, we discard the noisy segments that have large absolute values resulting around 785,000 clear samples, around 2,180 hours. These samples from different datasets are shuffled before being fed into the model. We pre-trained BrainPro for 30 epochs. To improve training efficiency, samples from the same dataset are grouped within each training batch, ensuring that the spatial configuration requires only one spatial filter retrieval. During training, the gradients of samples belonging to different brain processes are passed to their corresponding encoders and decoders, while the shared encoder processes all samples to capture universal EEG representations.

\begin{table}[!ht]
\centering
\caption{Categories of the pretraining datasets.}
\label{tab:pre-train_dataset_category}
\begin{threeparttable}
\resizebox{\textwidth}{!}{
\begin{tabularx}{\linewidth}{lY}
\toprule
\textbf{Brain Process Category} & \textbf{Datasets} \\
\midrule
Affect & Emobrain, SEED, SEED-IV, SEED-GER, SEED-FRA, and SHJT EEG Data \\
Motor & Grasp and Lift EEG Challenge, EEG Motor Movement/Imagery, KU, and HGD \\
Others & Raw EEG Data, Resting State EEG Data, SPIS Resting State, TUEP, TUSZ, and TUSL \\
\bottomrule
\end{tabularx}}
\end{threeparttable}
\end{table}

\subsection{Hyper-parameters}
\label{app:pre_training_hyper_parameters}
The pretraining of BrainPro is in an end-to-end manner, and the hyper-parameter settings are shown in Table~\ref{tab:pretraining_hyper_parameters}.

\label{app:details_pre_training}
\begin{table}[!ht]
\centering
\caption{Hyperparameters for pretraining.}
\label{tab:pretraining_hyper_parameters}
\begin{threeparttable}
\begin{tabular}{@{}lll@{}}
\toprule
\textbf{Type} & \textbf{Factor} & \textbf{Value} \\
\midrule
\multirow{5}{*}{Temporal Encoder} 
  & Input channels & \{1, 32, 32\} \\
  & Output channels & \{32, 32, 32\} \\
  & Kernel size & \{15, 3, 3\} \\
  & Stride & \{1, 1, 1\} \\
  & Padding & \{7, 1, 1\} \\
\midrule
\multirow{5}{*}{Spatial Encoder} 
  & Channel-wise Filter Number & 32 \\
  & Region-wise Filter Number & 32 \\
  & Total EEG Channel & 60 \\
  & Total Brain Region & 24 \\
\midrule
\multirow{3}{*}{Patch Maker} 
  & Patch Length & 20 \\
  & Patch Stride & 20 \\
  & MLP size & 64 \\
\midrule
\multirow{4}{*}{Transformer Encoder} 
  & Layers & 4 \\
  & Hidden size & 32 \\
  & MLP size & 64 \\
  & Attention heads & 32 \\
\midrule
\multirow{4}{*}{Transformer Decoder} 
  & Layers & 2 \\
  & Hidden size & 32 \\
  & MLP size & 64 \\
  & Attention heads & 32 \\
\midrule
\multirow{11}{*}{Training} 
  & Batch size & 160 (32 per GPU) \\
  & Peak learning rate & 1e-4 \\
  & Minimal learning rate & 1e-5 \\
  & Scheduler & Cosine \\
  & Optimizer & AdamW \\
  & Adam $\beta$ & (0.9, 0.98) \\
  & Weight decay & 0.05 \\
  & Total epochs & 30 \\
  & Warmup epochs & 2 \\
  & Gradient clipping & 3 \\
  & Mask ratio & 0.5 \\
\bottomrule
\end{tabular}
\end{threeparttable}
\end{table}

\subsection{Compute resources}
\label{app:compute_resources}
Pretraining was performed on a server with 5 NVIDIA A800 GPUs, approximately 512 GB RAM, and 1 TB storage. The pretraining run took approximately 3 days, corresponding to about 360 A800 GPU-hours. Downstream fine-tuning experiments were run on the same hardware and required substantially less compute than pretraining.

\section{More Details For Downstream Tasks}
\label{app:details_downstream_tasks}
\subsection{Hyper-parameters}
\label{app:downstream_hyper_parameters}
Table~\ref{tab:finetune_hyper_params_tune} lists the fine-tuned hyper-parameters for each downstream task, while the common training hyper-parameters are shown in Table~\ref{tab:finetune_hyper_params_fixed}.
\begin{table}[!ht]
\centering
\caption{Tuned hyper-parameters for downstream tasks.}
\label{tab:finetune_hyper_params_tune}
\begin{threeparttable}
\begin{tabular}{lllll}
\toprule
\textbf{Dataset} & \textbf{Learning Rate} & \textbf{Hidden Factor} & \textbf{Dropout} & \textbf{Token Selection} \\
\midrule
FACED                    & 5e-4   & 8   & 0.1   & All  \\
SEED-V                   & 3e-4   & 1   & 0.1   & All  \\
SEED-VII                 & 3e-4   & 8   & 0.1   & All  \\
SHU                      & 1e-4   & 1   & 0.1   & Aggr  \\
BCI-IV-2A                & 1e-3   & 1   & 0.2   & Aggr  \\
Mumtaz2016               & 1e-4   & 2   & 0.1   & Aggr  \\
Mental Arithmetic        & 1e-4   & 1   & 0.1   & Mean  \\ 
ATTEN                    & 1e-3   & 4   & 0.1   & Aggr  \\
BCIC2020-3               & 3e-4   & 1   & 0.1   & Mean   \\
\bottomrule
\end{tabular}
\end{threeparttable}
\end{table}

\begin{table}[!ht]
\centering
\caption{Fixed hyper-parameters for fine-tuning.}
\label{tab:finetune_hyper_params_fixed}
\begin{threeparttable}
\begin{tabular}{ll}
\toprule
\textbf{Hyper-parameters} & \textbf{Settings} \\ \midrule
Epochs & 50 (Others) / 30 (ATTEN) \\
Batch size & 64 \\
Optimizer & AdamW \\
Adam $\beta$ & (0.9, 0.999) \\
Adam $\epsilon$ & 1e-8 \\
Weight decay & 5e-2 / 1e-1 (BCIC2020-3)\\
Scheduler & CosineAnnealingLR \\
Cosine cycle epochs & 50 \\
Minimal learning rate & 1e-6 \\
Clipping gradient norm & 1 \\
Label smoothing (multi-class classification) & 0.1 \\ 
\bottomrule
\end{tabular}
\end{threeparttable}
\end{table}

\subsection{Downstream Datasets}
\label{app:downstream_dataset}
We evaluate BrainPro on nine downstream datasets covering diverse BCI applications, including emotion recognition, motor imagery, imagined speech, mental disorder diagnosis, stress detection, and attention decoding. 
These datasets differ substantially in subjects, channel configurations, sampling rates, trial durations, and class definitions, providing a comprehensive and challenging testbed for evaluating EEG foundation models. 
All EEG signals are resampled to 200 Hz, segmented into fixed-length windows according to task protocol, and the detailed training, validation, and test split is introduced in this section.

The \textbf{FACED} dataset~\citep{chen2023} is used for the task of emotion recognition. It consists of 32-channel EEG recordings sampled at 250 Hz from 123 participants, each exposed to 28 video clips designed to elicit nine distinct emotional states: amusement, inspiration, joy, tenderness, anger, fear, disgust, sadness, and neutral. Each EEG trial lasts 10 seconds and is resampled to 200 Hz, yielding a total of 10,332 clean EEG segments. Following \cite{wang2025cbramod}, subjects 1-80 are used for training, 81-100 for validation, and 101-123 for testing, ensuring subject-independent evaluation.  

The \textbf{SEED-V} dataset~\citep{liu2021comparing} is another benchmark for emotion recognition, covering five emotional categories: happy, sad, neutral, disgust, and fear. It contains 62-channel EEG recordings sampled at 1000 Hz from 16 subjects, each completing three sessions with 15 trials per session. The signals are segmented into 1-second windows and resampled to 200 Hz, resulting in 117,744 samples. Each session is evenly divided into training, validation, and testing subsets (5 trials each).

The \textbf{SEED-VII} dataset~\citep{10731546} extends the SEED series to seven emotion categories: neutral, sad, fear, disgust, happy, surprise, and anger. It contains 62-channel EEG recordings at 1000 Hz from 20 subjects, each watching 80 film clips designed to elicit emotions in four sessions. Using the same pre-processing steps as SEED-V, EEG signals are segmented into 1-second windows and resampled to 200 Hz, producing 281,679 labeled samples. As there are 20 trials in each session, they are split at the trial level into training (10), validation (5), and test subsets (5).  

The \textbf{BCI-IV-2A} dataset~\citep{katb-zv89-24} is a standard benchmark for motor imagery. It consists of EEG from 9 subjects recorded with 22 channels at 250 Hz while performing four classes of motor imagery: left hand, right hand, both feet, and tongue. Each trial lasts 4 seconds, and signals are resampled to 200 Hz. We applied a 0.1-70 Hz band-pass filter. In total, 5,088 trials are obtained. We applied subject-independent, ensuring that train, validation, and test partitions are disjoint at the subject level.  

The \textbf{SHU-MI} dataset~\citep{Ma2022} supports binary motor imagery classification. EEG was collected from 25 participants imagining either left- or right-hand movements, recorded with 32 channels at 250 Hz. The data are resampled to 200 Hz and segmented into 4-second non-overlapping windows, yielding 11,988 labeled samples. Following~\cite{wang2025cbramod}, subjects 1-15 are used for training, 16-20 for validation, and 21-25 for testing.   

The \textbf{BCIC2020-3} dataset~\citep{jeong20222020} focuses on imagined speech recognition. It includes EEG recordings from 64 channels at 256 Hz, where 15 participants imagine speaking one of five words. Each trial lasts 4 seconds and is resampled to 200 Hz, yielding 5,250 labeled samples. We use the official validation set as our test set, and the official training data are further divided into training and validation sets at a ratio of 9:1.

The \textbf{Mumtaz2016} dataset~\citep{mumtaz2016} is designed for mental disorder diagnosis, specifically distinguishing patients with major depressive disorder (MDD) from healthy controls. EEG signals were recorded from 19 electrodes at 256 Hz, preprocessed with 0.3-75 Hz bandpass and 50 Hz notch filters, and then resampled to 200 Hz. Segments of 4 seconds are extracted, producing 8,860 labeled samples. We follow~\cite{wang2025cbramod} for subject-wise splits to ensure reliable evaluation.  

The \textbf{Mental Arithmetic} dataset~\citep{data4010014} supports the task of mental stress detection. EEG was recorded from 36 subjects using 20 electrodes at 500 Hz, under two cognitive states: resting (“no stress”) and active mental arithmetic (“stress”). The data are resampled to 200 Hz, band-pass filtered (0.5-45 Hz), and segmented into 4-second windows, producing 1,080 labeled samples. We under-sample the class with more data samples to make the class balanced as~\cite{10025569}. Following~\cite{wang2025cbramod}, subjects 1-28 are used for training, 29-32 for validation, and 33-36 for testing.  

The \textbf{ATTEN} dataset is used for mental attention detection. It consists of EEG data from 26 subjects, recorded using 28 channels at 500 Hz during the Discrimination/Selection Response (DSR) task, which assesses cognitive attention. We adopt the pre-processing steps described in~\citep{10763464}. The signals are resampled to 200 Hz and segmented into 4-second non-overlapping windows, yielding 4,680 labeled samples. Subjects 1-20 are used as the training set, subjects 21-23 as the validation set, and the last 3 subjects as the test set.

\subsection{Baselines}
\label{app:baseline}
\textbf{EEGNet}~\citep{Lawhern_2018}: It is a lightweight convolutional neural network tailored for EEG-based BCI applications. By employing depthwise and separable convolutions, it efficiently extracts discriminative EEG features, enabling effective and computationally efficient EEG decoding.

\textbf{Conformer}~\citep{9991178}: It integrates CNNs with Transformers to model the spatio-temporal characteristics of EEG signals. It employs 1-D CNNs to extract local features and leverages a self-attention mechanism to capture global temporal dependencies. 

\textbf{BIOT}~\citep{NEURIPS2023_f6b30f3e}: It is a transformer-based model developed to address cross-dataset EEG classification challenges under domain shifts. It incorporates a domain-invariant attention mechanism alongside contrastive representation learning, which together improve generalization across diverse recording setups and subject groups.

\textbf{LaBraM}~\citep{jiang2024large}: It introduces a scalable transformer framework designed to learn universal EEG representations from large-scale brain signal datasets. Through pretraining on a broad range of EEG recordings, the model captures comprehensive temporal and spatial characteristics that can be effectively transferred to downstream BCI tasks. Its architecture integrates efficient self-attention operations with task-specific adapters, supporting flexible fine-tuning for varied applications.

\textbf{EEGPT}~\cite{NEURIPS2024_4540d267}: It adopts a dual self-supervised learning paradigm that combines masked autoencoding with spatio-temporal representation alignment. By emphasizing high signal-to-noise ratio (SNR) features rather than raw inputs, it improves the quality of learned representations. The model employs a hierarchical design that decouples spatial and temporal processing, thereby enhancing both computational efficiency and adaptability across BCI scenarios.

\textbf{CBraMod}~\cite{wang2025cbramod}: It is a transformer-based EEG foundation model tailored to capture the complex spatial and temporal dependencies in EEG data. It introduces a criss-cross transformer structure with parallel spatial and temporal attention modules, enabling simultaneous yet independent modeling of spatial and temporal dynamics.

\textbf{CodeBrain}~\citep{ma2026codebrain}: It is a two-stage EEG
foundation model that combines decoupled time-frequency tokenization with multi-scale representation learning. Its TFDual-Tokenizer learns separate temporal and frequency codebooks, while EEGSSM combines structured global convolution and sliding-window attention to capture long-range dependencies and local temporal patterns through masked token prediction.

\subsection{Evaluation Metrics}
In this section, we describe the evaluation metrics adopted in this work. Following LaBraM ~\citep{jiang2024large,wang2025cbramod}, we employ the following measures:

\begin{itemize}[leftmargin=10pt]
    \item \textbf{Balanced Accuracy}: Accounts for class imbalance by computing the average recall across all classes. It is applied in both binary and multi-class classification settings.  
    
    \item \textbf{AUC-PR}: The area under the precision-recall (PR) curve, used to evaluate performance in binary classification tasks.  
    
    \item \textbf{AUROC}: The area under the receiver operating characteristic (ROC) curve, a standard metric for assessing binary classification performance.  
    
    \item \textbf{Cohen’s Kappa}: A statistical measure of inter-rater agreement, commonly applied to imbalanced multi-class classification problems to quantify consistency beyond chance.  
    
    \item \textbf{Weighted F1}: The weighted average of class-wise F1-scores, where each class is weighted by its sample size. This provides a robust assessment of performance in multi-class classification tasks.  
\end{itemize}

\section{More Experimental Results}
\label{app:more_results}
\subsection{SEED-VII}
On the SEED-VII dataset (Table~\ref{tab:seedvii}), we observe lower overall performance compared to FACED or SEED-V, highlighting the challenge of distinguishing seven closely related emotions under high inter-subject variability. EEGNet perform poorly, while Conformer and BIOT provide moderate improvements. Foundation models achieve stronger results, with LaBraM performing best overall and CBraMod showing competitive accuracy. BrainPro reaches 0.3315/0.2223/0.3350 (ACC-B/Kappa/F1-W), closely matching CBraMod and LaBraM. Although LaBraM slightly outperforms BrainPro, the margin is small, and BrainPro maintains stable generalization, confirming its effectiveness even in this demanding multi-class setting. This suggests that BrainPro’s retrieval-based spatial modeling and mental-process-separation contribute to consistent robustness across diverse emotion recognition tasks.

\begin{table}[!ht]
\centering
\caption{Results on the SEED-VII dataset.}
\label{tab:seedvii}
\begin{threeparttable}
\begin{tabular}{l ccc}
\toprule
\textbf{Methods} & ACC-B & Kappa & F1-W \\
\midrule
EEGNet & 0.1892 ± 0.0056 & 0.0562 ± 0.0071 & 0.1472 ± 0.0123 \\
Conformer & 0.2719 ± 0.0032 & 0.1566 ± 0.0039 & 0.2712 ± 0.0033 \\
BIOT & 0.3005 ± 0.0054 & 0.1859 ± 0.0070 & 0.3055 ± 0.0069 \\
EEGPT & 0.2213 ± 0.0060 & 0.0954 ± 0.0071 & 0.2147 ± 0.0120 \\
LaBraM & \textbf{0.3346} ± 0.0122 & \textbf{0.2256} ± 0.0146 & \textbf{0.3391} ± 0.0121 \\
CBraMod & 0.3318 ± 0.0056 & 0.2236 ± 0.0065 & 0.3384 ± 0.0059 \\
CodeBrain & 0.3333 ± 0.0064 & 0.2249 ± 0.0085 & 0.3386 ± 0.0081 \\

\cellcolor{cyan!15}BrainPro & \cellcolor{cyan!15}0.3315 ± 0.0038 & \cellcolor{cyan!15}0.2223 ± 0.0055 & \cellcolor{cyan!15}0.3350 ± 0.0045 \\
\bottomrule
\end{tabular}
\begin{tablenotes}
\tiny
\item Note: \textbf{Bold} indicates the best performance. \cellcolor{cyan!15}{Cyan highlight} marks our BrainPro.
\end{tablenotes}
\end{threeparttable}
\end{table}

\subsection{Mental Disorder Diagnosis}
On the MDD dataset, all methods achieve relatively high accuracy, suggesting that distinguishing patients from controls is more tractable than fine-grained emotion recognition. EEGNet and Conformer already reach strong performance, while BIOT and LaBraM offer further improvements. BrainPro achieves 0.9161/0.9831/0.9803 (ACC-B/AUC-PR/AUROC), outperforming all baselines across metrics. These results confirm that BrainPro provides robust and reliable representations for clinical EEG applications, where high sensitivity and specificity are crucial.

\begin{table}[!ht]
\centering
\caption{Results on the Major Depressive Disorder dataset.}
\label{tab:mdd}
\begin{threeparttable}
\begin{tabular}{l ccc}
\toprule
\textbf{Methods} & ACC-B & AUC-PR & AUROC \\
\midrule
EEGNet & 0.9113 ± 0.0104 & 0.9632 ± 0.0045 & 0.9512 ± 0.0096 \\
Conformer & 0.8422 ± 0.0344 & 0.9442 ± 0.0502 & 0.9613 ± 0.0064 \\
BIOT & 0.8789 ± 0.0190 & 0.9744 ± 0.0083 & 0.9664 ± 0.0136 \\
EEGPT & 0.8475 ± 0.0933 & 0.9695 ± 0.0076 & 0.9669 ± 0.0069 \\
LaBraM & 0.8986 ± 0.0018 & 0.9791 ± 0.0041 & 0.9754 ± 0.0050 \\
CBraMod & 0.8909 ± 0.0037 & 0.9769 ± 0.0047 & 0.9726 ± 0.0062 \\
CodeBrain & 0.9079 ± 0.0027 & 0.9826 ± 0.0051 & \textbf{0.9804} ± 0.0057 \\
\cellcolor{cyan!15}BrainPro & \cellcolor{cyan!15}\textbf{0.9161} ± 0.0054 & \cellcolor{cyan!15}\textbf{0.9831} ± 0.0038 & \cellcolor{cyan!15}0.9803 ± 0.0046 \\
\bottomrule
\end{tabular}
\begin{tablenotes}
\tiny
\item Note: \textbf{Bold} indicates the best performance. \cellcolor{cyan!15}{Cyan highlight} marks our BrainPro.
\end{tablenotes}
\end{threeparttable}
\end{table}

\subsection{Speech}
On the imagined speech dataset, performance is generally modest across methods, reflecting the inherent difficulty of decoding internally generated speech from EEG. Simpler models such as EEGNet show very low balanced accuracy, while Conformer achieves only small gains. Foundation models perform better, with LaBraM and CBraMod leading the results. BrainPro achieves 0.5253/0.4067/0.5244 (ACC-B/Kappa/F1-W), the highest among all methods, indicating that retrieval-based spatial learning and metnal-process-separation effectively capture subtle neural dynamics underlying imagined speech.

\begin{table}[!ht]
\centering
\caption{Results on the Speech dataset.}
\label{tab:speech}
\begin{threeparttable}
\begin{tabular}{l ccc}
\toprule
\textbf{Methods} & ACC-B & Kappa & F1-W \\
\midrule
EEGNet & 0.2755 ± 0.0183 & 0.0943 ± 0.0228 & 0.2737 ± 0.0180 \\
Conformer & 0.3339 ± 0.0104 & 0.1673 ± 0.0130 & 0.3269 ± 0.0153 \\
BIOT & 0.3016 ± 0.0206 & 0.1270 ± 0.0257 & 0.2999 ± 0.0217 \\
EEGPT & 0.2787 ± 0.0061 & 0.0983 ± 0.0076 & 0.2758 ± 0.0060 \\
LaBraM & 0.4880 ± 0.0161 & 0.3600 ± 0.0201 & 0.4871 ± 0.0165 \\
CBraMod & 0.5061 ± 0.0211 & 0.3827 ± 0.0263 & 0.5050 ± 0.0212 \\
CodeBrain
& 0.5124 ± 0.0043 & 0.3906 ± 0.0054 & 0.5117 ± 0.0033 \\
\cellcolor{cyan!15}BrainPro & \cellcolor{cyan!15}\textbf{0.5253} ± 0.0084 & \cellcolor{cyan!15}\textbf{0.4067} ± 0.0105 & \cellcolor{cyan!15}\textbf{0.5244} ± 0.0090 \\
\bottomrule
\end{tabular}
\begin{tablenotes}
\tiny
\item Note: \textbf{Bold} indicates the best performance. \cellcolor{cyan!15}{Cyan highlight} marks our BrainPro.
\end{tablenotes}
\end{threeparttable}
\end{table}

\begin{table}[!h]
\centering
\setlength{\tabcolsep}{4pt}
\renewcommand{\arraystretch}{1.12}

\begin{threeparttable}
\caption{Robustness under random channel dropping on FACED.
Values are mean $\pm$ std; $\Delta$ denotes the absolute score
change relative to no channel dropping. Bold indicates the
highest mean in each row.}
\label{tab:drop_channel_details}

\begin{tabular}{@{}c cr cr cr@{}}
\toprule
\textbf{Drop}
& \multicolumn{2}{c}{\textbf{BrainPro}}
& \multicolumn{2}{c}{\textbf{CBraMod}}
& \multicolumn{2}{c}{\textbf{LaBraM}} \\
\cmidrule(lr){2-3}
\cmidrule(lr){4-5}
\cmidrule(l){6-7}
& Mean $\pm$ std & $\Delta$
& Mean $\pm$ std & $\Delta$
& Mean $\pm$ std & $\Delta$ \\
\midrule

\multicolumn{7}{@{}l}{\textbf{ACC-B} $\uparrow$} \\
\addlinespace[2pt]
0\%
& $\mathbf{0.5937}\pm0.0087$ & $0$
& $0.5669\pm0.0094$ & $0$
& $0.5224\pm0.0116$ & $0$ \\
10\%
& $\mathbf{0.5747}\pm0.0014$ & $-0.0190$
& $0.5218\pm0.0042$ & $-0.0451$
& $0.4972\pm0.0036$ & $-0.0252$ \\
20\%
& $\mathbf{0.5373}\pm0.0015$ & $-0.0564$
& $0.4890\pm0.0054$ & $-0.0779$
& $0.4725\pm0.0086$ & $-0.0499$ \\
30\%
& $\mathbf{0.4995}\pm0.0056$ & $-0.0942$
& $0.4473\pm0.0035$ & $-0.1196$
& $0.4230\pm0.0085$ & $-0.0994$ \\
40\%
& $\mathbf{0.4639}\pm0.0054$ & $-0.1298$
& $0.4111\pm0.0094$ & $-0.1558$
& $0.3851\pm0.0084$ & $-0.1373$ \\
50\%
& $\mathbf{0.4270}\pm0.0076$ & $-0.1667$
& $0.3671\pm0.0036$ & $-0.1998$
& $0.3397\pm0.0036$ & $-0.1827$ \\

\midrule
\multicolumn{7}{@{}l}{\textbf{Kappa} $\uparrow$} \\
\addlinespace[2pt]
0\%
& $\mathbf{0.5418}\pm0.0092$ & $0$
& $0.5112\pm0.0110$ & $0$
& $0.4610\pm0.0126$ & $0$ \\
10\%
& $\mathbf{0.5187}\pm0.0015$ & $-0.0231$
& $0.4585\pm0.0045$ & $-0.0527$
& $0.4319\pm0.0043$ & $-0.0291$ \\
20\%
& $\mathbf{0.4772}\pm0.0016$ & $-0.0646$
& $0.4234\pm0.0060$ & $-0.0878$
& $0.4049\pm0.0101$ & $-0.0561$ \\
30\%
& $\mathbf{0.4352}\pm0.0063$ & $-0.1066$
& $0.3775\pm0.0039$ & $-0.1337$
& $0.3490\pm0.0097$ & $-0.1120$ \\
40\%
& $\mathbf{0.3949}\pm0.0054$ & $-0.1469$
& $0.3365\pm0.0111$ & $-0.1747$
& $0.3057\pm0.0094$ & $-0.1553$ \\
50\%
& $\mathbf{0.3547}\pm0.0093$ & $-0.1871$
& $0.2876\pm0.0037$ & $-0.2236$
& $0.2544\pm0.0046$ & $-0.2066$ \\

\midrule
\multicolumn{7}{@{}l}{\textbf{F1-W} $\uparrow$} \\
\addlinespace[2pt]
0\%
& $\mathbf{0.6023}\pm0.0061$ & $0$
& $0.5729\pm0.0105$ & $0$
& $0.5259\pm0.0108$ & $0$ \\
10\%
& $\mathbf{0.5785}\pm0.0020$ & $-0.0238$
& $0.5228\pm0.0041$ & $-0.0501$
& $0.4977\pm0.0041$ & $-0.0282$ \\
20\%
& $\mathbf{0.5419}\pm0.0011$ & $-0.0604$
& $0.4919\pm0.0050$ & $-0.0810$
& $0.4736\pm0.0088$ & $-0.0523$ \\
30\%
& $\mathbf{0.5053}\pm0.0055$ & $-0.0970$
& $0.4525\pm0.0029$ & $-0.1204$
& $0.4226\pm0.0082$ & $-0.1033$ \\
40\%
& $\mathbf{0.4699}\pm0.0053$ & $-0.1324$
& $0.4165\pm0.0094$ & $-0.1564$
& $0.3817\pm0.0091$ & $-0.1442$ \\
50\%
& $\mathbf{0.4329}\pm0.0078$ & $-0.1694$
& $0.3728\pm0.0040$ & $-0.2001$
& $0.3349\pm0.0038$ & $-0.1910$ \\
\bottomrule
\end{tabular}
\end{threeparttable}
\end{table}

\subsection{Linear Probing on Representative Affective and Motor-related Tasks}
\label{app:linear_probe}

To further evaluate the quality of pretrained representations, we conduct linear probing on two representative downstream tasks corresponding to the primary process branches in BrainPro: FACED for affective EEG decoding and SHU-MI for motor imagery decoding. The pretrained encoder is frozen and only a linear classifier is trained. This protocol evaluates the linear separability of downstream labels without task-specific encoder adaptation. As shown in Table~\ref{tab:linear_probe}, BrainPro achieves the strongest frozen-encoder
performance among the evaluated EEG foundation-model baselines on both datasets. On FACED, BrainPro improves ACC-B from 0.3146 to 0.3581 over CBraMod. On SHU-MI, BrainPro improves ACC-B from 0.5233 to 0.5992 over LaBraM. These results provide additional evidence that BrainPro learns more linearly separable pretrained representations for affective and motor-related EEG decoding. Full fine-tuning results are reported in Table~\ref{tab:main}, where task-specific adaptation further improves performance.

\begin{table}[ht]
\centering
\caption{Linear probing results on representative affective and motor-related downstream tasks.}
\label{tab:linear_probe}
\resizebox{\textwidth}{!}{
\begin{tabular}{l ccc ccc}
\toprule
\multirow{2}{*}{\textbf{Method}}
& \multicolumn{3}{c}{\textbf{FACED (9-Class)}} 
& \multicolumn{3}{c}{\textbf{SHU-MI (2-Class)}} \\
\cmidrule(lr){2-4} \cmidrule(lr){5-7}
& \textbf{ACC-B} & \textbf{Kappa} & \textbf{F1-W}
& \textbf{ACC-B} & \textbf{AUC-PR} & \textbf{AUROC} \\
\midrule
BIOT (Fixed)
& 0.2775 $\pm$ 0.0318
& 0.1839 $\pm$ 0.0512
& 0.2599 $\pm$ 0.0271
& 0.5123 $\pm$ 0.0206
& 0.5215 $\pm$ 0.0197
& 0.5146 $\pm$ 0.0264 \\

LaBraM (Fixed)
& 0.3004 $\pm$ 0.0458
& 0.2377 $\pm$ 0.0617
& 0.2943 $\pm$ 0.0375
& 0.5233 $\pm$ 0.0272
& 0.5329 $\pm$ 0.0284
& 0.5218 $\pm$ 0.0275 \\

CBraMod (Fixed)
& 0.3146 $\pm$ 0.0346
& 0.2579 $\pm$ 0.0542
& 0.3077 $\pm$ 0.0298
& 0.5217 $\pm$ 0.0247
& 0.5304 $\pm$ 0.0253
& 0.5238 $\pm$ 0.0317 \\

\rowcolor{cyan!15}
BrainPro (Fixed)
& \textbf{0.3581 $\pm$ 0.0052}
& \textbf{0.2749 $\pm$ 0.0060}
& \textbf{0.3564 $\pm$ 0.0055}
& \textbf{0.5992 $\pm$ 0.0104}
& \textbf{0.6289 $\pm$ 0.0018}
& \textbf{0.6483 $\pm$ 0.0023} \\
\bottomrule
\end{tabular}
}
\end{table}

\section{Effects of Learning Rate}
\begin{figure}[!ht]
    \centering
    \includegraphics[width=0.8\textwidth]{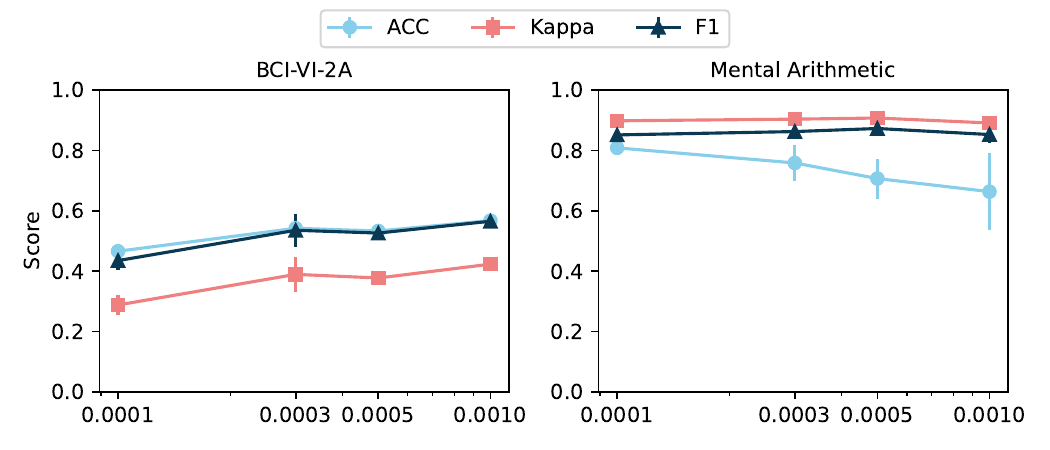}
    \caption{Effects of learning rate on BrainPro fine-tuning.}
    \label{fig:learning_rate_effects}
\end{figure}

Figure~\ref{fig:learning_rate_effects} shows how different learning rates affect BrainPro on BCI-IV-2a and Mental Arithmetic. On BCI-IV-2a, the highest discriminative performance is achieved at a relatively larger learning rate ($1\times10^{-3}$), whereas very small values reduce accuracy and kappa, suggesting that sufficient step size is needed for stable convergence on motor-related tasks. In contrast, Mental Arithmetic benefits from smaller learning rates, with accuracy gradually improving as the rate decreases, while global metrics such as AUC-PR and AUROC remain largely stable. Although these results are based on only two datasets, they highlight that learning rate can influence fine-tuning in a dataset-dependent manner, and that adapting optimization strategies to task characteristics may further enhance BrainPro’s generalization ability.

\section{Effect of Encoder Count and MLP Size on Performance}
\label{app:improvement_is_not_MLP}
Using multiple encoders improves performance over single-encoder variants, while enlarging the MLP alone does not. Table~\ref{tab:model_size} shows how downstream performance varies with different encoder counts and MLP sizes. Models that use only a single encoder perform worse on both FACED and BCI-IV-2A, even when their MLP size is increased to match or exceed that of multi-encoder configurations. In contrast, BrainPro, which uses two encoders, consistently outperforms the single-encoder variants despite having a comparable MLP ratio. Although adding a third encoder can provide additional gains in some settings, the main performance improvement arises from using more than one encoder rather than from simply increasing model capacity. These results indicate that architectural structure plays a larger role in performance than scaling the MLP alone, specifically incorporating multiple complementary encoders.

\begin{table}[!ht]
\centering
\caption{Effect of encoder count and relative MLP size on performance. MLP Ratio is computed w.r.t.\ BrainPro.}
\label{tab:model_size}
\resizebox{\textwidth}{!}{
\begin{threeparttable}
\begin{tabular}{c cccc cccc}
\toprule

\multirow{2}{*}{\textbf{Encoder No.}} 
& \multicolumn{4}{c}{\textbf{FACED}} 
& \multicolumn{4}{c}{\textbf{BCI-IV-2A}} \\
\cmidrule(lr){2-5} \cmidrule(lr){6-9}
& ACC-B & Kappa & F1-W & MLP Ratio
& ACC-B & Kappa & F1-W & MLP Ratio \\
\midrule

1 
& 0.5635 ± 0.0031 & 0.5088 ± 0.0050 & 0.5710 ± 0.0049 & 0.51
& 0.5527 ± 0.0218 & 0.4036 ± 0.0290 & 0.5504 ± 0.0236 & 0.59 \\ 

1
& 0.5808 ± 0.0048 & 0.5255 ± 0.0068 & 0.5840 ± 0.0079 & 1.02
& 0.5403 ± 0.0206 & 0.3871 ± 0.0275 & 0.5368 ± 0.0245 & 1.16 \\

\cellcolor{cyan!15}2
& \cellcolor{cyan!15}0.5937 ± 0.0087
& \cellcolor{cyan!15}0.5418 ± 0.0092
& \cellcolor{cyan!15}0.6023 ± 0.0061
& \cellcolor{cyan!15}1.00
& \cellcolor{cyan!15}0.5674 ± 0.0148
& \cellcolor{cyan!15}0.4232 ± 0.0198
& \cellcolor{cyan!15}0.5653 ± 0.0169
& \cellcolor{cyan!15}1.00 \\

3
& 0.5938 ± 0.0069 & 0.5427 ± 0.0069 & 0.6039 ± 0.0042 & 0.93
& \textbf{0.5889} ± 0.0165 & \textbf{0.4519} ± 0.0220 & \textbf{0.5868} ± 0.0157 & 1.07 \\

3
& \textbf{0.5988} ± 0.0057 
& \textbf{0.5477} ± 0.0058 
& \textbf{0.6098} ± 0.0043 
& \textbf{1.49}
& 0.5797 ± 0.0229 & 0.4396 ± 0.0305 & 0.5764 ± 0.0245 
& 1.41 \\

\bottomrule
\end{tabular}
\begin{tablenotes}
\tiny
\item Note: \textbf{Bold} marks the best performance per dataset block. \cellcolor{cyan!15}{Cyan highlight} marks BrainPro.
\end{tablenotes}
\end{threeparttable}}
\end{table}

\section{Effects of Clear Data Pretraining}
\begin{figure}[!ht]
    \centering   \includegraphics[width=0.8\textwidth]{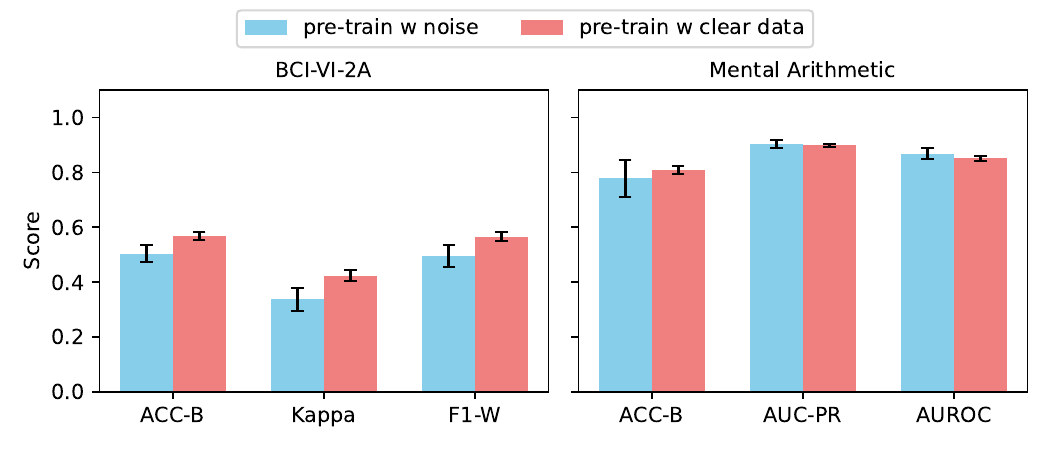}
    \caption{Effects of noisy samples on BrainPro pretraining.}
    \label{fig:noisy_effects}
\end{figure}

Figure~\ref{fig:noisy_effects} illustrates the effect of noisy samples during pretraining. On the BCI-IV-2a dataset, models pre-trained with clear data outperform those trained with noisy data across all classification metrics, with substantial gains in accuracy, Cohen’s kappa, and F1-score. On the Mental Arithmetic dataset, the differences are less pronounced, with only moderate improvements in AUC-PR and AUROC, and in some cases noisy pretraining yields comparable results. These findings suggest that while global performance metrics are relatively robust to noise, discriminability is significantly compromised when noisy data are included in pretraining. This emphasizes that high-quality pretraining data is critical for enabling BrainPro to learn robust and generalizable EEG representations.

\section{Effects of Resetting the Temporal Position Embedding}
\label{app:reset_t_pos_embd}
Table~\ref{tab:reset_pos_embed} reports the results of resetting vs. not resetting temporal position embeddings on both BCI-VI-2A and Mental Arithmetic datasets. The results show that resetting consistently improves performance across all metrics. On BCI-VI-2A, resetting yields notable gains, with ACC-B increasing from 0.5222 to 0.5674 and Kappa improving from 0.3630 to 0.4232, indicating better discriminative ability and robustness. Similarly, for the Mental Arithmetic dataset, resetting provides smaller but consistent improvements across ACC-B, AUC-PR, and AUROC.

The resetting strategy is applied after loading the pre-trained weights by re-initializing the temporal position embeddings with a Xavier uniform distribution. Importantly, after resetting, the embeddings remain learnable and are optimized during downstream fine-tuning. This ensures that the model retains the ability to encode task-specific temporal dynamics, while avoiding potential misalignment introduced by directly transferring positional priors from pretraining. The motivation behind this strategy lies in the observation that, although pretraining captures general mental-process patterns, the temporal ordering of such patterns may vary substantially across unseen downstream tasks. Resetting the embeddings allows the model to relearn temporal structures tailored to the target dataset, which likely explains the consistent improvements observed

\begin{table}[!ht]
\centering
\scriptsize
\caption{Effect of resetting temporal position embedding on BCI-IV-2A and Mental Arithmetic datasets.}
\label{tab:reset_pos_embed}
\resizebox{\textwidth}{!}{
\begin{threeparttable}
\begin{tabular}{l ccc ccc}
\toprule
\multirow{2}{*}{\textbf{Setting}} & \multicolumn{3}{c}{\textbf{BCI-IV-2A}} & \multicolumn{3}{c}{\textbf{Mental Arithmetic}} \\
\cmidrule(lr){2-4} \cmidrule(lr){5-7}
 & ACC-B & Kappa & F1-W & ACC-B & AUC-PR & AUROC \\
\midrule
w/o reset & 0.5222 ± 0.0406 & 0.3630 ± 0.0541 & 0.5109 ± 0.0490 & 0.8017 ± 0.0122 & 0.8955 ± 0.0113 & 0.8471 ± 0.0184 \\
reset     & \textbf{0.5674} ± 0.0148 & \textbf{0.4232} ± 0.0198 & \textbf{0.5653} ± 0.0169 & \textbf{0.8083} ± 0.0156 & \textbf{0.8980} ± 0.0052 & \textbf{0.8512} ± 0.0083 \\
\bottomrule
\end{tabular}
\begin{tablenotes}
\scriptsize
\item \textbf{Note:} Bold values indicate superior performance compared to the non-reset baseline.
\end{tablenotes}
\end{threeparttable}
}
\end{table}

\section{Effects of Different Token Selection Mode}
\label{app:token_selection}

Table~\ref{tab:token_merge} compares different token selection strategies for constructing segment-level representations. On BCI-IV-2a, the aggregation-based strategy (“Aggr”) achieves the best performance across all three metrics, outperforming both mean pooling and using all tokens, which suggests that aggregation provides a more compact and discriminative summary of motor imagery signals. In contrast, on the Mental Arithmetic dataset, mean pooling yields the best results on all metrics, with substantial gains in AUC-PR and AUROC, highlighting that a simpler averaging strategy is more effective for cognitive-related signals. Using all tokens performs worse on both datasets, likely due to redundancy and noise accumulation. These results demonstrate that the optimal token merging strategy is dataset-dependent, and that BrainPro’s flexible design allows for effective adaptation across different EEG domains.

\begin{table}[!ht]
\centering
\scriptsize
\caption{Comparison of different token selection strategies on BCI-VI-2A and Mental Arithmetic datasets.}
\label{tab:token_merge}
\renewcommand{\arraystretch}{1.2}
\resizebox{\textwidth}{!}{
\begin{threeparttable}
\begin{tabular}{l ccc ccc}
\toprule
\multirow{2}{*}{\textbf{Token Merge}} & \multicolumn{3}{c}{\textbf{BCI-VI-2A}} & \multicolumn{3}{c}{\textbf{Mental Arithmetic}} \\
\cmidrule(lr){2-4} \cmidrule(lr){5-7}
 & ACC-B & Kappa & F1-W & ACC-B & AUC-PR & AUROC \\
\midrule
Aggr & \textbf{0.5674} $\pm$ 0.0148 & \textbf{0.4232} $\pm$ 0.0198 & \textbf{0.5653} $\pm$ 0.0169 & 0.7367 $\pm$ 0.0315 & 0.8353 $\pm$ 0.0175 & 0.7822 $\pm$ 0.0141 \\
Mean & 0.4090 $\pm$ 0.0470 & 0.2120 $\pm$ 0.0627 & 0.3664 $\pm$ 0.0703 & \textbf{0.8083} $\pm$ 0.0156 & \textbf{0.8980} $\pm$ 0.0052 & \textbf{0.8512} $\pm$ 0.0083 \\
All & 0.5373 $\pm$ 0.0315 & 0.3831 $\pm$ 0.0421 & 0.5295 $\pm$ 0.0335 & 0.5883 $\pm$ 0.0347 & 0.6753 $\pm$ 0.0131 & 0.6293 $\pm$ 0.0160 \\
\bottomrule
\end{tabular}
\begin{tablenotes}
\scriptsize
\item \textbf{Note:} \textbf{Bold} indicates the best performance for each dataset.
\end{tablenotes}
\end{threeparttable}
}
\end{table}

\section{Parameters size and FLOPS}
Table~\ref{tab:model_complexity} compares the parameter counts and FLOPs of different EEG models, using the BCI-IV-2A dataset as the example for evaluation. Lightweight baselines such as EEGNet and Conformer have very few parameters (0.005M and 0.17M, respectively) and correspondingly low FLOPs, making them efficient but limited in representational capacity. In contrast, LaBraM, BIOT, EEGPT, CBraMod, and our proposed BrainPro are representative large EEG models, with parameter counts ranging from 3M to over 50M and FLOPs spanning hundreds of millions to several billions. Among them, LaBraM is reported with all tokens enabled, since this configuration achieves stronger performance despite higher computational cost. For a fair comparison, the MLP structure and token selection strategy are kept consistent across all foundation models. Notably, BrainPro contains 7.69M parameters and requires 531.04M FLOPs, positioning it as a mid-sized large model that balances complexity and efficiency. Compared to EEGPT (51.66M parameters, 9.31G FLOPs), BrainPro achieves competitive model capacity at a fraction of the computational cost, demonstrating a favorable trade-off between scalability and practicality.

\begin{table}[!ht]
\centering
\caption{Comparison of model complexity.}
\label{tab:model_complexity}
\begin{threeparttable}
\begin{tabular}{lcc}
\toprule
\textbf{Method} & \textbf{Parameters} & \textbf{FLOPs} \\
\midrule
EEGNet      & 0.005M  & 10.03M \\
Conformer   & 0.17M   & 50.09M \\
BIOT        & 3.28M   & 229.35M \\
EEGPT       & 51.66M  & 9.31G \\
LaBraM      & 9.43M   & 392.68M \\
CBraMod     & 8.45M   & 350.79M \\
\rowcolor{cyan!15}BrainPro    & 7.69M   & 531.04M \\
\bottomrule
\end{tabular}
\end{threeparttable}
\end{table}

\section{Visualization of Learned Spatial Filters in Pretraining}
\label{app:visual_spatial_filters}
\begin{figure*}[!ht]
    \centering
    
    \begin{subfigure}{0.15\textwidth}
        \includegraphics[width=\linewidth]{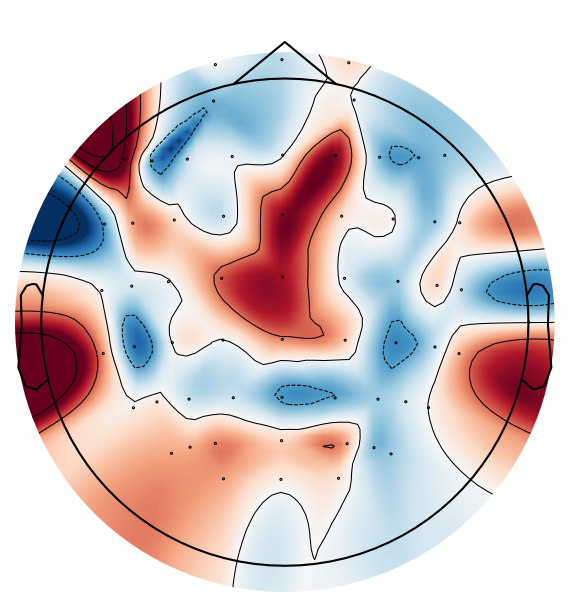}
        \caption{Affect 1}
    \end{subfigure}
    \begin{subfigure}{0.15\textwidth}
        \includegraphics[width=\linewidth]{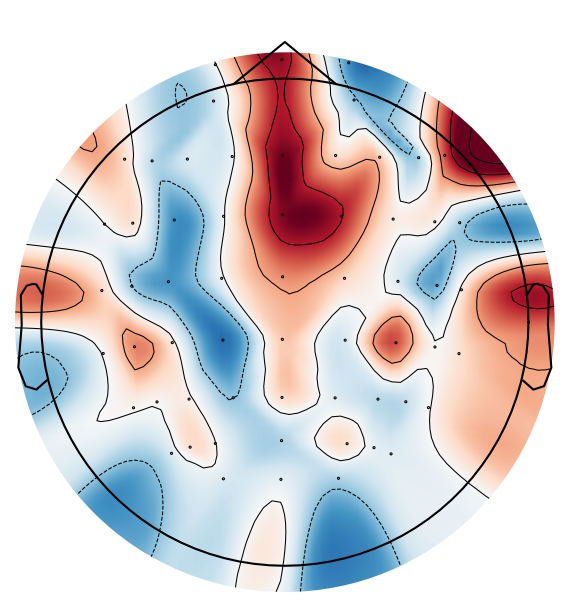}
        \caption{Affect 2}
    \end{subfigure}
    \begin{subfigure}{0.15\textwidth}
        \includegraphics[width=\linewidth]{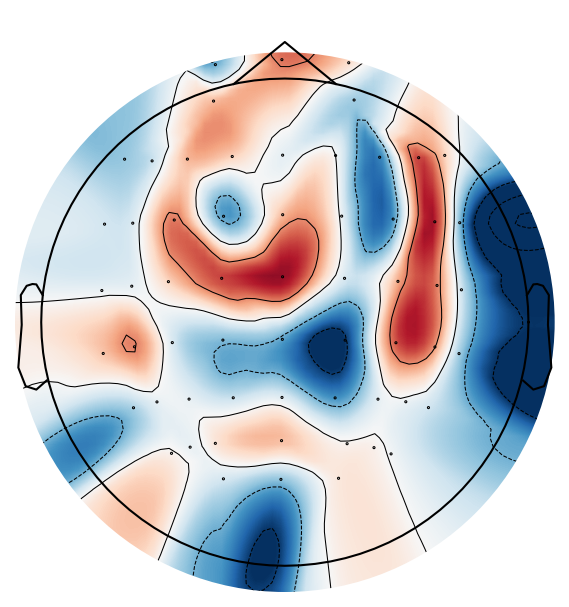}
        \caption{Affect 3}
    \end{subfigure}
    \begin{subfigure}{0.15\textwidth}
        \includegraphics[width=\linewidth]{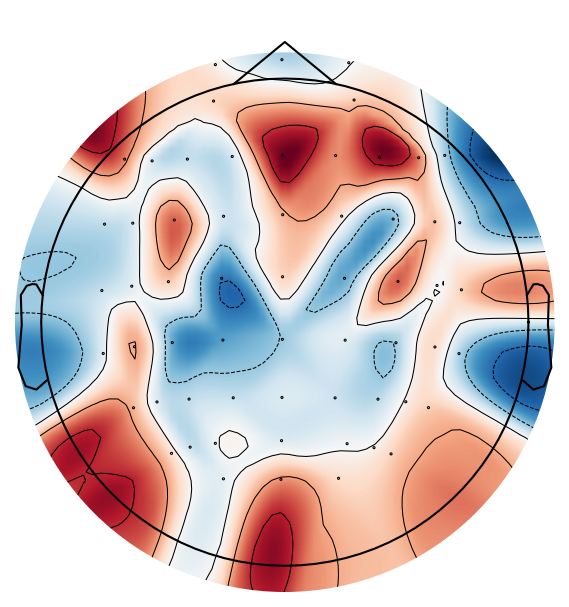}
        \caption{Affect 4}
    \end{subfigure}
    \begin{subfigure}{0.15\textwidth}
        \includegraphics[width=\linewidth]{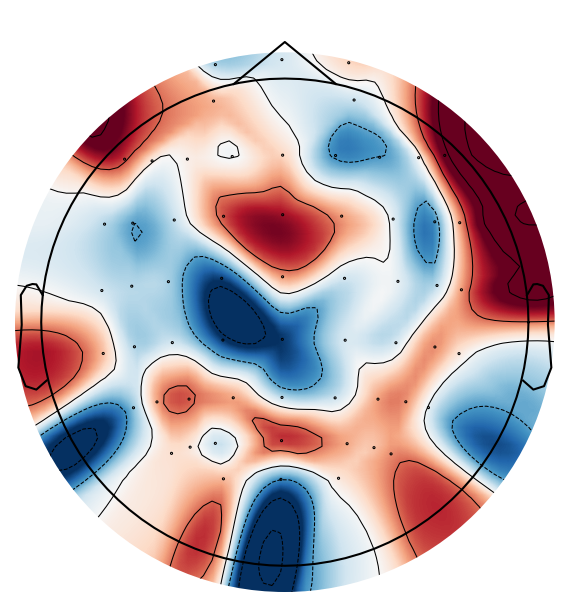}
        \caption{Affect 5}
    \end{subfigure}
    \begin{subfigure}{0.15\textwidth}
        \includegraphics[width=\linewidth]{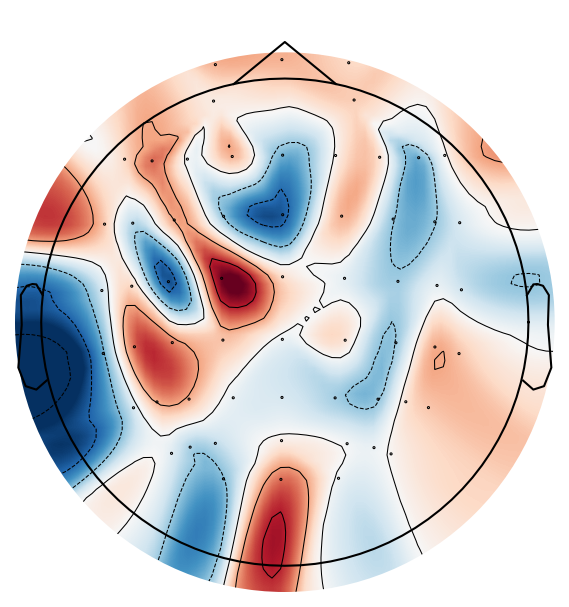}
        \caption{Affect 6}
    \end{subfigure}

    \vspace{0.5em}
    
    \begin{subfigure}{0.15\textwidth}
        \includegraphics[width=\linewidth]{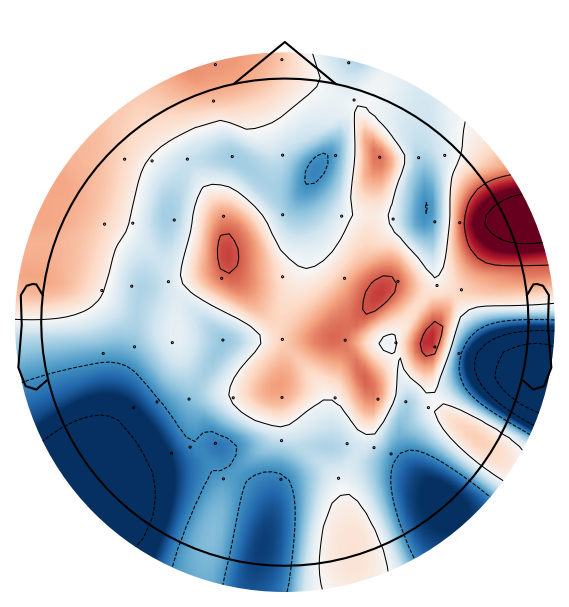}
        \caption{Motor 1}
    \end{subfigure}
    \begin{subfigure}{0.15\textwidth}
        \includegraphics[width=\linewidth]{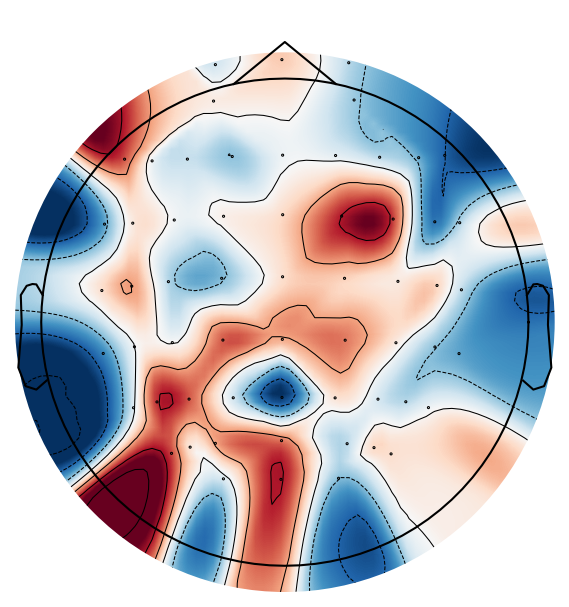}
        \caption{Motor 2}
    \end{subfigure}
    \begin{subfigure}{0.15\textwidth}
        \includegraphics[width=\linewidth]{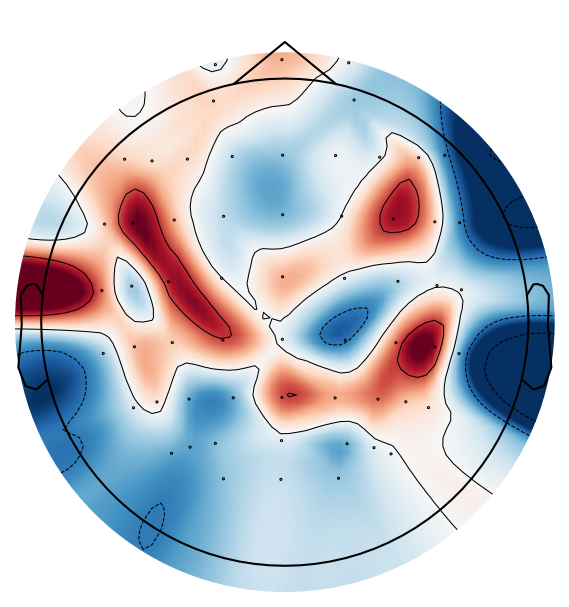}
        \caption{Motor 3}
    \end{subfigure}
    \begin{subfigure}{0.15\textwidth}
        \includegraphics[width=\linewidth]{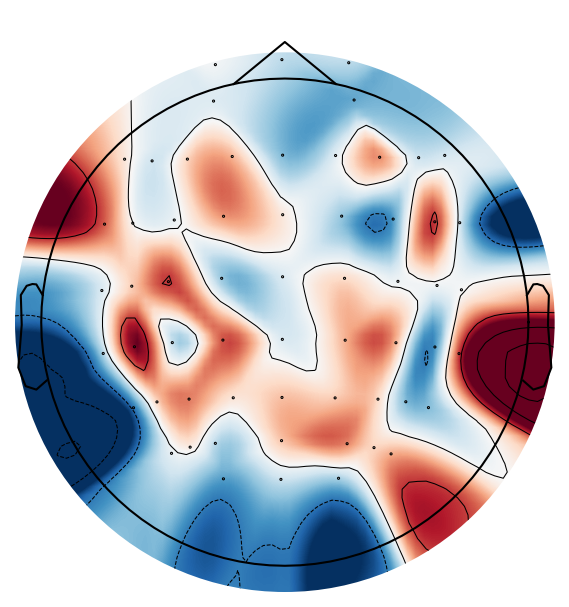}
        \caption{Motor 4}
    \end{subfigure}
    \begin{subfigure}{0.15\textwidth}
        \includegraphics[width=\linewidth]{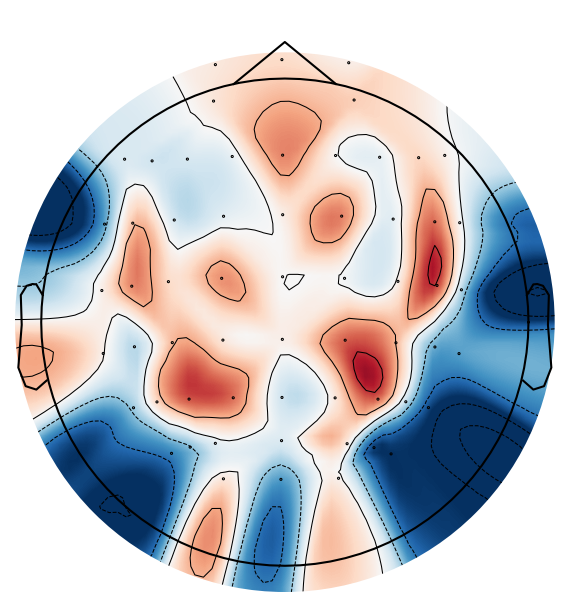}
        \caption{Motor 5}
    \end{subfigure}
    \begin{subfigure}{0.15\textwidth}
        \includegraphics[width=\linewidth]{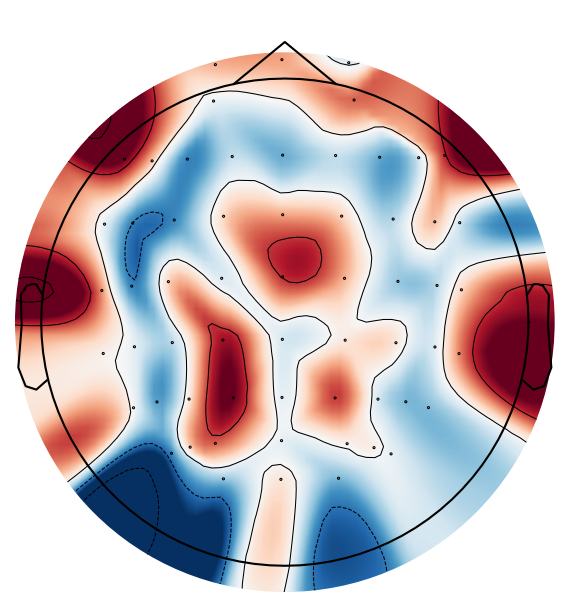}
        \caption{Motor 6}
    \end{subfigure}

    \vspace{0.5em}
    
    \begin{subfigure}{0.15\textwidth}
        \includegraphics[width=\linewidth]{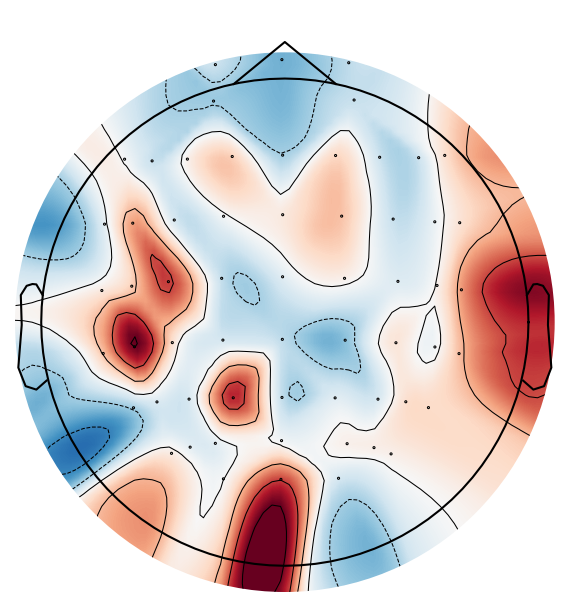}
        \caption{Shared 1}
    \end{subfigure}
    \begin{subfigure}{0.15\textwidth}
        \includegraphics[width=\linewidth]{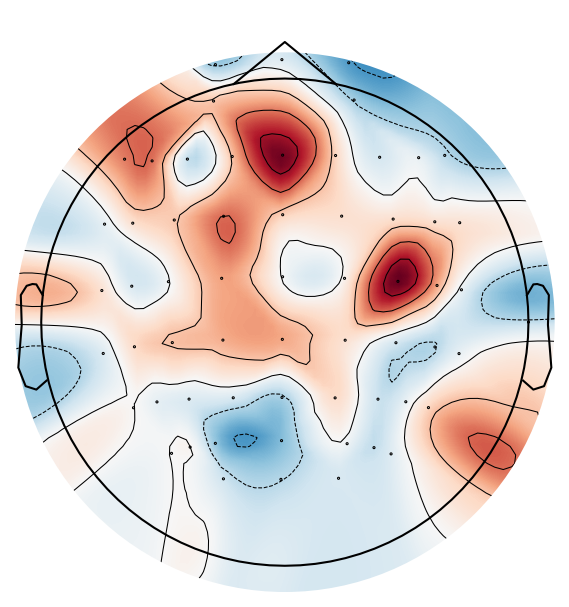}
        \caption{Shared 2}
    \end{subfigure}
    \begin{subfigure}{0.15\textwidth}
        \includegraphics[width=\linewidth]{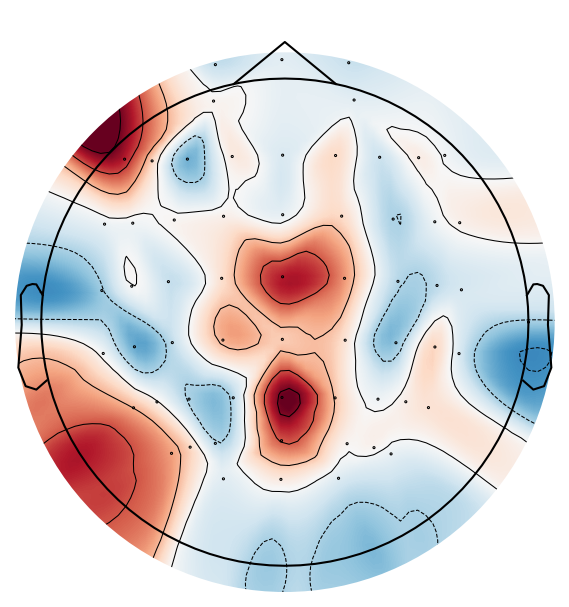}
        \caption{Shared 3}
    \end{subfigure}
    \begin{subfigure}{0.15\textwidth}
        \includegraphics[width=\linewidth]{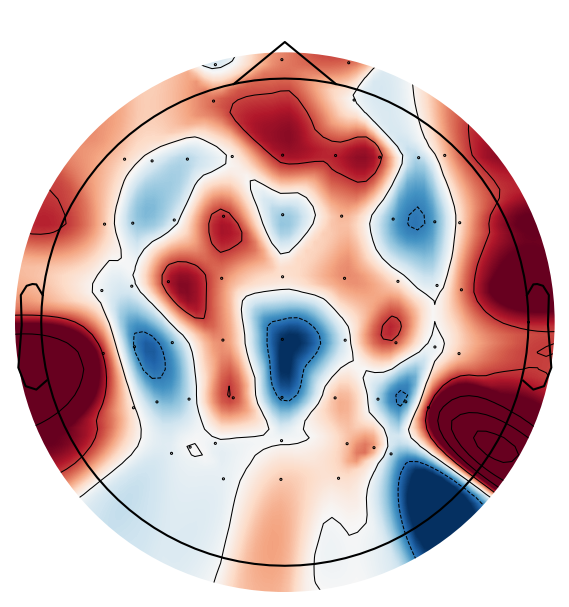}
        \caption{Shared 4}
    \end{subfigure}
    \begin{subfigure}{0.15\textwidth}
        \includegraphics[width=\linewidth]{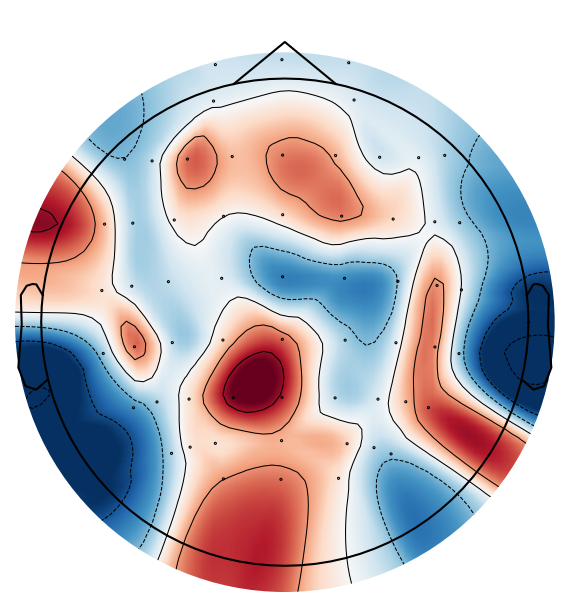}
        \caption{Shared 5}
    \end{subfigure}
    \begin{subfigure}{0.15\textwidth}
        \includegraphics[width=\linewidth]{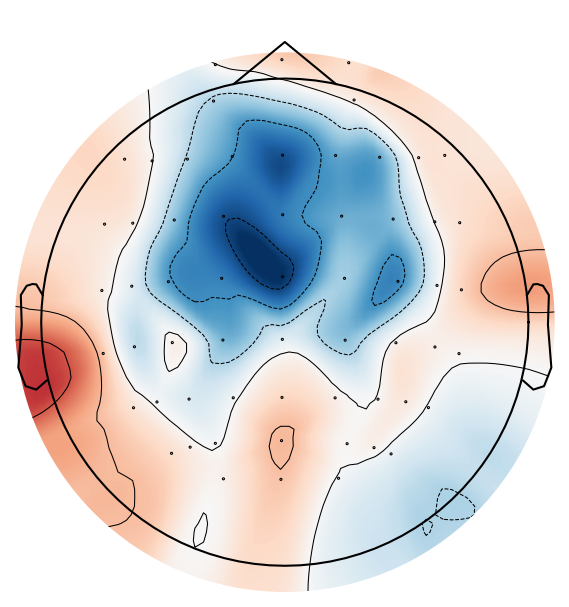}
        \caption{Shared 6}
    \end{subfigure}
    
    \caption{Visualization of learned filters for three groups of encoders after pretraining: Affect (top row), Motor (middle row), and Shared (bottom row).}
    \label{fig:spatial_filters}
\end{figure*}

As shown in Figure~\ref{fig:spatial_filters}, the affect encoders predominantly emphasize frontal and temporal regions, which are consistent with neural substrates associated with emotional processing~\citep{7946165,gao2021novel}, whereas the motor encoders focus on central and parietal regions, aligning with sensorimotor activity~\citep{PFURTSCHELLER2006153}. In contrast, the shared encoders exhibit more diverse and distributed spatial patterns, suggesting that they capture cross-domain representations that generalize across both affective and motor mental processes. These results demonstrate that the proposed pretraining strategy learns spatial filters that not only support downstream performance but also provide qualitative evidence that the learned spatial filters are consistent with coarse neurophysiological priors.

\section{Additional Comparisons and Diagnostics}
\label{app:rebuttal_evidence}
\subsection{Further Analysis on SEED-VII}
\label{app:new_benchmark}

\begin{figure}[!ht]
    \centering
    \includegraphics[width=\linewidth]
        {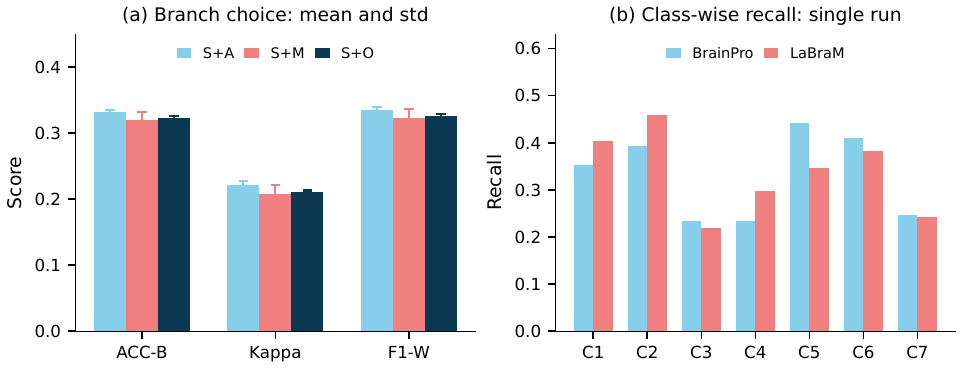}
    \caption{SEED-VII analysis. (a) Performance of three branch configurations. (b) Class recall (7 classes) calculated as $n_{ii}/\sum_j n_{ij}$ from single-run confusion matrices.}
    \label{fig:new_seedvii}
\end{figure}

Figure~\ref{fig:new_seedvii} examines branch selection and class-level performance on SEED-VII. S+A achieves the highest mean scores among the three branch configurations. Its ACC-B is 0.3315, compared with 0.3346 for LaBraM. The recall comparison shows class-dependent differences between BrainPro and LaBraM. These results identify SEED-VII as a benchmark for further improvement.

\subsection{Data Scale and Separation-margin Sensitivity}
\label{app:new_scaling}

\begin{figure}[!ht]
    \centering
    \includegraphics[width=\linewidth]
        {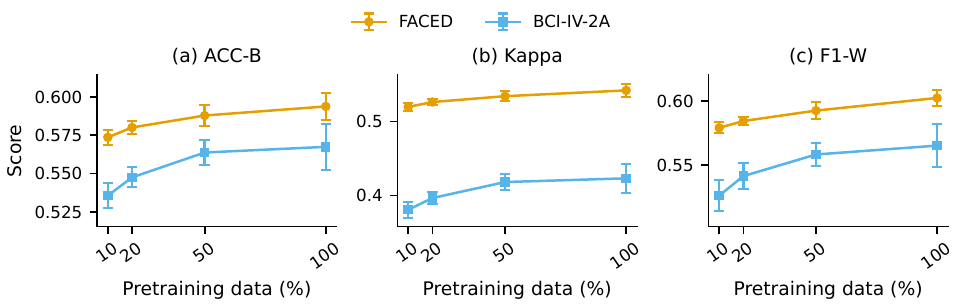}
    \caption{Effect of pretraining data size on FACED and BCI-IV-2A. Each panel uses a restricted vertical range to show performance changes.}
    \label{fig:data_scaling}
\end{figure}

\paragraph{Data scaling.}
As shown in Figure~\ref{fig:data_scaling}, mean scores increase across the 10\%, 20\%, 50\%, and 100\% data settings on both tasks. FACED ACC-B increases from 0.5736 to 0.5937. BCI-IV-2A ACC-B increases from 0.5356 to 0.5674. These changes correspond to gains of 2.01 and 3.18 percentage points, respectively. Increasing the data fraction from 50\% to 100\% yields further gains of 0.59 and 0.37 percentage points. The full retained corpus contains 2,180.38 hours. Proportional estimates for the four data fractions are approximately 218, 436, 1,090, and 2,180 hours. The results show consistent improvements with increasing pretraining data within the evaluated range.

\begin{figure}[!htp]
    \centering
    \includegraphics[width=\linewidth]
        {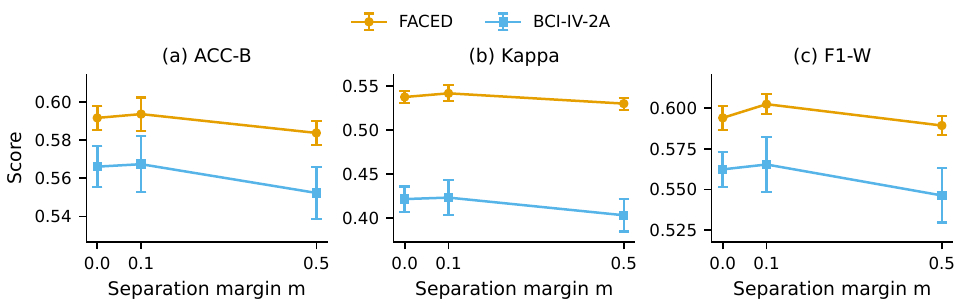}
    \caption{Effect of the separation margin on FACED and BCI-IV-2A. The three panels show ACC-B, Kappa, and F1-W.}
    \label{fig:margin}
\end{figure}

\paragraph{Separation margin.}
The default margin is fixed at $m=0.1$. We further compare $m\in\{0,0.1,0.5\}$ on both emotion and motor imagery tasks. As shown in Figure~\ref{fig:margin}, margins of 0 and 0.1 yield similar mean scores. A margin of 0.5 yields lower mean scores. In $\max(\mathrm{sim}_{\cos}-m,0)$, a larger margin reduces the penalty for a given similarity. The observed performance decrease at $m=0.5$ is consistent with weaker separation pressure.

\subsection{Region Weighting, Spatial retrieval, and Fusion}
\label{app:new_design}

\begin{figure}[h]
    \centering
    \includegraphics[width=\linewidth]
        {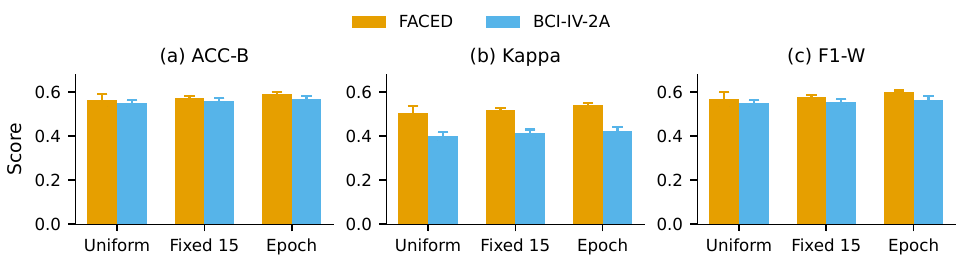}
    \caption{Comparison of regional weighting schedules on FACED and BCI-IV-2A. The settings use uniform weighting, fixed sharpness of 15, and epoch-dependent sharpness.}
    \label{fig:region_weighting}
\end{figure}

\paragraph{Region weighting.}
As shown in Figure~\ref{fig:region_weighting}, epoch-dependent sharpness achieves the highest mean scores among the three weighting settings on both tasks. Relative to fixed sharpness, ACC-B increases by 1.90 \% on FACED and 0.74 \% on BCI-IV-2A. The sharpness parameter controls the contrast between regional weights. The weights remain bounded between 0.5 and 1.5 throughout training. The schedule therefore increases regional emphasis within a fixed weight range.

\begin{figure}[h]
    \centering
    \includegraphics[width=\linewidth]
        {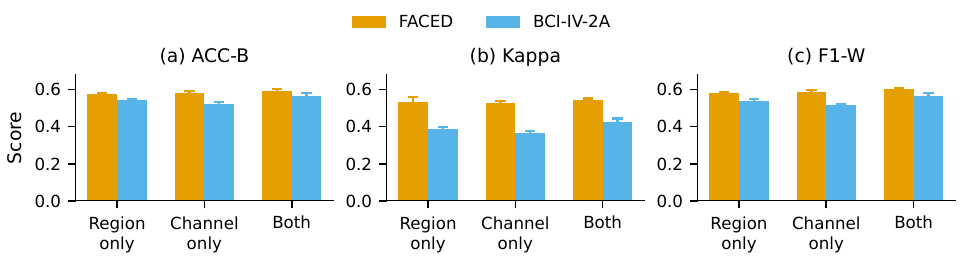}
    \caption{Comparison of spatial retrieval configurations on FACED and BCI-IV-2A. Region only uses regional retrieval. Channel only uses channel retrieval. Both combines the two spatial scales.}
    \label{fig: ablation_retrieval}
\end{figure}

\begin{figure}[!h]
    \centering
    \includegraphics[width=\linewidth]
        {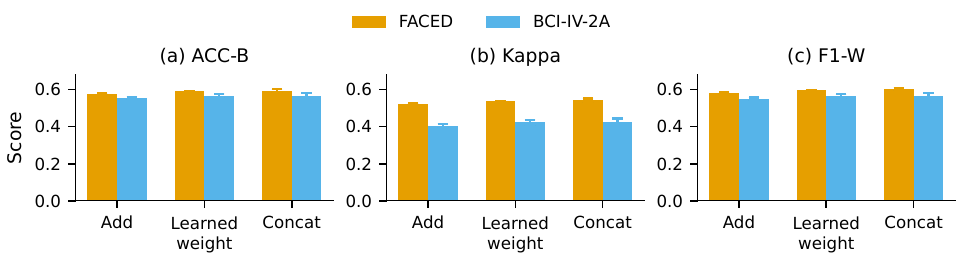}
    \caption{Comparison of feature fusion methods on FACED and BCI-IV-2A. The alternatives are addition, learned weighting, and concatenation.}
    \label{fig:ablation_fusion}
\end{figure}

\paragraph{Complementary spatial scales.}
Combined channel and region retrieval achieves the highest mean scores across all three metrics on both tasks shown in Figure~\ref{fig: ablation_retrieval}. This result supports the complementary roles of the two spatial scales in the evaluated architecture.

\paragraph{Fusion.}
Concatenation retains fine-scale and coarse-scale features as separate components before Transformer encoding.
As shown in Figure~\ref{fig:ablation_fusion}, concatenation achieves the highest mean scores among the three evaluated fusion methods. Learned weighting yields similar performance on BCI-IV-2A. Its ACC-B is 0.5667, compared with 0.5674 for concatenation. These results support concatenation in the evaluated configuration.

\subsection{Process Information and Dataset Identity}
\label{app:new_process}
Process categories may correlate with source datasets, devices, laboratories, and preprocessing pipelines. We assess this confounding through dataset identity prediction within process families and process prediction across datasets. 
\begin{figure}[h]
    \centering
    \includegraphics[width=\linewidth]
        {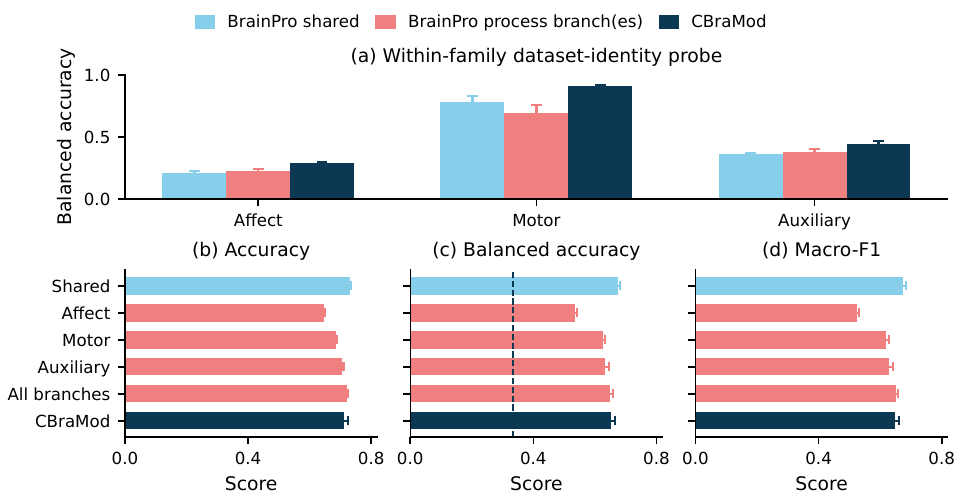}
    \caption{Linear probes on frozen representations. (a) Dataset identity prediction within each process family. The matched process branches are A, M, and O. Panels (b), (c), and (d) show process prediction with one dataset excluded from probe training. Dataset identity comparisons are made within each family. The dashed line indicates chance balanced accuracy of $1/3$.}
    \label{fig:new_process}
\end{figure}

\paragraph{Probe settings.}
We train class-balanced logistic regression probes on standardized, mean-pooled features from frozen encoders. Dataset identity prediction uses an approximately 80/20 split within each dataset. The test set is a randomly positioned block of sampled windows ordered by their original corpus indices to avoid data leakage. We evaluate five block-selection seeds. For process prediction, we use a leave-one-dataset-out protocol. Each probe is trained on the remaining datasets and evaluated on the held-out dataset. Predictions are pooled across held-out datasets to calculate accuracy, balanced accuracy, and macro-F1.

\paragraph{Results.}
Figure~\ref{fig:new_process}(a) shows lower dataset identity prediction scores for the matched BrainPro branches than for CBraMod in all three process families. The affective and auxiliary branches yield higher scores than the shared encoder. The shared encoder achieves the highest process balanced accuracy of 0.675. CBraMod achieves 0.652. The process-specific encoders achieve balanced accuracies between 0.535 and 0.634. All exceed the three-class chance level of $1/3$. These results indicate that BrainPro representations contain both dataset-specific and process-associated information. Process prediction transfers across dataset boundaries, while dataset identity remains decodable within process families. Establishing independence from device, laboratory, and preprocessing effects requires evaluations that control these factors directly.

\begin{figure}[!ht]
    \centering
    \includegraphics[width=\linewidth]
        {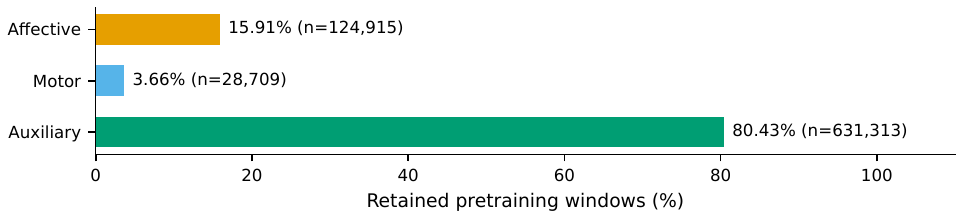}
    \caption{Branch exposure after data cleaning. Bars show each process family's proportion of retained pretraining windows. Labels give the corresponding window counts.}
    \label{fig:branch_exposure}
\end{figure}

\subsection{Branch Activation}
\label{app:branch_activation}

We quantify branch usage during pretraining to characterize the balance of encoder exposure. Data cleaning retains 784,937 windows, corresponding to 92.64\% of generated windows and 2,180.38 hours. As shown in Figure~\ref{fig:branch_exposure}, the affective, motor, and auxiliary groups contribute 124,915, 28,709, and 631,313 retained windows, respectively. Their proportions are 15.91\%, 3.66\%, and 80.43\%. Under sampling proportional to retained window counts, these proportions give the activation probabilities of the corresponding process encoders. The shared encoder is active for every sample. Pretraining exposure is therefore uneven, with the auxiliary encoder receiving the most samples and the motor encoder receiving the fewest. Despite this imbalance, BrainPro improves downstream decoding performance across the evaluated tasks. Balanced or hierarchical process sampling may further improve performance.

\begin{figure}[!ht]
    \centering
    \includegraphics[width=\linewidth]
        {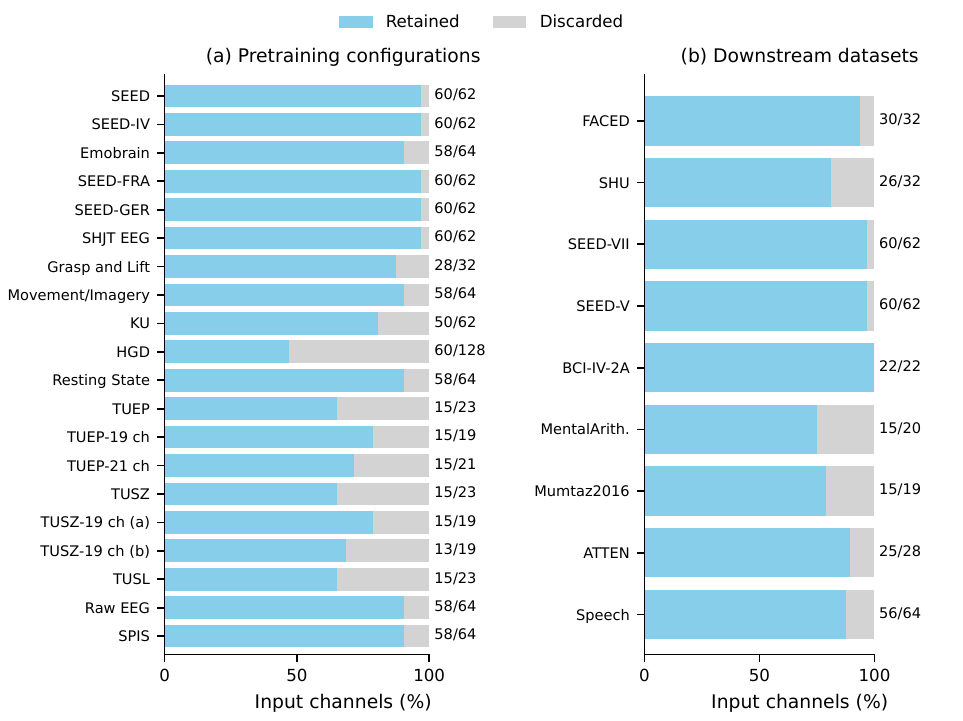}
    \caption{Montage coverage for (a) 19 pretraining configurations
    and (b) nine downstream datasets.
    Skyblue indicates retained channels.
    Light gray indicates discarded channels.
    Labels give retained and input channel counts.
    Retained template indices are unique in all evaluated mappings.
    Repeated TUEP and TUSZ rows represent distinct montage
    configurations.}
    \label{fig:new_mapping}
\end{figure}

\subsection{Montage Mapping}
\label{app:new_mapping}
Figure~\ref{fig:new_mapping} shows that the mapping preserves most input channels in all evaluated configurations except HGD. HGD retains 60 of 128 channels. The high retention across the other configurations supports broad montage compatibility while preserving channel coverage. All evaluated mappings have unique retained template indices. Unsupported channels are excluded.

\section{Discussion}
\label{app:discussion}
\paragraph{New insights.}
This work shows that incorporating coarse process conditioning into large EEG models yields consistent improvements across diverse decoding tasks. 
The spatial filters learned by the affective, motor-related, and shared encoders form distinct and interpretable topographies, and the channel-drop analysis suggests that these spatial representations remain effective when electrodes are missing.
The encoder-combination results indicate that process-associated branches capture task-aligned and complementary features beyond the shared encoder.

\paragraph{Limitations.}
\label{app:discussion_limitation}
Despite these advances, BrainPro has several limitations. First, the current process taxonomy is coarse, mainly covering affective, motor-related data. It does not explicitly model finer-grained processes such as attention, memory, language, or workload, which may limit branch specialization beyond the affective and motor-related settings emphasized in this work. Second, the auxiliary residual branch groups heterogeneous non-affective and non-motor data into a single pathway.

\paragraph{Future work.}
Several directions remain open. Extending BrainPro to richer process annotations, potentially through hierarchical process labels or dynamically allocated process-associated branches, could improve flexibility and interpretability. Incorporating multimodal neurophysiological signals such as fNIRS, MEG, eye tracking, or peripheral physiological signals may further enrich representation learning. Improving efficiency through parameter sharing, pruning, adapter-based tuning, or distillation will be important for real-time and on-device BCI applications. Overall, BrainPro provides a practical step toward scalable EEG pretraining that combines montage-aware spatial modeling with coarse process-conditioned representation learning.

\section{Broader impacts}
\label{app:broader_impacts}
BrainPro may improve robust EEG representation learning across heterogeneous montages and protocols, benefiting BCI, affective computing, and clinical decision-support research. However, EEG data are sensitive biometric and health-related signals, so use requires informed consent, secure data handling, and ethical oversight. Future use should evaluate bias across subjects, devices, and populations, and avoid non-consensual monitoring or high-stakes standalone decisions.


\newpage
\section*{NeurIPS Paper Checklist}

\begin{enumerate}

\item {\bf Claims}
    \item[] Question: Do the main claims made in the abstract and introduction accurately reflect the paper's contributions and scope?
    \item[] Answer: \answerYes{}
    \item[] Justification: The abstract and introduction state the main contributions: topographic spatial retrieval, process-conditioned encoder branches, and process-guided region-weighted reconstruction. The empirical claims are limited to the reported BCI benchmarks and are supported by the main results, ablations, channel-drop analysis, encoder-configuration studies, and spatial-filter visualizations; we also clarify that BrainPro uses coarse mental-process cues and does not identify latent neural states.
    \item[] Guidelines:
    \begin{itemize}
        \item The answer \answerNA{} means that the abstract and introduction do not include the claims made in the paper.
        \item The abstract and/or introduction should clearly state the claims made, including the contributions made in the paper and important assumptions and limitations. A \answerNo{} or \answerNA{} answer to this question will not be perceived well by the reviewers.
        \item The claims made should match theoretical and experimental results, and reflect how much the results can be expected to generalize to other settings.
        \item It is fine to include aspirational goals as motivation as long as it is clear that these goals are not attained by the paper.
    \end{itemize}

\item {\bf Limitations}
    \item[] Question: Does the paper discuss the limitations of the work performed by the authors?
    \item[] Answer: \answerYes{}
    \item[] Justification: The paper includes a limitations discussion (Appendix~\ref{app:discussion_limitation}). It states that the current process taxonomy is coarse, mainly covers affective and motor-related data, does not explicitly model finer-grained processes such as attention, memory, language, and workload, and uses an auxiliary residual branch for heterogeneous non-affective and non-motor data.
    \item[] Guidelines:
    \begin{itemize}
        \item The answer \answerNA{} means that the paper has no limitation while the answer \answerNo{} means that the paper has limitations, but those are not discussed in the paper.
        \item The authors are encouraged to create a separate ``Limitations'' section in their paper.
        \item The paper should point out any strong assumptions and how robust the results are to violations of these assumptions (e.g., independence assumptions, noiseless settings, model well-specification, asymptotic approximations only holding locally). The authors should reflect on how these assumptions might be violated in practice and what the implications would be.
        \item The authors should reflect on the scope of the claims made, e.g., if the approach was only tested on a few datasets or with a few runs. In general, empirical results often depend on implicit assumptions, which should be articulated.
        \item The authors should reflect on the factors that influence the performance of the approach. For example, a facial recognition algorithm may perform poorly when image resolution is low or images are taken in low lighting. Or a speech-to-text system might not be used reliably to provide closed captions for online lectures because it fails to handle technical jargon.
        \item The authors should discuss the computational efficiency of the proposed algorithms and how they scale with dataset size.
        \item If applicable, the authors should discuss possible limitations of their approach to address problems of privacy and fairness.
        \item While the authors might fear that complete honesty about limitations might be used by reviewers as grounds for rejection, a worse outcome might be that reviewers discover limitations that aren't acknowledged in the paper. The authors should use their best judgment and recognize that individual actions in favor of transparency play an important role in developing norms that preserve the integrity of the community. Reviewers will be specifically instructed to not penalize honesty concerning limitations.
    \end{itemize}

\item {\bf Theory assumptions and proofs}
    \item[] Question: For each theoretical result, does the paper provide the full set of assumptions and a complete (and correct) proof?
    \item[] Answer: \answerNA{}
    \item[] Justification: The paper does not present formal theoretical results as main contributions. It provides mathematical formulations and appendix-level justification to motivate retrieval-based spatial encoding in Appendix~\ref{app:Justify_spatial_learning}, while the main claims are methodological and empirical rather than formal theoretical guarantees.
    \item[] Guidelines:
    \begin{itemize}
        \item The answer \answerNA{} means that the paper does not include theoretical results.
        \item All the theorems, formulas, and proofs in the paper should be numbered and cross-referenced.
        \item All assumptions should be clearly stated or referenced in the statement of any theorems.
        \item The proofs can either appear in the main paper or the supplemental material, but if they appear in the supplemental material, the authors are encouraged to provide a short proof sketch to provide intuition.
        \item Inversely, any informal proof provided in the core of the paper should be complemented by formal proofs provided in appendix or supplemental material.
        \item Theorems and Lemmas that the proof relies upon should be properly referenced.
    \end{itemize}

\item {\bf Experimental result reproducibility}
    \item[] Question: Does the paper fully disclose all the information needed to reproduce the main experimental results of the paper to the extent that it affects the main claims and/or conclusions of the paper (regardless of whether the code and data are provided or not)?
    \item[] Answer: \answerYes{}
    \item[] Justification: The paper describes the model architecture, topographic spatial retrieval, process-conditioned branches, masking strategy, reconstruction and separation objectives, downstream adaptation protocol, datasets, preprocessing, evaluation metrics, data-split protocol, random seeds, optimizer, and hyperparameters. Additional details are provided in Section 4 and Appendices G-H, including dataset descriptions, pretraining settings, downstream settings, and baseline information.
    \item[] Guidelines:
    \begin{itemize}
        \item The answer \answerNA{} means that the paper does not include experiments.
        \item If the paper includes experiments, a \answerNo{} answer to this question will not be perceived well by the reviewers: Making the paper reproducible is important, regardless of whether the code and data are provided or not.
        \item If the contribution is a dataset and\slash or model, the authors should describe the steps taken to make their results reproducible or verifiable.
        \item Depending on the contribution, reproducibility can be accomplished in various ways. For example, if the contribution is a novel architecture, describing the architecture fully might suffice, or if the contribution is a specific model and empirical evaluation, it may be necessary to either make it possible for others to replicate the model with the same dataset, or provide access to the model. In general. releasing code and data is often one good way to accomplish this, but reproducibility can also be provided via detailed instructions for how to replicate the results, access to a hosted model (e.g., in the case of a large language model), releasing of a model checkpoint, or other means that are appropriate to the research performed.
        \item While NeurIPS does not require releasing code, the conference does require all submissions to provide some reasonable avenue for reproducibility, which may depend on the nature of the contribution. For example
        \begin{enumerate}
            \item If the contribution is primarily a new algorithm, the paper should make it clear how to reproduce that algorithm.
            \item If the contribution is primarily a new model architecture, the paper should describe the architecture clearly and fully.
            \item If the contribution is a new model (e.g., a large language model), then there should either be a way to access this model for reproducing the results or a way to reproduce the model (e.g., with an open-source dataset or instructions for how to construct the dataset).
            \item We recognize that reproducibility may be tricky in some cases, in which case authors are welcome to describe the particular way they provide for reproducibility. In the case of closed-source models, it may be that access to the model is limited in some way (e.g., to registered users), but it should be possible for other researchers to have some path to reproducing or verifying the results.
        \end{enumerate}
    \end{itemize}

\item {\bf Open access to data and code}
    \item[] Question: Does the paper provide open access to the data and code, with sufficient instructions to faithfully reproduce the main experimental results, as described in supplemental material?
    \item[] Answer: \answerYes{}
    \item[] Justification: The experiments use released EEG datasets, and code is provided.
    \item[] Guidelines:
    \begin{itemize}
        \item The answer \answerNA{} means that paper does not include experiments requiring code.
        \item Please see the NeurIPS code and data submission guidelines (\url{https://neurips.cc/public/guides/CodeSubmissionPolicy}) for more details.
        \item While we encourage the release of code and data, we understand that this might not be possible, so \answerNo{} is an acceptable answer. Papers cannot be rejected simply for not including code, unless this is central to the contribution (e.g., for a new open-source benchmark).
        \item The instructions should contain the exact command and environment needed to run to reproduce the results. See the NeurIPS code and data submission guidelines (\url{https://neurips.cc/public/guides/CodeSubmissionPolicy}) for more details.
        \item The authors should provide instructions on data access and preparation, including how to access the raw data, preprocessed data, intermediate data, and generated data, etc.
        \item The authors should provide scripts to reproduce all experimental results for the new proposed method and baselines. If only a subset of experiments are reproducible, they should state which ones are omitted from the script and why.
        \item At submission time, to preserve anonymity, the authors should release anonymized versions (if applicable).
        \item Providing as much information as possible in supplemental material (appended to the paper) is recommended, but including URLs to data and code is permitted.
    \end{itemize}

\item {\bf Experimental setting/details}
    \item[] Question: Does the paper specify all the training and test details (e.g., data splits, hyperparameters, how they were chosen, type of optimizer) necessary to understand the results?
    \item[] Answer: \answerYes{}
    \item[] Justification: The paper reports the pretraining corpus construction, preprocessing pipeline, segmentation length, resampling frequency, scaling procedure, masking ratio, optimizer, learning-rate schedule, number of epochs, model components, downstream datasets, evaluation metrics, non-overlapping train/validation/test split protocol, validation-based model selection, five random seeds, and hardware. Full hyperparameter settings are provided in Appendices G--H.
    \item[] Guidelines:
    \begin{itemize}
        \item The answer \answerNA{} means that the paper does not include experiments.
        \item The experimental setting should be presented in the core of the paper to a level of detail that is necessary to appreciate the results and make sense of them.
        \item The full details can be provided either with the code, in appendix, or as supplemental material.
    \end{itemize}

\item {\bf Experiment statistical significance}
    \item[] Question: Does the paper report error bars suitably and correctly defined or other appropriate information about the statistical significance of the experiments?
    \item[] Answer: \answerYes{}
    \item[] Justification: The paper reports mean $\pm$ standard deviation over five random seeds $\{0,1,2,3,4\}$ for the main downstream results, ablation studies, encoder-configuration analysis, and appendix experiments. The standard deviation is computed across the five independent runs and captures variability due to random initialization and training randomness under the same experimental protocol.
    \item[] Guidelines:
    \begin{itemize}
        \item The answer \answerNA{} means that the paper does not include experiments.
        \item The authors should answer \answerYes{} if the results are accompanied by error bars, confidence intervals, or statistical significance tests, at least for the experiments that support the main claims of the paper.
        \item The factors of variability that the error bars are capturing should be clearly stated (for example, train/test split, initialization, random drawing of some parameter, or overall run with given experimental conditions).
        \item The method for calculating the error bars should be explained (closed form formula, call to a library function, bootstrap, etc.)
        \item The assumptions made should be given (e.g., Normally distributed errors).
        \item It should be clear whether the error bar is the standard deviation or the standard error of the mean.
        \item It is OK to report 1-sigma error bars, but one should state it. The authors should preferably report a 2-sigma error bar than state that they have a 96\% CI, if the hypothesis of Normality of errors is not verified.
        \item For asymmetric distributions, the authors should be careful not to show in tables or figures symmetric error bars that would yield results that are out of range (e.g., negative error rates).
        \item If error bars are reported in tables or plots, the authors should explain in the text how they were calculated and reference the corresponding figures or tables in the text.
    \end{itemize}

\item {\bf Experiments compute resources}
    \item[] Question: For each experiment, does the paper provide sufficient information on the computer resources (type of compute workers, memory, time of execution) needed to reproduce the experiments?
    \item[] Answer: \answerYes{}
    \item[] Justification: The paper reports that pretraining was conducted on 5 NVIDIA A800 GPUs for approximately 3 days, corresponding to roughly 360 A800 GPU-hours, on a machine with approximately 512 GB RAM and 1 TB storage (Appendix~\ref{app:compute_resources}).
    \item[] Guidelines:
    \begin{itemize}
        \item The answer \answerNA{} means that the paper does not include experiments.
        \item The paper should indicate the type of compute workers CPU or GPU, internal cluster, or cloud provider, including relevant memory and storage.
        \item The paper should provide the amount of compute required for each of the individual experimental runs as well as estimate the total compute.
        \item The paper should disclose whether the full research project required more compute than the experiments reported in the paper (e.g., preliminary or failed experiments that didn't make it into the paper).
    \end{itemize}

\item {\bf Code of ethics}
    \item[] Question: Does the research conducted in the paper conform, in every respect, with the NeurIPS Code of Ethics \url{https://neurips.cc/public/EthicsGuidelines}?
    \item[] Answer: \answerYes{}
    \item[] Justification: The study uses existing public or controlled-access EEG datasets and does not involve new data collection, deception, crowdsourcing, or intervention on human participants. We follow provider-specific dataset terms, use non-overlapping train/validation/test splits to reduce data leakage, and discuss limitations and responsible-use considerations.
    \item[] Guidelines:
    \begin{itemize}
        \item The answer \answerNA{} means that the authors have not reviewed the NeurIPS Code of Ethics.
        \item If the authors answer \answerNo, they should explain the special circumstances that require a deviation from the Code of Ethics.
        \item The authors should make sure to preserve anonymity (e.g., if there is a special consideration due to laws or regulations in their jurisdiction).
    \end{itemize}

\item {\bf Broader impacts}
    \item[] Question: Does the paper discuss both potential positive societal impacts and negative societal impacts of the work performed?
    \item[] Answer: \answerYes{}
    \item[] Justification: The paper discusses potential benefits for EEG representation learning, BCI research, assistive technologies, and robust decoding under heterogeneous montages in Appendix~\ref{app:broader_impacts}. It also acknowledges potential risks of EEG modeling, including privacy concerns, misuse for intrusive monitoring or affective inference, and limitations for clinical or high-stakes deployment without additional validation.
    \item[] Guidelines:
    \begin{itemize}
        \item The answer \answerNA{} means that there is no societal impact of the work performed.
        \item If the authors answer \answerNA{} or \answerNo, they should explain why their work has no societal impact or why the paper does not address societal impact.
        \item Examples of negative societal impacts include potential malicious or unintended uses (e.g., disinformation, generating fake profiles, surveillance), fairness considerations (e.g., deployment of technologies that could make decisions that unfairly impact specific groups), privacy considerations, and security considerations.
        \item The conference expects that many papers will be foundational research and not tied to particular applications, let alone deployments. However, if there is a direct path to any negative applications, the authors should point it out. For example, it is legitimate to point out that an improvement in the quality of generative models could be used to generate Deepfakes for disinformation. On the other hand, it is not needed to point out that a generic algorithm for optimizing neural networks could enable people to train models that generate Deepfakes faster.
        \item The authors should consider possible harms that could arise when the technology is being used as intended and functioning correctly, harms that could arise when the technology is being used as intended but gives incorrect results, and harms following from (intentional or unintentional) misuse of the technology.
        \item If there are negative societal impacts, the authors could also discuss possible mitigation strategies (e.g., gated release of models, providing defenses in addition to attacks, mechanisms for monitoring misuse, mechanisms to monitor how a system learns from feedback over time, improving the efficiency and accessibility of ML).
    \end{itemize}

\item {\bf Safeguards}
    \item[] Question: Does the paper describe safeguards that have been put in place for responsible release of data or models that have a high risk for misuse (e.g., pre-trained language models, image generators, or scraped datasets)?
    \item[] Answer: \answerNA{}
    \item[] Justification: We do not release a high-risk generative model, scraped dataset, or system intended for deployment in high-stakes settings. The paper focuses on EEG representation learning and discusses responsible-use considerations, including that the model should not be used for clinical diagnosis, surveillance, or high-stakes decision-making without proper consent, privacy protection, and task-specific validation.
    \item[] Guidelines:
    \begin{itemize}
        \item The answer \answerNA{} means that the paper poses no such risks.
        \item Released models that have a high risk for misuse or dual-use should be released with necessary safeguards to allow for controlled use of the model, for example by requiring that users adhere to usage guidelines or restrictions to access the model or implementing safety filters.
        \item Datasets that have been scraped from the Internet could pose safety risks. The authors should describe how they avoided releasing unsafe images.
        \item We recognize that providing effective safeguards is challenging, and many papers do not require this, but we encourage authors to take this into account and make a best faith effort.
    \end{itemize}

\item {\bf Licenses for existing assets}
    \item[] Question: Are the creators or original owners of assets (e.g., code, data, models), used in the paper, properly credited and are the license and terms of use explicitly mentioned and properly respected?
    \item[] Answer: \answerNo{}
    \item[] Justification: We cite the original sources for the public and controlled-access EEG datasets and baseline methods used in this study, and dataset details are summarized in the Appendix~\ref{app:pretrainingdatasets} and ~\ref{app:downstream_dataset}. Because the study uses many third-party EEG datasets with heterogeneous access conditions, including datasets requiring registration or application, we do not enumerate every dataset license in the checklist and do not redistribute third-party datasets; users should obtain each dataset from its original provider and comply with the corresponding terms of use.
    \item[] Guidelines:
    \begin{itemize}
        \item The answer \answerNA{} means that the paper does not use existing assets.
        \item The authors should cite the original paper that produced the code package or dataset.
        \item The authors should state which version of the asset is used and, if possible, include a URL.
        \item The name of the license (e.g., CC-BY 4.0) should be included for each asset.
        \item For scraped data from a particular source (e.g., website), the copyright and terms of service of that source should be provided.
        \item If assets are released, the license, copyright information, and terms of use in the package should be provided. For popular datasets, \url{paperswithcode.com/datasets} has curated licenses for some datasets. Their licensing guide can help determine the license of a dataset.
        \item For existing datasets that are re-packaged, both the original license and the license of the derived asset (if it has changed) should be provided.
        \item If this information is not available online, the authors are encouraged to reach out to the asset's creators.
    \end{itemize}

\item {\bf New assets}
    \item[] Question: Are new assets introduced in the paper well documented and is the documentation provided alongside the assets?
    \item[] Answer: \answerYes{}
    \item[] Justification: We introduce the BrainPro model and document its architecture, pretraining objective, topographic spatial retrieval module, downstream adaptation protocol, preprocessing settings, hyperparameters, and evaluation protocol in the main paper and appendices. Any released implementation materials are intended to document and facilitate use of the core BrainPro components, while third-party EEG datasets must be obtained from their original providers.
    \item[] Guidelines:
    \begin{itemize}
        \item The answer \answerNA{} means that the paper does not release new assets.
        \item Researchers should communicate the details of the dataset\slash code\slash model as part of their submissions via structured templates. This includes details about training, license, limitations, etc.
        \item The paper should discuss whether and how consent was obtained from people whose asset is used.
        \item At submission time, remember to anonymize your assets (if applicable). You can either create an anonymized URL or include an anonymized zip file.
    \end{itemize}

\item {\bf Crowdsourcing and research with human subjects}
    \item[] Question: For crowdsourcing experiments and research with human subjects, does the paper include the full text of instructions given to participants and screenshots, if applicable, as well as details about compensation (if any)?
    \item[] Answer: \answerNA{}
    \item[] Justification: We do not conduct crowdsourcing experiments or collect new human-subject data. All experiments use existing public or controlled-access EEG datasets, and participant instructions, compensation, and consent procedures are governed by the original dataset collection studies cited in the paper.
    \item[] Guidelines:
    \begin{itemize}
        \item The answer \answerNA{} means that the paper does not involve crowdsourcing nor research with human subjects.
        \item Including this information in the supplemental material is fine, but if the main contribution of the paper involves human subjects, then as much detail as possible should be included in the main paper.
        \item According to the NeurIPS Code of Ethics, workers involved in data collection, curation, or other labor should be paid at least the minimum wage in the country of the data collector.
    \end{itemize}

\item {\bf Institutional review board (IRB) approvals or equivalent for research with human subjects}
    \item[] Question: Does the paper describe potential risks incurred by study participants, whether such risks were disclosed to the subjects, and whether Institutional Review Board (IRB) approvals (or an equivalent approval/review based on the requirements of your country or institution) were obtained?
    \item[] Answer: \answerNA{}
    \item[] Justification: This work does not involve new human-subject recruitment, intervention, or data collection. The experiments use existing public or controlled-access EEG datasets, for which ethics approvals, consent procedures, and risk disclosures are governed by the original dataset providers and cited dataset papers.
    \item[] Guidelines:
    \begin{itemize}
        \item The answer \answerNA{} means that the paper does not involve crowdsourcing nor research with human subjects.
        \item Depending on the country in which research is conducted, IRB approval (or equivalent) may be required for any human subjects research. If you obtained IRB approval, you should clearly state this in the paper.
        \item We recognize that the procedures for this may vary significantly between institutions and locations, and we expect authors to adhere to the NeurIPS Code of Ethics and the guidelines for their institution.
        \item For initial submissions, do not include any information that would break anonymity (if applicable), such as the institution conducting the review.
    \end{itemize}

\item {\bf Declaration of LLM usage}
    \item[] Question: Does the paper describe the usage of LLMs if it is an important, original, or non-standard component of the core methods in this research? Note that if the LLM is used only for writing, editing, or formatting purposes and does \emph{not} impact the core methodology, scientific rigor, or originality of the research, declaration is not required.
    \item[] Answer: \answerNA{}
    \item[] Justification: The core method, experiments, and scientific contributions do not involve LLMs as an important, original, or non-standard component. Any use of standard writing, editing, or formatting assistance, if applicable, does not affect the methodology, experimental design, scientific claims, or originality of the research.
    \item[] Guidelines:
    \begin{itemize}
        \item The answer \answerNA{} means that the core method development in this research does not involve LLMs as any important, original, or non-standard components.
        \item Please refer to our LLM policy in the NeurIPS handbook for what should or should not be described.
    \end{itemize}

\end{enumerate}

\end{document}